\documentclass{article}

\usepackage{microtype}
\usepackage{graphicx}
\usepackage{subfigure}
\usepackage{booktabs} 

\usepackage{hyperref}
\usepackage{marginnote}

\usepackage[accepted]{icml2026}

\usepackage{amsmath}
\usepackage{amssymb}
\usepackage{mathtools}
\usepackage{amsthm}

\usepackage[capitalize,noabbrev]{cleveref}

\usepackage{algorithm}
\usepackage{algorithmic}

\usepackage[textsize=tiny]{todonotes}

\usepackage{enumitem}
\usepackage{tabularx}

\usepackage{tikz}
\usetikzlibrary{positioning,arrows.meta,fit,calc}

\usepackage{xcolor}
\usepackage{graphicx}

\theoremstyle{plain}

\theoremstyle{definition}

\theoremstyle{remark}

\icmltitlerunning{LatentTrack: Sequential Weight Generation via Latent Filtering}

\begin{document}

\twocolumn[
\icmltitle{LatentTrack: Sequential Weight Generation via Latent Filtering}

\begin{icmlauthorlist}
\icmlauthor{Omer Haq}{yyy}
\end{icmlauthorlist}

\icmlaffiliation{yyy}{Independent Researcher, Baltimore, MD, USA}
\icmlcorrespondingauthor{Omer Haq}{haqomer1@gmail.com}

\icmlkeywords{Machine Learning, Sequential Models, Uncertainty, Filtering, Hypernetworks}

\vskip 0.3in
]

\printAffiliationsAndNotice{\icmlEqualContribution}  

\begin{abstract}
We introduce \emph{LatentTrack} (LT), a sequential neural architecture for online probabilistic prediction under nonstationary dynamics. LT performs causal Bayesian filtering in a low-dimensional latent space and uses a lightweight hypernetwork to generate predictive model parameters at each time step, enabling constant-time online adaptation without per-step gradient updates.

At each time step, a learned latent model predicts the next latent distribution, which is updated via amortized inference using new observations, yielding a predict--generate--update filtering framework in function space. The formulation supports both structured (Markovian) and unstructured latent dynamics within a unified objective, while Monte Carlo inference over latent trajectories produces calibrated predictive mixtures with fixed per-step cost. Evaluated on long-horizon online regression using the Jena Climate benchmark, LT consistently achieves lower negative log-likelihood and mean squared error than stateful sequential and static uncertainty-aware baselines, with competitive calibration, demonstrating that latent-conditioned function evolution is an effective alternative to traditional latent-state modeling under distribution shift.
\end{abstract}

\section{Introduction}
\label{sec:intro}

\paragraph{LT in a nutshell.}
LT is a \emph{filter in function space}: instead of filtering over observations or outputs, it tracks a low-dimensional latent state $z_t$ whose dynamics generate the evolving weights of a predictive model.
A learned transition predicts $z_t$, a hypernetwork maps $z_t$ to model parameters $\theta_t$, and new data update an amortized posterior over $z_t$.
This predict--generate--update cycle performs online inference directly over \emph{functions}, enabling continual adaptation and calibrated uncertainty without retraining.
Predictive uncertainty arises naturally from Monte Carlo mixtures over latent trajectories, yielding temporally coherent predictions with constant-time test-time updates.

\noindent\textbf{Scope.}
We focus on supervised streaming settings with known covariates, i.e., $\mathcal D_t=(x_t,y_t)$ observed at predict time.
The framework applies broadly to regression, classification, and value or policy heads, though we do not model covariate dynamics.

\noindent\textbf{Filtering and beyond.}
While our empirical evaluation focuses on causal online filtering, the LT framework naturally extends to forecasting and open-loop generation by rolling forward the learned latent transition.
We focus on filtering to isolate stability and adaptation under distribution shift, but LT is best viewed as a general latent dynamical model over predictive functions.

\noindent\textbf{Capacity allocation and structural learning.}
A central design choice in LT is how representational capacity is allocated.
Rather than increasing the expressivity of latent inference components such as the prior, posterior, or transition, LT concentrates capacity in a hypernetwork that maps latent states to predictive functions.
This shifts modeling burden from latent inference to function generation, allowing relatively simple latent dynamics to control a rich and flexible hypothesis class.
We hypothesize that this allocation yields more stable adaptation, improved pointwise accuracy, stronger predictive likelihood, and better-calibrated uncertainty under nonstationarity than architectures that rely on increasingly expressive latent dynamics.

\paragraph{Contributions and novelty.}
\begin{itemize}[leftmargin=12pt,itemsep=2pt,topsep=2pt]
\item \textbf{Function-space filtering.}
A framework for sequential Bayesian filtering in latent weight space, tracking an evolving generator of predictive functions.

\item \textbf{Predict--generate--update objective.}
A filtering ELBO derived from streaming evidence, supporting both structured (Markovian) and unstructured latent dynamics.

\item \textbf{Structural learning via weight generation.}
A capacity allocation that concentrates expressivity in function generation rather than latent inference, enabling stable online adaptation with calibrated uncertainty.

\item \textbf{Constant-time adaptation.}
$\mathcal O(1)$ per-step inference via amortized updates, with no test-time gradients or inner-loop optimization.

\item \textbf{Fair evaluation.}
A capacity--compute matching protocol that equalizes parameters and predictive samples across methods.

\item \textbf{Empirical validation.}
On long-horizon online regression using real-world climate data, LT achieves lower error on the majority of time steps, improved robustness across random seeds, and well-behaved predictive uncertainty relative to stateful and static baselines.
\end{itemize}

\begin{figure}[t]
  \centering
  \resizebox{\linewidth}{!}{%
  \begin{tikzpicture}[
    font=\small,
    >=Latex,
    box/.style={
      draw,
      rounded corners=2pt,
      align=center,
      inner sep=4.5pt,
      text width=2.55cm,
      minimum height=0.90cm,
      fill=blue!6
    },
    box2/.style={
      draw,
      rounded corners=2pt,
      align=center,
      inner sep=4.5pt,
      text width=2.55cm,
      minimum height=0.90cm,
      fill=green!6
    },
    box3/.style={
      draw,
      rounded corners=2pt,
      align=center,
      inner sep=4.5pt,
      text width=2.65cm,
      minimum height=0.90cm,
      fill=orange!7
    },
    smallbox/.style={
      draw,
      rounded corners=2pt,
      align=center,
      inner sep=5pt,
      text width=3.15cm,
      minimum height=0.90cm,
      fill=gray!8
    },
    note/.style={font=\scriptsize, align=center, text=black!75},
    solid/.style={-Latex, line width=0.9pt},
    dashedarr/.style={-Latex, line width=0.9pt, dashed},
    lab/.style={midway, fill=white, inner sep=1pt}
  ]

  \node[box] (prior) {Prior / Transition\\$p(z_t\mid \mathcal \cdot)$};

  \node[box2, right=1.35cm of prior] (hyper) {Hypernetwork\\$g_\eta(z_t)$};

  \node[box3, right=1.35cm of hyper] (pred)
    {Predictor\\$\;p_\vartheta(y\mid x;\theta_t)$};

  \node[smallbox, below=1.15cm of pred] (data)
    {Observed set\\$\mathcal D_t$};

  \node[box, fill=purple!6,
        left=2.05cm of data] (post)
    {Amortized update\\$q_\psi(z_t\mid \mathcal D_{1:t})$};

  \draw[solid] (prior) -- node[lab] {$z_t$} (hyper);
  \draw[solid] (hyper) -- node[lab] {$\theta_t$} (pred);
  \draw[solid] (pred) -- node[lab, right, font=\scriptsize] {observe} (data);

  \draw[dashedarr] (data.west) -- node[lab] {evidence} (post.east);
  \draw[dashedarr] (post.north) -- ++(0,0.50) -| node[lab] {propagate belief} (prior.south);


  \end{tikzpicture}%
  } 


    \caption{\textbf{LT} predict--generate--update loop (supervised setting).
    A latent belief $z_t$ is propagated via a causal prior/transition conditioned on a
    running summary of past observations, mapped to predictor parameters $\theta_t$
    by the hypernetwork $g_\eta$, and used to define the predictive model
    $p_\vartheta(y_t \mid x_t;\theta_t)$.
    Incoming observations then amortize a posterior update over $z_t$, enabling
    temporally causal belief filtering over predictor parameters.
    }

  \label{fig:teaser}
\end{figure}
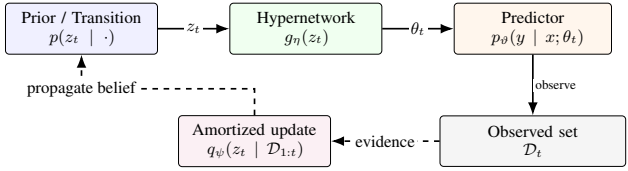

\section{Method}
\label{sec:method}

\subsection{Setup (General $\mathcal D$)}
\label{sec:method:gen}
We observe a stream $\mathcal D_{1:T}=(\mathcal D_1,\ldots,\mathcal D_T)$, where each $\mathcal D_t$ may be a singleton or a mini-batch.  
LT posits a low-dimensional latent state $z_t\in\mathbb R^d$ with emission $p(\mathcal D_t\mid z_t)$.
When $\mathcal D_t$ contains supervised pairs, a lightweight hypernetwork $g_\eta$ maps $z_t$ to predictor weights
$\theta_t=g_\eta(z_t)$ used in the emission (see~\cref{sec:method:specialize}).
We defer any discussion of controlled/summary states to~\cref{sec:method:online-variants}.

\subsection{Prior Predictive and One-Step Prior}
\label{sec:method:pp}
By the conditional independence $(\mathcal D_t\!\perp\!\mathcal D_{1:t-1}, z_{1:t-1})\mid z_t$,
the one-step evidence is
\begin{equation}
\label{eq:pp-general}
p(\mathcal D_t \mid \mathcal D_{1:t-1})
= \int p(\mathcal D_t \mid z_t)\; p_\phi(z_t \mid \mathcal D_{1:t-1})\, \mathrm dz_t .
\end{equation}
Here $p_\phi(z_t\!\mid\!\mathcal D_{1:t-1})$ denotes the \emph{one-step prior}: any causal distribution over $z_t$
that depends only on past data $\mathcal D_{1:t-1}$.

\subsection{Filtering ELBO (general, $\mathcal D$-only)}
\label{sec:method:elbo-general}
Applying Jensen’s inequality to $\log p(\mathcal D_t\!\mid\!\mathcal D_{1:t-1})$ with the amortized posterior
$q_\psi(z_t\mid\mathcal D_{1:t})$ gives the per-step filtering ELBO
\begin{equation}
\label{eq:elbo-general}
\begin{aligned}
\mathcal L_t
&= \mathbb{E}_{q_\psi(z_t\mid \mathcal D_{1:t})}
   \big[\log p(\mathcal D_t\mid z_t)\big] \\
&\quad - \mathrm{KL}\!\Big(
    q_\psi(z_t\mid \mathcal D_{1:t})
    \,\Big\|\,
    p_\phi(z_t\mid \mathcal D_{1:t-1})
   \Big).
\end{aligned}
\end{equation}
Summing over $t$ yields $\sum_{t=1}^T \mathcal L_t \le \log p(\mathcal D_{1:T})$ by the chain rule.

\paragraph{Structured replacement.}
A looser but often more stable variant replaces the marginal prior in the KL with the
transition kernel conditioned on the previous latent:
\begingroup
\setlength{\abovedisplayskip}{6pt}
\setlength{\belowdisplayskip}{6pt}
\setlength{\belowdisplayshortskip}{6pt}
\begin{equation}
\label{eq:structured-kl-replace}
\begin{alignedat}{1}
&\mathrm{KL}\!\big(q_\psi(z_t\mid \mathcal D_{1:t}) \,\|\, p_\phi(z_t\mid \mathcal D_{1:t-1})\big)\ \Rightarrow\\[-1pt]
&\mathbb{E}_{q_\psi(z_{t-1}\mid \mathcal D_{1:t-1})}
\Big[
  \mathrm{KL}\!\big(
    q_\psi(z_t\mid \mathcal D_{1:t})
    \,\|\, 
    p_\phi(z_t\mid z_{t-1}, \mathcal D_{1:t-1})
  \big)
\Big].
\end{alignedat}
\end{equation}
\endgroup

\noindent Here the outer expectation is with respect to $z_{t-1}\!\sim\!q_\psi(z_{t-1}\mid \mathcal D_{1:t-1})$.
This replacement yields the \emph{structured ELBO}; see Appendix~\ref{app:structured-elbo} for the full derivation.

\paragraph{Structured vs.\ original KL.}
The structured replacement encourages \emph{temporal coherence} by aligning the posterior at step $t$ to the
\emph{transition kernel} from $z_{t-1}$, empirically reducing latent drift and stabilizing training. It also regularizes the transition parameters $\phi$ more directly (the prior is judged on its one-step predictions), often producing smoother $z_t$ trajectories. The trade-off is that the bound is \emph{looser} (by KL convexity), so it can understate misfit to the marginalized prior, and optimization inherits extra variance from the outer expectation over $z_{t-1}$ (or its samples). Practically, it adds mild compute (sampling/integrating over $z_{t-1}$) and can be more sensitive when the
transition is misspecified; in such cases the original marginal-KL objective may provide a crisper likelihood signal.

We denote the transition-conditioned objective in Equation \eqref{eq:structured-kl-replace} as
\emph{LT-Structured} (LT-Structured), emphasizing explicit temporal structure in the latent dynamics,
and the original marginal-prior formulation as \emph{LT-Unstructured} (LT-Unstructured).

\subsection{Specializing $\mathcal D_t=(x_t,y_t)$}
\label{sec:method:specialize}
For supervised pairs $\mathcal D_t=(x_t,y_t)$, the emission factorizes as
\begin{equation}
\label{eq:emission-factor}
p(\mathcal D_t\mid z_t)=p(y_t,x_t\mid z_t)
= p(y_t\mid x_t,z_t)\,p(x_t\mid z_t),
\end{equation}
with $p(y_t\mid x_t,z_t)$ parameterized by $\theta_t=g_\eta(z_t)$ (e.g., Gaussian mean $f_{\theta_t}(x_t)$ and noise head $\Sigma_{\theta_{t}}(x_{t})$ for regression, categorical/Bernoulli heads for classification, or value/advantage heads for RL targets).

\paragraph{Known-covariate assumption (one-step forecasting).}
If $x_t$ is observed at predict time and we do not model its process,
$p(x_t\mid\mathcal D_{1:t-1})=\delta_{x_t^\star}(x_t)$ and
\begin{align}
p(\mathcal D_t\mid\mathcal D_{1:t-1})
&=p(y_t\mid x_t,\mathcal D_{1:t-1}),\\
p(y_t\mid x_t,\mathcal D_{1:t-1})
&=\!\int p_\vartheta(y_t\mid x_t;g_\eta(z_t))\,
     p_\phi(z_t\mid\mathcal D_{1:t-1})\,\mathrm dz_t .
\label{eq:pp-conditional-int}
\end{align}
Under this assumption, \eqref{eq:elbo-general} specializes to
\begin{equation}
\label{eq:elbo-conditional}
\begin{aligned}
\mathcal L_t
&= \mathbb E_{q_\psi(z_t\mid\mathcal D_{1:t})}
   \big[\log p_\vartheta(y_t\mid x_t; g_\eta(z_t))\big] \\
&\quad - \mathrm{KL}\!\Big(
   q_\psi(z_t\mid\mathcal D_{1:t})
   \,\Big\|\,
   p_\phi(z_t\mid\mathcal D_{1:t-1})
\Big).
\end{aligned}
\end{equation}
\noindent
Equation~\eqref{eq:elbo-conditional} therefore defines the core training step of \textbf{LT}: an online evidence lower bound that performs amortized Bayesian filtering over latent-generated weights. This objective underlies the \emph{predict--generate--update} loop, illustrated schematically in Figure~\ref{fig:teaser} for the online regression and filtering setting, where the latent prior/ transition distribution is conditioned causally on the running summary of past observations.



\subsection{Online training \& inference: variants and credit weighting}
\label{sec:method:online-variants}

\paragraph{Causal sequence backbone.}
LT assumes a causal summarizer that maintains a fixed-size state
$h_t = h_\psi(h_{t-1}, e_t)$ from encoded observations $e_t=\mathrm{Enc}_\psi(\mathcal D_t)$.
Any backbone exposing this interface is compatible with the framework.
All experiments in this work use a gated recurrent unit (GRU), which provides stable updates under nonstationarity with constant per-step memory.
This choice is orthogonal to LT’s probabilistic components: the prior and posterior heads consume the summarizer state without imposing assumptions on its internal structure.

\paragraph{Online training and inference.}
LT operates in an online setting, using variational filtering during training and streaming prediction at inference.
In all experiments, training is performed with chunked truncated backpropagation through time (TBPTT) over a fixed window (Algorithm~\ref{alg:latentrack-chunk}), which bounds gradient horizon while maintaining low latency.
Within each window, per-step ELBOs are combined using fixed recency weighting to prioritize recent observations.
Alternative training schedules and weighting schemes are described in Appendix~\ref{app:alt-train-algos}.

At test time, the unstructured variant predictions are generated using the prior path only.
Given the causal summary $h_{t-1}$, the prior head $\mathrm{Head}_{p_\phi}(h_{t-1})$ outputs parameters $(\mu^{(p)}_t,\log\sigma^{(p)}_t)$ defining $p_\phi(z_t\mid h_{t-1})$.
We draw $K$ samples $z_t^{(k)}\sim\mathcal N\!\big(\mu^{(p)}_t,\mathrm{diag}(\sigma^{(p)2}_t)\big)$, map them through the hypernetwork to $\theta_t^{(k)}=g_\eta(z_t^{(k)})$, and form the predictive mixture
\[
\hat p(y_t\mid x_t,\mathcal D_{1:t-1})
\;\approx\;
\frac{1}{K}\sum_{k=1}^K p_\vartheta\!\big(y_t\mid x_t;\theta_t^{(k)}\big).
\]
For Gaussian heads, this yields predictive mean
$\hat{\mu}_t(x_t)=\frac{1}{K}\sum_k \mu_{\theta_t^{(k)}}(x_t)$
and variance
$\hat{\sigma}_t^2(x_t)=\frac{1}{K}\sum_k \sigma^2_{\theta_t^{(k)}}(x_t)
+\operatorname{Var}_k\!\big[\mu_{\theta_t^{(k)}}(x_t)\big]$,
corresponding to aleatoric and epistemic components.
After observing $\mathcal D_t$, the summary state is updated causally without gradients,
$h_t = h_\psi(h_{t-1}, \mathrm{Enc}_\psi(\mathcal D_t))$, so subsequent predictions incorporate the new information.


\section{Complexity, Capacity, and Fairness}
\label{sec:ledger}

To ensure fair and interpretable comparisons, we control model capacity,
training protocol, and inference-time cost across baselines while respecting
structural differences between stateful and static models. Detailed hyperparameters and parameter count are provided in Appendix~\ref{app:table_and_figs}, Table~\ref{tab:training_protocol}, \ref{tab:param_counts}.

\noindent\textbf{Training protocol.}
All models are trained for 6 epochs and evaluated over 25 random seeds.
Extending training to 20 epochs on a subset of seeds yields no material
improvement, with performance saturating after approximately 4 epochs.
The 6-epoch budget therefore exceeds convergence while limiting overfitting
and computational variance.

Stateful models (LT, VRNN, DSSM) are trained using TBPTT with window length $W{=}256$ and fixed
recency weighting $\lambda{=}0.9$.
Latent and hidden dimensions are held constant across stateful methods
($\dim(z_t){=}8$, $\dim(h_t){=}64$) to isolate architectural differences
rather than state size.

Static baselines (MC-Dropout, Bayes-by-Backprop, Deep Ensembles) are trained
on overlapping sliding windows of length $W{=}256$ with stride $S{=}32$ and
the same recency weighting.
This results in approximately $8\times$ more parameter updates per epoch and
greater context overlap, which can improve optimization stability but does not
preserve persistent temporal state.

\noindent\textbf{KL scheduling.}
All stochastic models use linear $\beta$-annealing with $\beta_{\max}{=}1$.
Warmup horizons are aligned in terms of effective training progress:
$\beta_{\text{warmup}}{=}575$ updates for stateful models and
$\beta_{\text{warmup}}{=}4600$ for static models (approximately 3 epochs),
ensuring comparable regularization strength across update cadences.

\noindent\textbf{Capacity allocation and inference-time matching.}
LT generates the \emph{entire} predictor network at each time step via a
hypernetwork conditioned on $z_t$, requiring deliberate capacity reallocation.
Compared to VRNN and DSSM, LT employs lightweight linear-Gaussian prior,
posterior, and transition heads, allocating most parameters to predictive
weight generation.
This reflects the core hypothesis of LT—that modeling nonstationarity through
function evolution is more effective than increasing latent inference
expressivity—while remaining matched in total capacity and test-time cost.

All methods are matched by inference-time parameter count and predictive
sampling budget.
Total parameters are constrained to approximately $20\text{k}$ ($\pm10\%$),
with stricter matching of \emph{active test-time parameters} to $20\text{k}$
($\pm5\%$), excluding components not used during inference (e.g., amortized
posteriors in DSSM and LT-Unstructured).
Predictive sampling is equalized, with non-ensemble methods using $K{=}1$
sample during training and $K{=}100$ at evaluation. Deep Ensembles using
$M{=}10$ members throughout.


\begin{table*}[t]
\centering
\footnotesize
\setlength{\tabcolsep}{6pt}
\renewcommand{\arraystretch}{1.15}

\begin{tabular}{l|cccccc}
\toprule
\multicolumn{7}{c}{\textbf{Representative Seed: Temporal Metrics Across Time}} \\
\midrule
\multicolumn{7}{c}{\textbf{Negative Log-Likelihood (NLL)}} \\
\midrule
\textbf{Model} &
Mean $\downarrow$ &
Trimmed Mean $\downarrow$ &
Median $\downarrow$ &
Trimmed Median $\downarrow$ &
\% Rank-1 $\uparrow$ &
\% Rank $\geq$2 $\uparrow$ \\
\midrule
LT (Structured)
 & \textbf{6.29} & \textbf{2.44} & \textbf{2.33} & \textbf{2.32} & \textbf{58.8} & \textbf{79.6} \\
LT (Unstructured)
 & $1.22{\times}10^{10}$ & 2.73 & 2.62 & 2.61 & 17.6 & 65.8 \\
DSSM
 & 2.97 & 2.87 & 2.95 & 2.93 & 17.8 & 38.1 \\
VRNN
 & 3.44 & 3.41 & 3.39 & 3.38 & 0.9 & 4.6 \\
MC-Dropout
 & 23.19 & 10.92 & 4.56 & 4.49 & 0.9 & 3.4 \\
Deep Ensembles
 & 29.85 & 28.27 & 10.62 & 10.14 & 0.1 & 1.2 \\
BBB
 & 19.20 & 17.57 & 8.24 & 7.86 & 3.9 & 7.4 \\
\midrule
\multicolumn{7}{c}{\textbf{Mean Squared Error (MSE)}} \\
\midrule
\textbf{Model} &
Mean $\downarrow$ &
Trimmed Mean $\downarrow$ &
Median $\downarrow$ &
Trimmed Median $\downarrow$ &
\% Rank-1 $\uparrow$ &
\% Rank $\geq$2 $\uparrow$ \\
\midrule
LT (Structured)
 & \textbf{10095.7} & \textbf{6.32} & \textbf{1.98} & \textbf{1.93} & \textbf{51.4} & \textbf{75.2} \\
LT (Unstructured)
 & 16978.1 & 16.26 & 4.68 & 4.55 & 23.2 & 70.1 \\
DSSM
 & 69.68 & 45.29 & 15.38 & 14.82 & 14.5 & 31.5 \\
VRNN
 & 112.17 & 93.25 & 46.82 & 45.60 & 2.5 & 6.6 \\
MC-Dropout
 & 14906.4 & 114.46 & 65.81 & 64.18 & 4.0 & 7.7 \\
Deep Ensembles
 & 1584.85 & 139.94 & 76.92 & 74.61 & 2.1 & 3.8 \\
BBB
 & 3543.0 & 124.60 & 69.83 & 68.13 & 2.3 & 5.1 \\
\bottomrule
\end{tabular}
\caption{
Single-seed temporal statistics computed over all time steps for the representative seed.
Trimmed statistics remove the top $1\%$ of values.
\textbf{A single data anomaly dominates the untrimmed mean MSE, whereas trimmed and median statistics reflect typical tracking behavior.}
Rank percentages indicate the fraction of time steps in which a method achieves the best (Rank-1) or top-two (Rank $\geq$2) performance, emphasizing sustained per-step accuracy rather than sensitivity to isolated outliers.
}
\label{tab:single_seed_temporal}
\end{table*}

\begin{figure}[t]
  \centering
  \includegraphics[width=\linewidth]{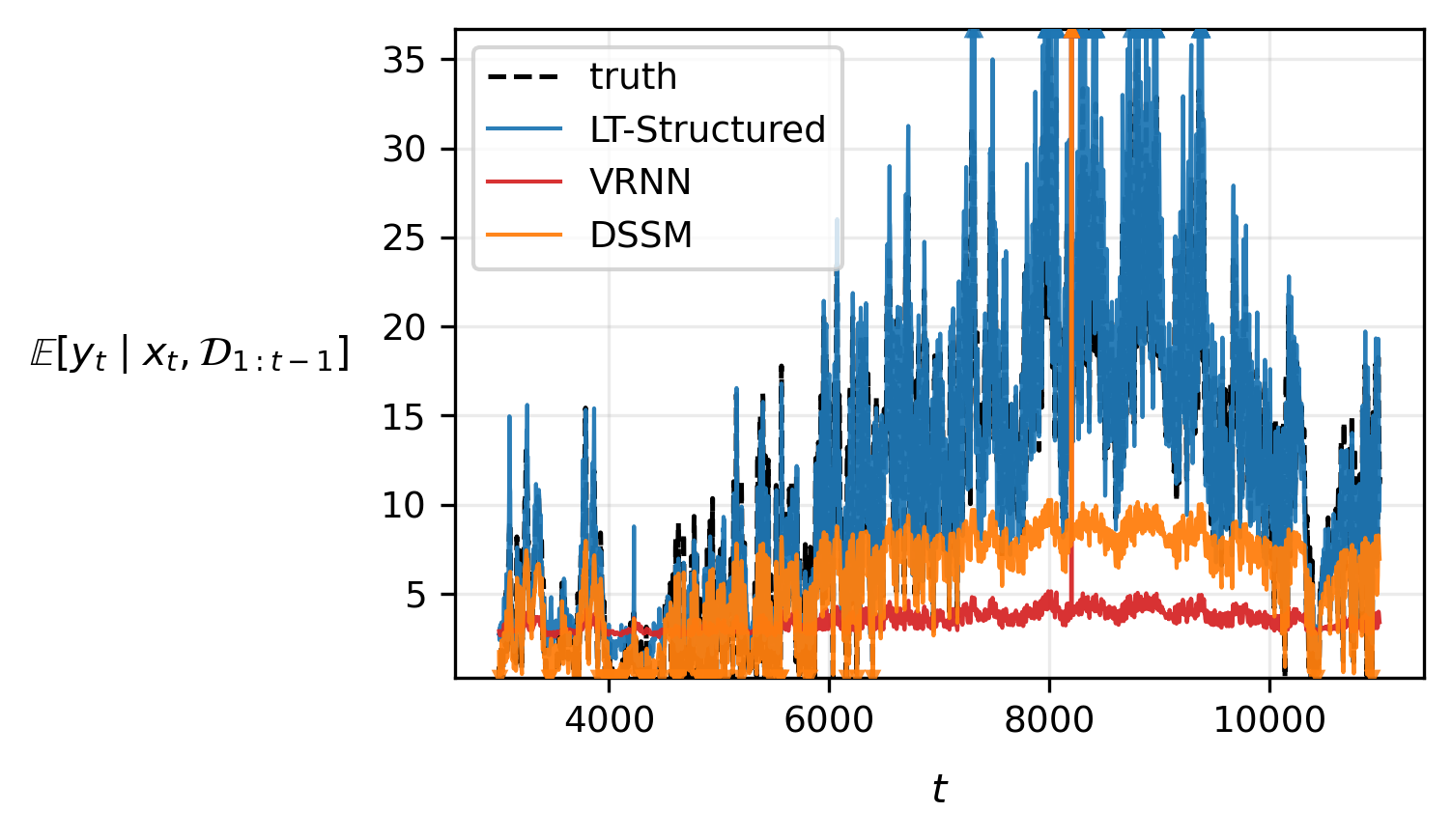}
  \caption{Predictive mean over time for the representative seed. LT tracks the signal more accurately than stateful baselines, but exhibits occasional instability consistent with the failure-rate statistics.}
  \label{fig:mean_vs_time_rep_seed}
\end{figure}



\begin{figure}[t]
\centering

\subfigure[Temporal NLL  Top 3 Winners Grid]{
    \includegraphics[width=\linewidth]{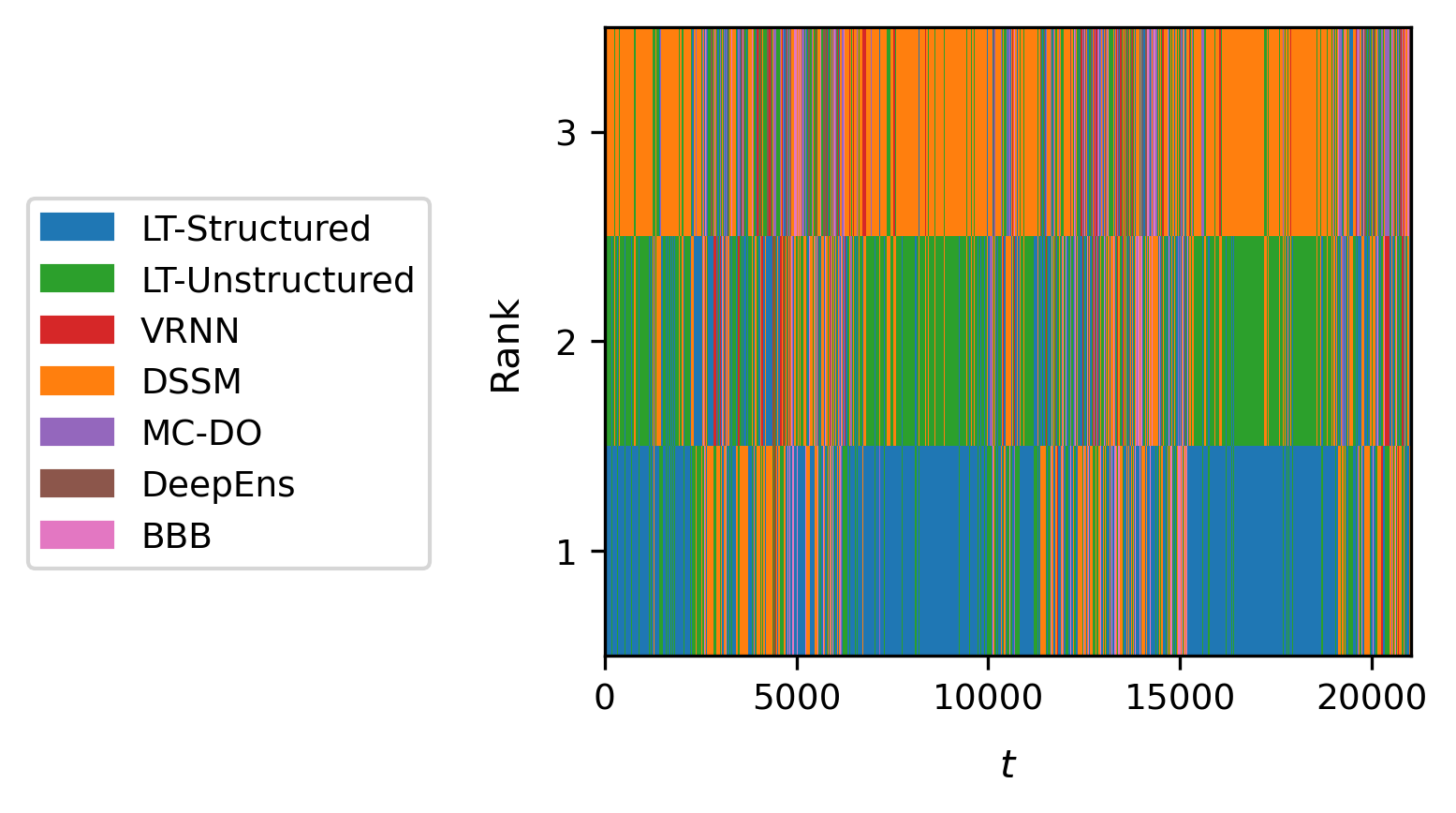}
}

\vspace{4mm}

\subfigure[Temporal MSE Top 3 Winner Grid]{
    \includegraphics[width=\linewidth]{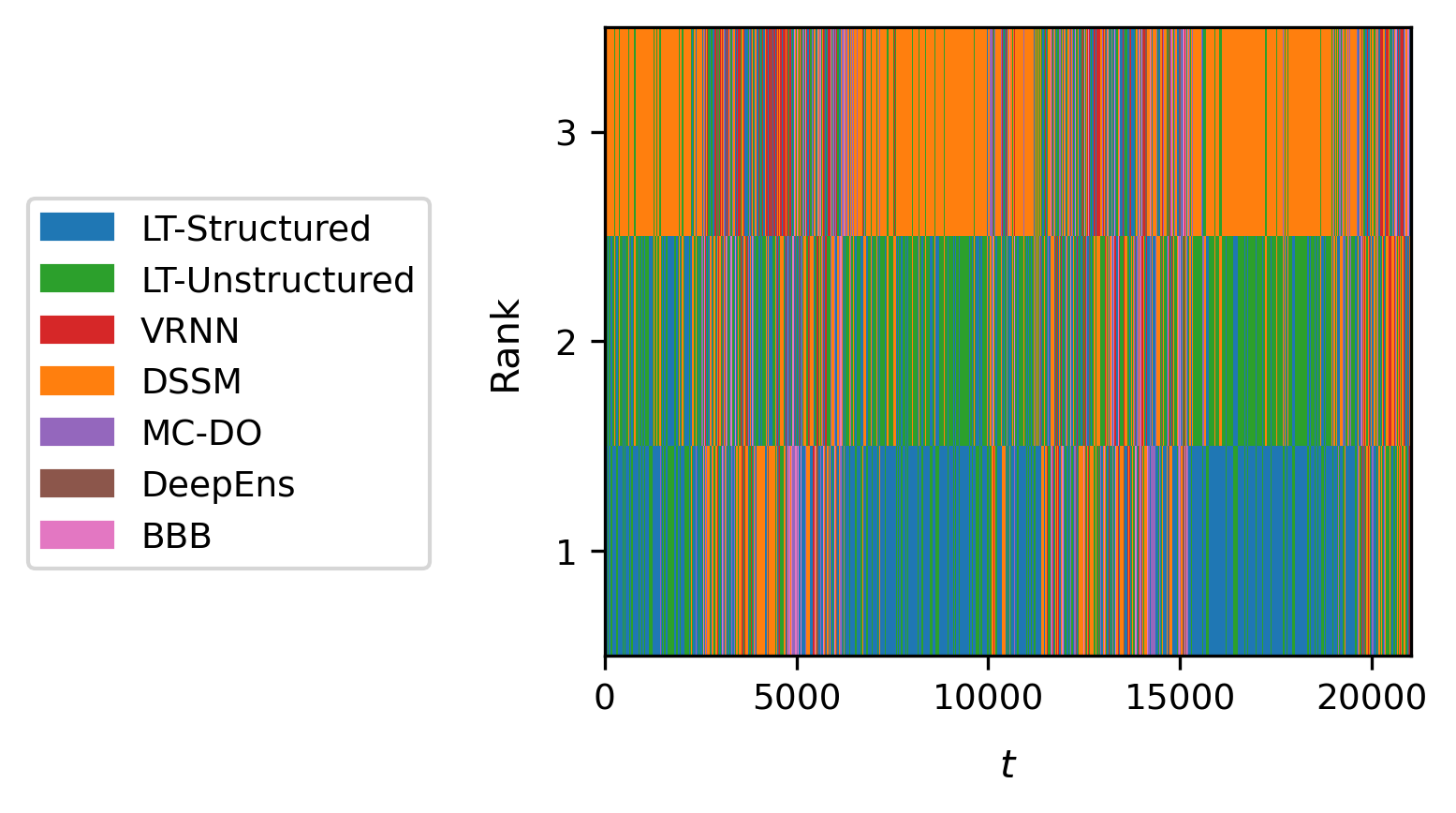}
}

\caption{Ranking stability across time.
Colors indicate the top-3 models at each time step.
(a) Per-time-step negative log-likelihood (NLL) ranking.
(b) Per-time-step mean squared error (MSE) ranking.
Lower ranks indicate better performance.
LT-Structured appears more frequently among the top-ranked methods across time,
indicating greater ranking stability under both metrics.
}
\label{fig:rank_grid}

\end{figure}



\section{Experiments}
\label{sec:expts}

\subsection{Task and Stream}
\label{sec:expts:tasks}

We evaluate LT on long-horizon online prediction using the \textbf{Jena Climate} dataset, a real-world multivariate time series with strong seasonality and nonstationary drift.
The dataset contains 14 meteorological variables recorded hourly; we downsample by a factor of 6 to obtain 6-hour resolution.
At each time step, the model predicts temperature 6 steps ahead (36 hours), conditioned on a rolling history of past observations and known time-of-day and day-of-week covariates.

LT operates in strictly causal filtering mode: predictions at time~$t$ use data up to $t\!-\!1$, after which the latent state is updated upon observing $\mathcal D_t$.
The first 70\% of the sequence is used for training, with evaluation performed on the remaining 30\%, which includes substantial seasonal transitions and regime changes.
Performance is assessed continuously over the evaluation stream, reflecting realistic online deployment without retraining or test-time optimization.

\subsection{Baselines}
\label{sec:expts:baselines}

We compare \textbf{LT} against both stateful sequential models and static uncertainty-aware baselines. As sequential baselines, we include Variational Recurrent Neural Networks (VRNNs)~\citep{chung2015vrnn,fraccaro2016sequential}, which combine recurrent dynamics with per-step latent variables trained via amortized variational inference, and Deep State-Space Models (DSSMs)~\citep{krishnan2015deep,karl2017deep,doerr2018prssm,rangapuram2018dssm}, which couple latent Markov dynamics with neural parameterizations of transitions and emissions. We also evaluate static uncertainty baselines, including MC-Dropout~\citep{gal2016dropout}, Deep Ensembles~\citep{lakshminarayanan2017simple}, and Bayes-by-Backprop (BBB)~\citep{blundell2015weight}. Full model specifications, parameterizations, and training details for VRNN and DSSM are provided in Appendix~\ref{app:vrnn-dssm}.

\subsection{Metrics and Evaluation}
\label{sec:expts:metrics}

We evaluate predictive accuracy, uncertainty quality, and temporal stability using negative log-likelihood (NLL) and mean squared error (MSE) computed over full streaming sequences. 
For each random seed, we compute temporal summaries over the entire horizon, including mean, median, and trimmed variants (excluding the top $1\%$ of values) for both metrics; trimmed and median statistics are emphasized, as temporal means are highly sensitive to rare extreme events. 
A representative seed for LT-Structured is selected as the one whose temporal mean NLL is closest to the median across seeds. 
In addition, we report per–time-step rankings for NLL and MSE.
Robustness across random initializations is assessed via a catastrophic failure rate, defined as the fraction of seeds whose maximum NLL exceeds $10^{6}$, isolating rare but severe divergence events. 
Additional uncertainty diagnostics, including probability integral transform (PIT) histograms, calibration curves, and epistemic/aleatoric variance decompositions, are reported in the appendix and omitted from the main text for brevity, as they are consistent with trends observed in NLL and MSE.

\begin{figure}[t]
\centering

\subfigure[CCDF of peak per-time-step NLL]{
    \includegraphics[width=\linewidth]{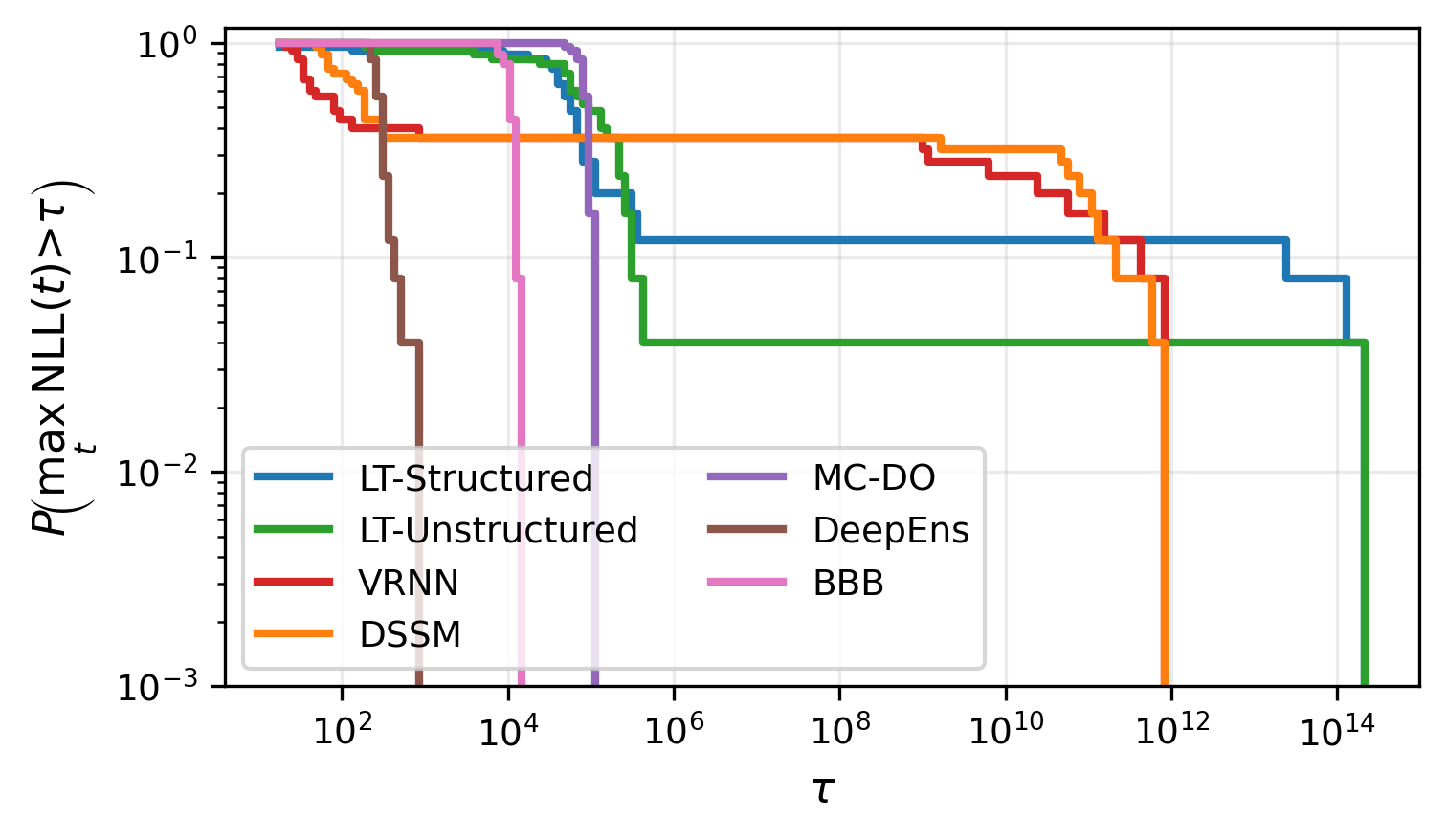}
}

\vspace{4mm}

\subfigure[Catastrophic failure rate at fixed threshold]{
    \includegraphics[width=\linewidth]{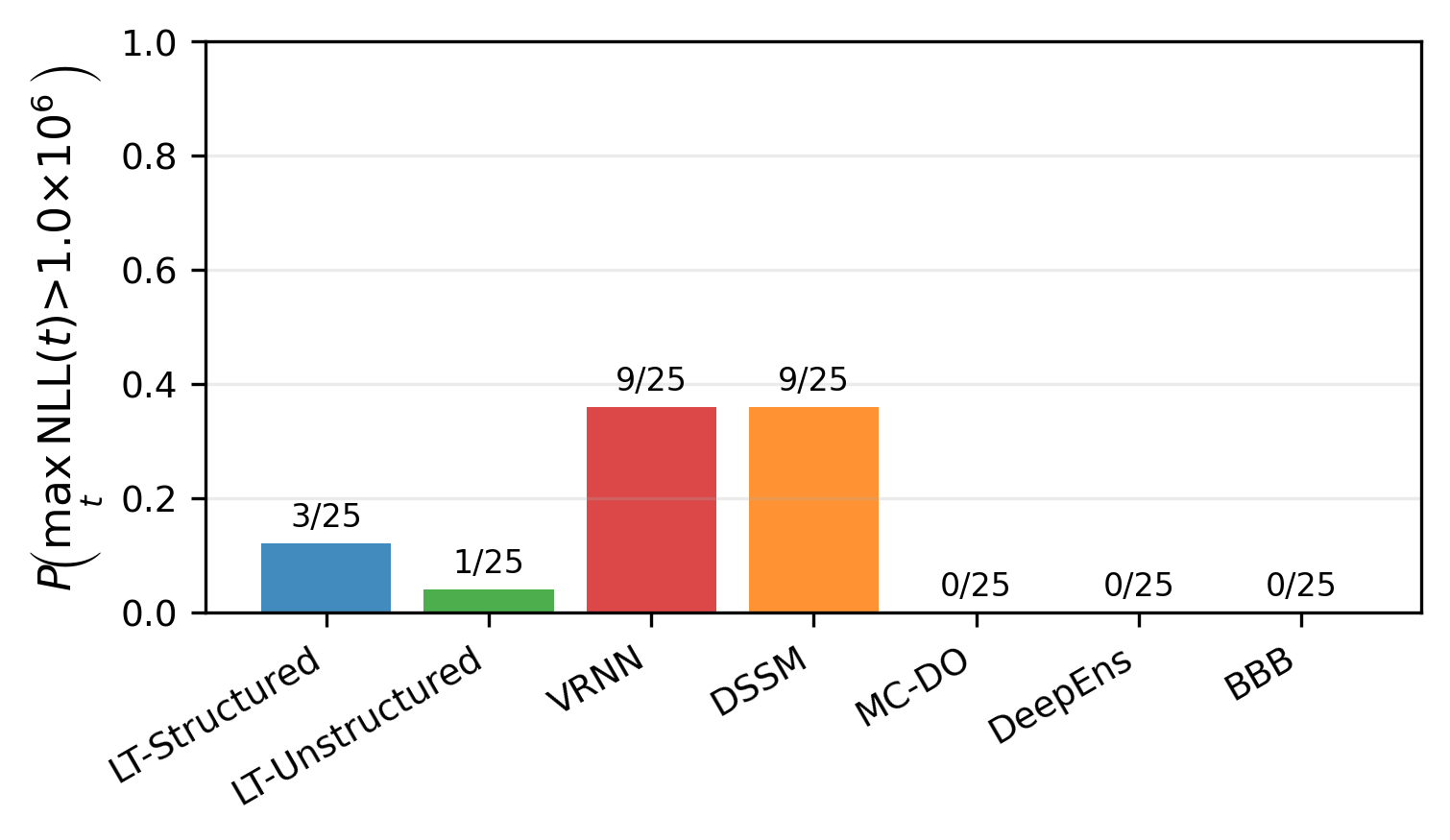}
}

\caption{
Temporal stability and catastrophic failure under distribution shift.
(a) Complementary cumulative distribution function (CCDF) of the maximum per-time-step negative log-likelihood, $\max_t \mathrm{NLL}(t)$, computed per run and aggregated across seeds, characterizing the tail behavior of rare but severe prediction failures.
(b) Fraction of runs whose peak NLL exceeds a fixed threshold $\tau = 10^6$, corresponding to a vertical slice of the CCDF in (a).
LT variants exhibit lighter tails and lower failure rates than stateful baselines, indicating improved robustness to transient instability while maintaining competitive predictive accuracy.}
\label{fig:seed_agg}

\end{figure}

\subsection{Results}

We evaluate LT under long-horizon online prediction on the Jena Climate benchmark, focusing on four aspects critical for streaming deployment: (i) sustained per-step predictive performance and accuracy rather than aggregate averages, (ii) robustness to catastrophic failures across random initializations, (iii) the role of structured versus unstructured latent dynamics, and (iv) the quality of predictive uncertainty.

\paragraph{Isolated data anomaly and evaluation implications.}
During evaluation on the Jena Climate benchmark, we observe a single isolated time step at which all models---including both stateful and static baselines---exhibit extreme predictive error. As shown in Appendix~\ref{app:table_and_figs}, (Figures~\ref{fig:anom_mean_vs_time}, \ref{fig:anom_var_vs_time}, \ref{fig:anom_mse_and_var_vs_time}, and \ref{fig:anom_nll_vs_time}), this event induces a simultaneous spike in prediction error (MSE), negative log-likelihood (NLL), and predictive uncertainty, affecting aleatoric, epistemic, and total variance. The synchronization of this response across all models indicates sensitivity to a shared extreme observation or data anomaly rather than model-specific instability.

Importantly, all models recover immediately following this event, with predictive means and uncertainty returning to baseline levels within a single time step. While such isolated extremes strongly influence temporal means and maximum-based statistics (e.g., CCDF tails), they do not reflect typical per-step behavior over long horizons and therefore should not dominate conclusions about sustained performance. For this reason, we emphasize trimmed and median statistics, together with per-time-step analyses, which more accurately characterize sustained predictive performance and calibration under nonstationary dynamics.

\paragraph{Claim 1: Sustained per-step performance (not averages).}
Figure~\ref{fig:mean_vs_time_rep_seed} shows predictive mean trajectories over a selected interval for the representative seed and the top two baselines. LT-Structured tracks the underlying signal more closely than stateful baselines, particularly during seasonal transitions and regime shifts. Over the full evaluation horizon (Appendix~\ref{app:table_and_figs}, Figure~\ref{fig:full_mean_vs_time}), LT maintains accurate tracking across extended nonstationary segments.

Table~\ref{tab:single_seed_temporal} reports temporal statistics for the representative seed, including negative log-likelihood (NLL) and mean squared error (MSE), summarized using means, medians, and trimmed variants that exclude the top 1\% of extreme values. Across both metrics, \textbf{LT-Structured achieves the strongest median and trimmed performance}, indicating superior typical per-step accuracy and probabilistic quality. These statistics emphasize representative behavior within a single long-horizon deployment, which is more informative than the temporal mean, which in this setting is dominated by a single isolated data anomaly.

Per-time-step rankings further reinforce this conclusion. As summarized in Table~\ref{tab:single_seed_temporal} and visualized in Figure~\ref{fig:rank_grid}, LT-Structured achieves first-place ranking for NLL in $58.8\%$ of time steps and ranks second or better in $79.6\%$ of time steps. For MSE, it attains first place in $51.4\%$ of time steps and ranks second or better in $75.2\%$ of time steps. LT-Unstructured also performs favorably, consistently trailing only the structured variant. Together, these results indicate that LT’s advantage reflects sustained per-step performance rather than sensitivity to aggregate averaging.

\paragraph{Claim 2: Robustness to nonstationarity across random initializations.}
To assess robustness across random initializations, we analyze catastrophic failures aggregated over seeds. Figure~\ref{fig:seed_agg} reports the complementary cumulative distribution function (CCDF) of the maximum per-time-step NLL attained within each run, together with a fixed-threshold failure rate defined as the fraction of seeds whose peak NLL exceeds $10^6$. This analysis isolates rare but severe divergence events that are not reflected in within-run median or ranking statistics and typically associated with the data anomalies discussed above.

Under this criterion, both LT variants exhibit \textbf{significantly lower failure probabilities} than VRNN and DSSM, corresponding to fewer runs that experience catastrophic divergence. LT-Unstructured achieves the lowest catastrophic failure probability ($4\%$), followed by LT-Structured ($12\%$), substantially outperforming other stateful models.

While the CCDF indicates that, conditional on failure, LT can exhibit larger peak NLL values than some baselines, we emphasize that such peaks typically occur at the same isolated time steps across all models (Appendix~\ref{app:table_and_figs}, Figure~\ref{fig:anom_mse_and_var_vs_time} and \ref{fig:anom_nll_vs_time}). Once the catastrophic threshold is exceeded, the run is already dominated by failure, and differences in peak magnitude (e.g., $10^6$ versus $10^{10}$) are of limited practical relevance. Accordingly, our robustness analysis focuses on the \emph{frequency} of catastrophic failures across seeds rather than the precise severity of individual divergence events.

Importantly, these failures are transient: NLL, MSE, and predictive variance typically spike for a single time step before rapidly recovering. This behavior indicates sensitivity to abrupt, shared stress points in the sequence rather than persistent model instability. Static baselines also exhibit catastrophic failures under this definition; however, these events are typically less severe in magnitude.

\paragraph{Claim 3: Structured vs.\ unstructured latent dynamics.}
Comparing LT variants reveals a stability--accuracy tradeoff rather than a single dominant configuration. LT-Structured achieves slightly stronger median and trimmed NLL and MSE, reflecting improved typical accuracy and temporal coherence over the majority of time steps. In contrast, LT-Unstructured exhibits the lowest catastrophic failure probability across random initializations ($4\%$), followed by LT-Structured ($12\%$), indicating reduced tail risk under rare misalignment events.

This behavior reflects the role of latent transition structure in shaping error profiles. Enforcing temporal structure improves coherence of the latent trajectory and yields more accurate predictions for most time steps, but can increase sensitivity to rare transition misalignment conditional on failure, when the learned dynamics momentarily fail to track abrupt changes. The unstructured variant, which relies on a marginal prior rather than an explicit transition model, reduces the likelihood of persistent divergence at the cost of slightly weaker typical accuracy.

Importantly, both variants substantially outperform baselines across all evaluation criteria, indicating that the primary gains arise from latent-conditioned weight generation itself rather than the specific form of the latent dynamics.

This tradeoff also manifests in how uncertainty is expressed under extreme observations, with structured and unstructured LT variants allocating uncertainty differently between epistemic and aleatoric components (Claim~4); additional comparisons between the two variants are provided in Appendix~\ref{app:table_and_figs} (Figures~\ref{fig:lt_comp_mean_vs_time}--\ref{fig:lt_comp_nll_vs_time}).

\paragraph{Claim 4: Well-behaved uncertainty, not overconfidence.}
The performance gains achieved by LT are not driven by overconfident or miscalibrated uncertainty estimates. Improvements in median and trimmed NLL coincide with stable variance behavior, indicating that predictive accuracy gains arise from better modeling of the data rather than artificial variance inflation.

Additional diagnostics reported in Appendix~\ref{app:table_and_figs} support this conclusion. Probability integral transform (PIT) histograms (Figure~\ref{fig:calib_pit}) and calibration curves (Figure~\ref{fig:calib_curve}) show reduced boundary mass and closer adherence to the uniform reference relative to baselines, consistent with improved probabilistic calibration. The calibration curves are computed over the central $[1,99]\%$ predictive mass, ensuring that they reflect typical behavior rather than being dominated by the isolated data anomaly.

Decompositions of predictive uncertainty into aleatoric and epistemic components over time (Figure~\ref{fig:full_var_vs_time}) show that LT maintains stable uncertainty levels over long horizons, with only transient deviations under extreme observations. All stateful models exhibit a synchronized spike in uncertainty at the isolated data anomaly, followed by immediate recovery to baseline levels, indicating short-lived stress rather than persistent miscalibration.

Decompositions of predictive uncertainty into aleatoric and epistemic components over time (Appendix~\ref{app:table_and_figs}, Figure~\ref{fig:full_var_vs_time}) show that LT maintains stable uncertainty levels over long horizons, with transient stress responses concentrated at a single extreme observation. Although all models exhibit a brief peak in total variance, MSE, and NLL at this anomaly (Figures~\ref{fig:anom_mse_and_var_vs_time}--\ref{fig:anom_nll_vs_time}), the \emph{composition} of that peak differs (Figure~\ref{fig:anom_var_vs_time}): LT-Unstructured drives the response almost entirely through an epistemic spike while its aleatoric variance collapses to near-zero, whereas LT-Structured shows a pronounced epistemic spike with only a modest aleatoric increase. In contrast, VRNN and DSSM respond primarily by inflating aleatoric variance with comparatively little epistemic increase, effectively treating the event as observation noise rather than explicit model mismatch. Across all methods, uncertainty returns to baseline within one time step, consistent with a shared isolated data anomaly rather than persistent overconfidence or long-term instability; from a filtering perspective, the LT variants (particularly LT-Unstructured) are therefore well suited for innovation-based gating or reset mechanisms, since anomalous events are expressed \emph{entirely} through epistemic uncertainty rather than being masked by variance inflation.

\noindent\textbf{Summary.}
Across all evaluations, LT demonstrates stronger sustained per-step performance and improved temporal robustness compared to all baselines. The structured and unstructured variants occupy complementary points on a stability--accuracy tradeoff, while both retain well-behaved and calibrated uncertainty. These gains are achieved despite deliberately lightweight latent inference components, supporting the hypothesis that modeling function evolution via latent-conditioned weight generation is a more effective allocation of capacity than increasing the expressivity of latent inference mechanisms alone.


\section{Related Work}
\label{sec:related}

\paragraph{Deep state--space and variational sequence models.}
LT is related to deep state--space models that learn latent dynamics for sequential data, including the Deep Kalman Filter (DKF) \citep{krishnan2015deep}, the Deep Variational Bayes Filter (DVBF) \citep{karl2017deep}, and the Variational Recurrent Neural Network (VRNN) family \citep{chung2015vrnn,fraccaro2016sequential}.
These approaches perform filtering in a latent space that directly governs observations.
In contrast, LT performs filtering in \emph{function space}: the latent variable generates the parameters of a predictive model rather than the observations themselves, enabling temporally coherent uncertainty over functions with constant-time online updates.

\noindent\textbf{Hypernetworks and weight--space inference.}
Hypernetworks and weight-generation methods \citep{ha2016hypernetworks,krueger2017bayesian,pawlowski2017implicit} treat model parameters as outputs of a secondary network, supporting amortized Bayesian inference in weight space.
LT extends these ideas to sequential settings by evolving the latent variable through time, inducing smooth weight trajectories.
This connects hypernetwork-based inference with classical Bayesian filtering ideas such as ensemble Kalman and particle filters, while remaining fully differentiable and end-to-end trainable.

\noindent\textbf{Online and continual learning.}
Prior work on online adaptation typically relies on gradient-based meta-learning \citep{finn2017maml,al2018continuous,nagabandi2018meta}, replay buffers \citep{lopez2017gradient,chaudhry2019continual}, or parameter regularization \citep{kirkpatrick2017ewc,zenke2017synaptic}.
LT instead performs amortized, gradient-free adaptation at each time step via latent filtering, enabling causal updates with constant test-time cost and without catastrophic forgetting.
This is related to amortized continual inference \citep{osterlund2022continual}, but operates explicitly in weight space rather than activation space.

\noindent\textbf{Function--space inference and temporal forecasting.}
Gaussian processes and Neural Processes \citep{garnelo2018neural,kim2019attentive} perform Bayesian inference over functions, while temporal forecasting models such as DeepAR \citep{rangapuram2018deepar}, Deep State Space Models (DSSM) \citep{doersch2019cold}, and probabilistic transformers \citep{gasthaus2020probabilistic} predict distributions over future observations.
Recent work on Function Distribution Networks (FDNs) \citep{haq2025fdn} similarly performs amortized inference in function space by learning distributions over network weights for static prediction tasks.
LT can be viewed as a sequential, parametric analogue that tracks a trajectory of function generators, yielding calibrated predictive mixtures with fixed per-step update cost in non-stationary streams.

\noindent\textbf{Summary.}
LT lies at the intersection of deep state--space modeling, hypernetwork-based weight generation, and amortized Bayesian filtering, distinguished by moving the filtering operation from data space to function space.


\section{Discussion and Limitations}
\label{sec:discussion}


LT reflects a deliberate allocation of model capacity: rather than increasing the expressivity of latent inference components such as the prior, posterior, or transition heads, most capacity is placed in a hypernetwork that generates predictive functions conditioned on a latent state. Consequently, LT employs substantially lighter latent inference components---in our implementation, simple linear heads---than stateful baselines such as VRNN and DSSM, which rely on more expressive nonlinear latent dynamics to compensate for fixed predictors. Despite this asymmetry, and despite static baselines receiving substantially more frequent parameter updates via overlapping windows, LT achieves stronger sustained performance, suggesting that modeling nonstationarity directly in function space is a more effective and stable use of capacity for long-horizon online prediction.

An important consideration in interpreting these results is the presence of an isolated extreme observation that induces a sharp, synchronized degradation across all evaluated models. In this case, catastrophic deviations in predictive error and likelihood occur simultaneously for stateful and static baselines alike, indicating sensitivity to a shared data anomaly rather than a model-specific failure mode. Notably, the more pronounced response exhibited by LT at this anomaly is consistent with its lower typical error elsewhere and may be advantageous in practice, as sharper excursions facilitate reliable detection and gating of extreme events. Such mitigation mechanisms are standard in filtering systems and are orthogonal to the core architectural contribution, which governs sustained per-step behavior over long horizons.

Finally, LT inherits the parameter scaling behavior of hypernetwork-based models, as increasing predictor or conditioning capacity increases the total parameter count. In practical deployments, this limitation can be mitigated by using LT as an adaptive prediction head on top of a fixed or pretrained backbone, preserving online adaptability while concentrating most capacity in shared representations.

\section{Conclusion}
\label{sec:conclusion}

LT introduces a predict--generate--update filtering framework in function space, enabling constant-time online adaptation under nonstationary data streams. By conditioning predictive functions on a latent state rather than directly evolving high-capacity latent dynamics, the model achieves superior sustained predictive performance and temporal robustness despite deliberately lightweight latent inference components. Empirical results on long-horizon benchmarks demonstrate consistently lower negative log-likelihood and mean squared error, with competitive calibration, relative to both stateful and static baselines. Together, these findings indicate that function-space adaptation provides a principled and effective alternative to traditional latent-state modeling for online probabilistic prediction under distribution shift.


\section*{Impact Statement}
This paper presents work whose goal is to advance the field of Machine Learning. There are many potential societal consequences of our work, none of which we feel must be specifically highlighted here. 

\newpage


\bibliography{example_paper}

\begin{thebibliography}{26}
\providecommand{\natexlab}[1]{#1}
\providecommand{\url}[1]{\texttt{#1}}
\expandafter\ifx\csname urlstyle\endcsname\relax
  \providecommand{\doi}[1]{doi: #1}\else
  \providecommand{\doi}{doi: \begingroup \urlstyle{rm}\Url}\fi

\bibitem[Al-Shedivat et~al.(2018)Al-Shedivat, Bansal, Burda, Sutskever, Mordatch, and Abbeel]{al2018continuous}
Al-Shedivat, M., Bansal, T., Burda, Y., Sutskever, I., Mordatch, I., and Abbeel, P.
\newblock Continuous adaptation via meta-learning in nonstationary and competitive environments.
\newblock In \emph{International Conference on Learning Representations}, 2018.

\bibitem[Blundell et~al.(2015)Blundell, Cornebise, Kavukcuoglu, and Wierstra]{blundell2015weight}
Blundell, C., Cornebise, J., Kavukcuoglu, K., and Wierstra, D.
\newblock Weight uncertainty in neural networks.
\newblock In \emph{Proceedings of the 32nd International Conference on Machine Learning (ICML)}, pp.\  1613--1622. PMLR, 2015.

\bibitem[Chaudhry et~al.(2019)Chaudhry, Ranzato, Rohrbach, and Elhoseiny]{chaudhry2019continual}
Chaudhry, A., Ranzato, M., Rohrbach, M., and Elhoseiny, M.
\newblock Continual learning with tiny episodic memories.
\newblock In \emph{Advances in Neural Information Processing Systems}, 2019.

\bibitem[Chung et~al.(2015)Chung, Kastner, Dinh, Goel, Courville, and Bengio]{chung2015vrnn}
Chung, J., Kastner, K., Dinh, L., Goel, K., Courville, A., and Bengio, Y.
\newblock A recurrent latent variable model for sequential data.
\newblock In \emph{Advances in Neural Information Processing Systems}, 2015.

\bibitem[Doerr et~al.(2018)Doerr, Daniel, Schiegg, Nguyen-Tuong, Schaal, Toussaint, and Trimpe]{doerr2018prssm}
Doerr, A., Daniel, C., Schiegg, M., Nguyen-Tuong, D., Schaal, S., Toussaint, M., and Trimpe, S.
\newblock Probabilistic recurrent state-space models.
\newblock In Dy, J. and Krause, A. (eds.), \emph{Proceedings of the 35th International Conference on Machine Learning}, volume~80 of \emph{Proceedings of Machine Learning Research}, pp.\  1280--1289, Stockholm, Sweden, 2018. PMLR.
\newblock URL \url{https://proceedings.mlr.press/v80/doerr18a.html}.

\bibitem[Doersch et~al.(2019)Doersch, Zoran, Garnelo, Eslami, Rezende, Botvinick, Osindero, and Gregor]{doersch2019cold}
Doersch, C., Zoran, D., Garnelo, M., Eslami, S. M.~A., Rezende, D.~J., Botvinick, M., Osindero, S., and Gregor, K.
\newblock Cold fusion: Training seq2seq models together with language models.
\newblock In \emph{International Conference on Learning Representations}, 2019.

\bibitem[Finn et~al.(2017)Finn, Abbeel, and Levine]{finn2017maml}
Finn, C., Abbeel, P., and Levine, S.
\newblock Model-agnostic meta-learning for fast adaptation of deep networks.
\newblock In \emph{International Conference on Machine Learning}, 2017.

\bibitem[Fraccaro et~al.(2016)Fraccaro, S{\o}nderby, Paquet, and Winther]{fraccaro2016sequential}
Fraccaro, M., S{\o}nderby, S.~K., Paquet, U., and Winther, O.
\newblock Sequential neural models with stochastic layers.
\newblock In \emph{Advances in Neural Information Processing Systems}, 2016.

\bibitem[Gal \& Ghahramani(2016)Gal and Ghahramani]{gal2016dropout}
Gal, Y. and Ghahramani, Z.
\newblock Dropout as a bayesian approximation: Representing model uncertainty in deep learning.
\newblock In \emph{Proceedings of the 33rd International Conference on Machine Learning (ICML)}, pp.\  1050--1059. PMLR, 2016.

\bibitem[Garnelo et~al.(2018)Garnelo, Rosenbaum, Maddison, Ramalho, Saxton, Shanahan, Teh, Rezende, and Eslami]{garnelo2018neural}
Garnelo, M., Rosenbaum, D., Maddison, C., Ramalho, T., Saxton, D., Shanahan, M., Teh, Y.~W., Rezende, D., and Eslami, S. M.~A.
\newblock Neural processes.
\newblock In \emph{International Conference on Machine Learning}, 2018.

\bibitem[Gasthaus et~al.(2020)Gasthaus, Benidis, Wang, Bohlke-Schneider, Flunkert, Salinas, Januschowski, Maddix, Rangapuram, and Kurle]{gasthaus2020probabilistic}
Gasthaus, J., Benidis, K., Wang, Y., Bohlke-Schneider, M., Flunkert, V., Salinas, D., Januschowski, T., Maddix, D., Rangapuram, S.~S., and Kurle, R.
\newblock Probabilistic transformer: Modelling time series with distributions.
\newblock In \emph{Advances in Neural Information Processing Systems}, 2020.

\bibitem[Ha et~al.(2016)Ha, Dai, and Le]{ha2016hypernetworks}
Ha, D., Dai, A., and Le, Q.~V.
\newblock Hypernetworks.
\newblock \emph{arXiv preprint arXiv:1609.09106}, 2016.

\bibitem[Haq(2025)]{haq2025fdn}
Haq, O.
\newblock Function distribution networks.
\newblock \emph{arXiv preprint arXiv:2510.17794}, 2025.

\bibitem[Karl et~al.(2017)Karl, Soelch, Bayer, and van~der Smagt]{karl2017deep}
Karl, M., Soelch, M., Bayer, J., and van~der Smagt, P.
\newblock Deep variational bayes filters: Unsupervised learning of state space models from raw data.
\newblock In \emph{International Conference on Learning Representations}, 2017.

\bibitem[Kim et~al.(2019)Kim, Mnih, Schwarz, Garnelo, Eslami, Rosenbaum, Vinyals, and Teh]{kim2019attentive}
Kim, H., Mnih, A., Schwarz, J., Garnelo, M., Eslami, S. M.~A., Rosenbaum, D., Vinyals, O., and Teh, Y.~W.
\newblock Attentive neural processes.
\newblock In \emph{International Conference on Learning Representations}, 2019.

\bibitem[Kirkpatrick et~al.(2017)Kirkpatrick, Pascanu, Rabinowitz, Veness, et~al.]{kirkpatrick2017ewc}
Kirkpatrick, J., Pascanu, R., Rabinowitz, N., Veness, J., et~al.
\newblock Overcoming catastrophic forgetting in neural networks.
\newblock In \emph{Proceedings of the National Academy of Sciences}, 2017.

\bibitem[Krishnan et~al.(2015)Krishnan, Shalit, and Sontag]{krishnan2015deep}
Krishnan, R.~G., Shalit, U., and Sontag, D.
\newblock Deep kalman filters.
\newblock In \emph{Proceedings of the 31st Conference on Uncertainty in Artificial Intelligence}, 2015.

\bibitem[Krueger et~al.(2017)Krueger, Huang, Islam, Turner, and Lacoste]{krueger2017bayesian}
Krueger, D., Huang, C., Islam, R., Turner, R., and Lacoste, A.
\newblock Bayesian hypernetworks.
\newblock In \emph{International Conference on Learning Representations}, 2017.

\bibitem[Lakshminarayanan et~al.(2017)Lakshminarayanan, Pritzel, and Blundell]{lakshminarayanan2017simple}
Lakshminarayanan, B., Pritzel, A., and Blundell, C.
\newblock Simple and scalable predictive uncertainty estimation using deep ensembles.
\newblock In \emph{Advances in Neural Information Processing Systems (NeurIPS)}, pp.\  6402--6413, 2017.

\bibitem[Lopez-Paz \& Ranzato(2017)Lopez-Paz and Ranzato]{lopez2017gradient}
Lopez-Paz, D. and Ranzato, M.
\newblock Gradient episodic memory for continual learning.
\newblock In \emph{Advances in Neural Information Processing Systems}, 2017.

\bibitem[Nagabandi et~al.(2018)Nagabandi, Finn, and Levine]{nagabandi2018meta}
Nagabandi, A., Finn, C., and Levine, S.
\newblock Learning to adapt in dynamic, real-world environments through meta-reinforcement learning.
\newblock In \emph{International Conference on Learning Representations}, 2018.

\bibitem[Osterlund et~al.(2022)Osterlund, Zoubin, and Turner]{osterlund2022continual}
Osterlund, E., Zoubin, G., and Turner, R.
\newblock Continual inference: A neural architecture for online bayesian inference.
\newblock \emph{arXiv preprint arXiv:2203.14209}, 2022.

\bibitem[Pawlowski et~al.(2017)Pawlowski, Brock, Lee, Rajchl, and Glocker]{pawlowski2017implicit}
Pawlowski, N., Brock, A., Lee, J., Rajchl, M., and Glocker, B.
\newblock Implicit weight uncertainty in neural networks.
\newblock In \emph{International Conference on Learning Representations}, 2017.

\bibitem[Rangapuram et~al.(2018{\natexlab{a}})Rangapuram, Seeger, Gasthaus, Stella, Wang, and Januschowski]{rangapuram2018deepar}
Rangapuram, S.~S., Seeger, M.~W., Gasthaus, J., Stella, L., Wang, Y., and Januschowski, T.
\newblock Deepar: Probabilistic forecasting with autoregressive recurrent networks.
\newblock In \emph{International Journal of Forecasting}, 2018{\natexlab{a}}.

\bibitem[Rangapuram et~al.(2018{\natexlab{b}})Rangapuram, Seeger, Gasthaus, Stella, Wang, and Januschowski]{rangapuram2018dssm}
Rangapuram, S.~S., Seeger, M.~W., Gasthaus, J., Stella, L., Wang, Y., and Januschowski, T.
\newblock Deep state space models for time series forecasting.
\newblock In \emph{Advances in Neural Information Processing Systems}, volume~31. Curran Associates, Inc., 2018{\natexlab{b}}.
\newblock URL \url{https://papers.nips.cc/paper/8004-deep-state-space-models-for-time-series-forecasting}.

\bibitem[Zenke et~al.(2017)Zenke, Poole, and Ganguli]{zenke2017synaptic}
Zenke, F., Poole, B., and Ganguli, S.
\newblock Continual learning through synaptic intelligence.
\newblock In \emph{International Conference on Machine Learning}, 2017.

\end{thebibliography}
\bibliographystyle{icml2026}

\newpage
\appendix
\onecolumn

\appendix

\section{Deriving the Looser Structured ELBO}
\label{app:structured-elbo}

\paragraph{From the prior predictive to an explicit transition.}
\begin{equation}
p(\mathcal D_t \mid \mathcal D_{1:t-1})
= \int p(\mathcal D_t \mid z_t)\, p_\phi(z_t \mid \mathcal D_{1:t-1})\, \mathrm dz_t ,
\end{equation}
with Chapman--Kolmogorov (CK)
\begin{equation}
\label{eq:ck}
p_\phi(z_t \mid \mathcal D_{1:t-1})
= \int p_\phi(z_t \mid z_{t-1}, \mathcal D_{1:t-1})\,
        p_\phi(z_{t-1} \mid \mathcal D_{1:t-1})\, \mathrm dz_{t-1}.
\end{equation}

\paragraph{Standard filtering ELBO.}
Using $q_\psi(z_t \mid \mathcal D_{1:t})$,
\begin{equation}
\label{eq:std-elbo-app}
\begin{aligned}
\log p(\mathcal D_t \mid \mathcal D_{1:t-1})
\;\ge\;
&\ \mathbb E_{q_\psi(z_t\mid\mathcal D_{1:t})}\!\big[\log p(\mathcal D_t \mid z_t)\big] \\
&\ -\ \mathrm{KL}\!\Big(q_\psi(z_t\mid\mathcal D_{1:t}) \,\Big\|\, p_\phi(z_t\mid \mathcal D_{1:t-1})\Big).
\end{aligned}
\end{equation}

\paragraph{Expand the KL as an integral (keep $q\log q$ intact).}
\begin{equation}
\label{eq:kl-int}
\mathrm{KL}\!\Big(q_\psi \,\Big\|\, p_\phi(\cdot\mid \mathcal D_{1:t-1})\Big)
=
\int q_\psi(z_t\mid\mathcal D_{1:t})
\Big[
  \underbrace{\log q_\psi(z_t\mid\mathcal D_{1:t})}_{\text{$q\log q$ (unchanged)}}
  - \log p_\phi(z_t\mid \mathcal D_{1:t-1})
\Big]\mathrm dz_t .
\end{equation}

\paragraph{Insert CK into the \(-\log\) term and apply Jensen (log-sum).}
Substitute \eqref{eq:ck} into the \(-\log)\) term of \eqref{eq:kl-int}:
\begin{equation*}
-\log p_\phi(z_t\mid \mathcal D_{1:t-1})
= -\log \!\int p_\phi(z_t\mid z_{t-1},\mathcal D_{1:t-1})\,
                 p_\phi(z_{t-1}\mid \mathcal D_{1:t-1})\, \mathrm dz_{t-1}.
\end{equation*}
By concavity of $\log$ (log-sum inequality), for each fixed $z_t$,
\begin{equation*}
-\log \!\int p_\phi(z_t\mid z_{t-1},\mathcal D_{1:t-1})\,
             p_\phi(z_{t-1}\mid \mathcal D_{1:t-1})\, \mathrm dz_{t-1}
\;\le\;
-\int p_\phi(z_{t-1}\mid \mathcal D_{1:t-1})\,
       \log p_\phi(z_t\mid z_{t-1},\mathcal D_{1:t-1})\, \mathrm dz_{t-1}.
\end{equation*}
Plugging this \emph{only} into the second term of \eqref{eq:kl-int} yields
\begin{align}
\mathrm{KL}\!\Big(q_\psi \,\Big\|\, p_\phi(\cdot\mid \mathcal D_{1:t-1})\Big)
&\le
\int q_\psi(z_t\mid\mathcal D_{1:t})
\bigg[
\log q_\psi(z_t\mid\mathcal D_{1:t})
-
\int p_\phi(z_{t-1}\mid \mathcal D_{1:t-1})\,
    \log p_\phi(z_t\mid z_{t-1},\mathcal D_{1:t-1})\,\mathrm dz_{t-1}
\bigg]\mathrm dz_t \nonumber\\
&=
\int p_\phi(z_{t-1}\mid \mathcal D_{1:t-1})
\Bigg(
\int q_\psi(z_t\mid\mathcal D_{1:t})
\Big[
\log q_\psi(z_t\mid\mathcal D_{1:t})
- \log p_\phi(z_t\mid z_{t-1},\mathcal D_{1:t-1})
\Big]\mathrm dz_t
\Bigg)\mathrm dz_{t-1} \nonumber\\
&=
\mathbb E_{p_\phi(z_{t-1}\mid \mathcal D_{1:t-1})}
\Big[
\mathrm{KL}\!\big(q_\psi(z_t\mid\mathcal D_{1:t}) \,\big\|\, p_\phi(z_t\mid z_{t-1},\mathcal D_{1:t-1})\big)
\Big].
\label{eq:kl-bound-correct}
\end{align}
(Interchanging integrals above is justified by Fubini/Tonelli.)

\paragraph{Structured ELBO (looser).}
Substitute \eqref{eq:kl-bound-correct} into \eqref{eq:std-elbo-app}:
\begin{equation}
\label{eq:elbo-structured-app}
\begin{aligned}
\mathcal L_t^{\mathrm{struct}}
&=
\mathbb E_{q_\psi(z_t\mid\mathcal D_{1:t})}\!\big[\log p(\mathcal D_t\mid z_t)\big]
-
\mathbb E_{p_\phi(z_{t-1}\mid \mathcal D_{1:t-1})}
\Big[
\mathrm{KL}\!\big(q_\psi(z_t\mid\mathcal D_{1:t}) \,\big\|\, p_\phi(z_t\mid z_{t-1},\mathcal D_{1:t-1})\big)
\Big],
\end{aligned}
\end{equation}
and thus $\mathcal L_t^{\mathrm{struct}} \le \mathcal L_t$.

\paragraph{Practical amortization.}
Replace the intractable outer expectation by the previous amortized filter:
\begin{equation}
\label{eq:elbo-structured-maintext}
\begin{aligned}
\mathcal L_t^{\mathrm{struct}}
&\approx
\mathbb E_{q_\psi(z_t\mid\mathcal D_{1:t})}\!\big[\log p(\mathcal D_t\mid z_t)\big]
-
\mathbb E_{q_\psi(z_{t-1}\mid \mathcal D_{1:t-1})}
\Big[
\mathrm{KL}\!\big(q_\psi(z_t\mid\mathcal D_{1:t}) \,\big\|\, p_\phi(z_t\mid z_{t-1},\mathcal D_{1:t-1})\big)
\Big].
\end{aligned}
\end{equation}

\paragraph{Interpretation.}
The $q\log q$ term remains unchanged; the looseness comes solely from bounding
\[
-\log\!\int p_\phi(z_t\mid z_{t-1},\mathcal D_{1:t-1})\,p_\phi(z_{t-1}\mid \mathcal D_{1:t-1})\,\mathrm dz_{t-1}
\ \le\
-\int p_\phi(z_{t-1}\mid \mathcal D_{1:t-1})\,\log p_\phi(z_t\mid z_{t-1},\mathcal D_{1:t-1})\,\mathrm dz_{t-1}.
\]
This converts the prior KL into an \emph{expected} KL to the transition, promoting temporal coherence at the cost of a looser bound.

\section{Online Training Algorithm}
\label{app:alt-train-algos}

This appendix describes the online training and inference procedures used for all stateful models evaluated in this work.
All models optimize an online variational objective over a streaming sequence and differ only in how temporal dependencies are handled during training, specifically in how gradients are accumulated, how latent states are detached, and how credit is assigned across time.
Inference is performed causally and without gradients for all methods.

All main experimental results use a single training configuration: \emph{chunked truncated backpropagation through time (TBPTT)} with \emph{fixed recency weighting}.
This choice provides a stable and computationally efficient trade-off between temporal credit assignment and memory cost, and is applied uniformly across all stateful baselines for fairness.
Other training schedules are included below for completeness and to clarify the design space, but are not used in the reported results.

In chunked TBPTT, the model processes the data stream sequentially while accumulating a credit-weighted sum of per-step ELBOs over a finite window of length $W$.
At window boundaries, a single optimizer update is performed and the recurrent state is detached to prevent gradients from propagating indefinitely backward in time.
This enforces a bounded temporal horizon while preserving causal training.

Within each window, per-step losses are combined using a \emph{fixed recency} weighting scheme,
$w_\tau = \lambda^{\,t-\tau}$ where $\lambda \in (0,1]$, which exponentially downweights older time steps and prioritizes recent observations.
The resulting weighted objective is normalized before backpropagation to ensure scale invariance across windows.
This simple causal weighting was found to be stable and effective in practice and is used throughout the paper.
We additionally implement a \emph{surprise-aware} variant that modulates the recency weights according to
$w_\tau = \lambda^{\,t-\tau}\cdot \exp\!\big(\alpha\,[\mathrm{NLL}_\tau - \mathrm{EMA\_NLL}_\tau]_+\big)$,
where $[\cdot]_+ = \max(\cdot,0)$ and $\mathrm{EMA\_NLL}_\tau$ denotes an exponential moving average of recent negative log-likelihood values.
This mechanism increases credit on unexpectedly high-error steps while preserving causality.
It is included for completeness but is not used in the reported experiments.
Pseudocode for all training and inference variants is provided below.

\begin{algorithm}[H]
\caption{LT (training): \textbf{Exact rolling-window}}
\label{alg:latentrack-exact-stride}
\begin{algorithmic}[1]
\STATE \textbf{Inputs:} window $W$, update stride $S$, weights $w_\tau>0$ (\emph{fixed recency or surprise-aware, detached}), KL schedule $\beta_t$, mixture samples $K$
\STATE Init summary $h_0$; params $\phi,\psi,\eta,\vartheta$; prior $p_\phi(z_1)$. \ \textit{// keep ring buffer of recent $\mathcal D_\tau$ and detached $s_\tau$}
\FOR{$t=1$ to $T$}
  \STATE \textbf{Advance stream (no graph kept):}
  \STATE \quad $(x_t,y_t)\!\leftarrow\!\mathcal D_t$;\ $e_t=\mathrm{Enc}_\psi(\mathcal D_t)$;\ $h_t^{\text{det}}=\mathrm{RNN}_\psi(h_{t-1}^{\text{det}},e_t)$;\ \textit{// store $h_t^{\text{det}}=\mathrm{stop\_grad}(h_t^{\text{det}})$}
  \IF{$(t \bmod S = 0)$ \textbf{or} $t=T$}
     \STATE \textbf{Recompute last $W$ under current params (fresh graph):}
     \STATE \quad $t_{\min}=\max(1,\,t-W+1)$;\ $h\leftarrow h_{t_{\min}-1}^{\text{det}}$;\ $L_{\text{acc}}\leftarrow 0$;\ $Z\leftarrow 0$
     \FOR{$\tau=t_{\min}$ to $t$}
        \STATE $(\mu^{(p)}_\tau,\log\sigma^{(p)}_\tau)=\mathrm{Head}_{p_\phi}(s)$
        \STATE $e_\tau=\mathrm{Enc}_\psi(\mathcal D_\tau)$;\ $s=\mathrm{RNN}_\psi(s,e_\tau)$
        \STATE $(\mu^{(q)}_\tau,\log\sigma^{(q)}_\tau)=\mathrm{Head}_{q_\psi}(s)$
        \STATE Sample $z_\tau^{(k)}\!\sim\!\mathcal N(\mu^{(q)}_\tau,\mathrm{diag}(\sigma^{(q)2}_\tau))$, $k=1..K$;\ $\theta_\tau^{(k)}=g_\eta(z_\tau^{(k)})$
        \STATE $\widehat{\mathbb E}\log p_\tau=\frac{1}{K}\sum_k \log p_\vartheta(y_\tau\mid x_\tau;\theta_\tau^{(k)})$
        \STATE $KL_\tau=\mathrm{KL}\!\big(\mathcal N(\mu^{(q)}_\tau,\sigma^{(q)2}_\tau)\,\big\|\,\mathcal N(\mu^{(p)}_\tau,\sigma^{(p)2}_\tau)\big)$ \ \textit{\small (structured: replace by $\ \mathbb E_{q_\psi(z_{\tau-1}\mid\mathcal D_{1:\tau-1})}\![\mathrm{KL}(q_\psi(z_\tau\mid\cdot)\|p_\phi(z_\tau\mid z_{\tau-1},\cdot))]$; see App.)}
        \STATE $\mathcal L_\tau=\widehat{\mathbb E}\log p_\tau - \beta_\tau\,KL_\tau$
        \STATE \textbf{Credit weight:} set $w_\tau$ (fixed recency or surprise-aware), then $w_\tau\!\leftarrow\!\mathrm{stop\_grad}(w_\tau)$;\quad $L_{\text{acc}}\!+\!=w_\tau\,\mathcal L_\tau$;\ $Z\!+\!=w_\tau$
     \ENDFOR
     \STATE \textbf{Backprop \& step:}\quad $\text{loss}=L_{\text{acc}}/\max(Z,\varepsilon)$;\ $\text{opt.zero\_grad(); loss.backward(); opt.step()}$
     \STATE \textbf{(optional) detach checkpoint:}\ set $h_t^{\text{det}}\leftarrow \mathrm{stop\_grad}(h_t)$ \ \textit{// keeps a clean checkpoint for future recomputes}
  \ENDIF
\ENDFOR
\end{algorithmic}
\end{algorithm}



\begin{algorithm}[H]
\caption{LT (training): \textbf{Chunk stride} (partial-window updates with TBPTT)}
\label{alg:latentrack-chunk}
\begin{algorithmic}[1]
\STATE \textbf{Inputs:} window $W$, weights $w_\tau>0$ (\emph{fixed recency or surprise-aware, detached}), KL schedule $\beta_t$, mixture samples $K$
\STATE Init $h_0$; params $\phi,\psi,\eta,\vartheta$; prior $p_\phi(z_1)$
\STATE $L_{\text{acc}}\leftarrow 0$;\quad $Z\leftarrow 0$
\FOR{$t=1$ to $T$}
  \STATE $(\mu^{(p)}_t,\log\sigma^{(p)}_t)=\mathrm{Head}_{p_\phi}(h_{t-1})$
  \STATE $(x_t,y_t)\leftarrow\mathcal D_t$;\; $e_t=\mathrm{Enc}_\psi(\mathcal D_t)$;\; $h_t=\mathrm{RNN}_\psi(h_{t-1},e_t)$
  \STATE $(\mu^{(q)}_t,\log\sigma^{(q)}_t)=\mathrm{Head}_{q_\psi}(h_t)$
  \STATE Sample $z_t^{(k)}$;\; $\theta_t^{(k)}=g_\eta(z_t^{(k)})$
  \STATE $\widehat{\mathbb E}\log p_t=\frac{1}{K}\sum_k \log p_\vartheta(y_t\mid x_t;\theta_t^{(k)})$
  \STATE $KL_t=\mathrm{KL}\big(\mathcal N(\mu^{(q)}_t,\sigma^{(q)2}_t)\,\big\|\,\mathcal N(\mu^{(p)}_t,\sigma^{(p)2}_t)\big)$ \ \textit{\small (structured: same replacement as Exact; see App.)}
  \STATE $\mathcal L_t=\widehat{\mathbb E}\log p_t - \beta_t\,KL_t$
  \STATE \textbf{Credit weight:} choose $w_t$ (fixed recency or surprise-aware), then $w_t\!\leftarrow\!\mathrm{stop\_grad}(w_t)$;\quad $L_{\text{acc}}\!+\!=w_t\,\mathcal L_t$;\quad $Z\!+\!=w_t$
  \IF{$(t-t_0+1)\ \bmod W = 0$}
     \STATE \textbf{Stride update on partial chunk:}\quad $\text{loss}=L_{\text{acc}}/\max(Z,\varepsilon)$;\; step optimizer; \textbf{reset} $L_{\text{acc}},Z$; $h_t\!\leftarrow\!\mathrm{stop\_grad}(h_t)$
  \ENDIF
\ENDFOR
\end{algorithmic}
\end{algorithm}

\begin{algorithm}[H]
\caption{LT (training): \textbf{Approx stride} micro-steps + window-end TBPTT}
\label{alg:latentrack-approx}
\begin{algorithmic}[1]
\STATE \textbf{Inputs:} window $W$, stride $S$, weights $w_\tau>0$ (\emph{fixed recency or surprise-aware, detached}), KL schedule $\beta_t$, mixture samples $K$
\STATE Init $h_0$; params $\phi,\psi,\eta,\vartheta$; prior $p_\phi(z_1)$
\STATE $t_0\leftarrow 1$
\FOR{$t=1$ to $T$}
  \IF{$t=t_0$} \STATE $h_{t-1}\leftarrow \mathrm{stop\_grad}(h_{t-1})$ \ENDIF
  \STATE $(\mu^{(p)}_t,\log\sigma^{(p)}_t)=\mathrm{Head}_{p_\phi}(h_{t-1})$
  \STATE $(x_t,y_t)\leftarrow\mathcal D_t$;\; $e_t=\mathrm{Enc}_\psi(\mathcal D_t)$;\; $h_t=\mathrm{RNN}_\psi(h_{t-1},e_t)$
  \STATE $(\mu^{(q)}_t,\log\sigma^{(q)}_t)=\mathrm{Head}_{q_\psi}(h_t)$
  \STATE Sample $z_t^{(k)}$;\; $\theta_t^{(k)}=g_\eta(z_t^{(k)})$
  \STATE $\widehat{\mathbb E}\log p_t=\frac{1}{K}\sum_k \log p_\vartheta(y_t\mid x_t;\theta_t^{(k)})$
  \STATE $KL_t=\mathrm{KL}\big(\mathcal N(\mu^{(q)}_t,\sigma^{(q)2}_t)\,\big\|\,\mathcal N(\mu^{(p)}_t,\sigma^{(p)2}_t)\big)$ \ \textit{\small (structured: same replacement as Exact; see App.)}
  \STATE $\mathcal L_t=\widehat{\mathbb E}\log p_t - \beta_t\,KL_t$
  \IF{$(t-t_0+1)\ \bmod S = 0$ \textbf{and} $(t-t_0+1)<W$}
     \STATE \textbf{Micro-step (short horizon):} treat $h_{t-1}$ as detached just for this step; step optimizer on $\mathcal L_t$
  \ENDIF
  \IF{$((t-t_0+1)=W)$ \textbf{or} $t=T$}
     \STATE \textbf{Recompute last $W$ under current params} (fresh graph for TBPTT):
     \STATE \quad $t_{\min}=\max(1,\,t-W+1)$;\ $L_{\text{acc}}\leftarrow 0$;\ $Z\leftarrow 0$;\ $h\leftarrow h_{t_{\min}-1}$ (detached)
     \FOR{$\tau=t_{\min}$ to $t$}
        \STATE \emph{// repeat the per-step block to rebuild $\mathcal L_\tau$} (sample with $K$, compute KL and $\mathcal L_\tau$ as above)
        \STATE \textbf{Credit weight:} set $w_\tau$ (fixed recency or surprise-aware, detached)
        \STATE $L_{\text{acc}}\!+\!=w_\tau\,\mathcal L_\tau$;\ $Z\!+\!=w_\tau$
     \ENDFOR
     \STATE \textbf{Window-end TBPTT:}\quad $\text{loss}=L_{\text{acc}}/\max(Z,\varepsilon)$;\; step optimizer
     \STATE $h_t\leftarrow \mathrm{stop\_grad}(h_t)$;\; $t_0\leftarrow t+1$
  \ENDIF
\ENDFOR
\end{algorithmic}
\end{algorithm}

\begin{algorithm}[H]
\caption{LT (inference): streaming prediction without gradients}
\label{alg:latentrack-infer}
\begin{algorithmic}[1]
\STATE Carry forward summary $h_{t-1}$
\STATE \textbf{Predict (prior):} $(\mu^{(p)}_t,\log\sigma^{(p)}_t)=\mathrm{Head}_{p_{\phi}}(h_{t-1})$
\STATE \textbf{Produce prediction (known covariates):} for $k=1..K$, sample $z_t^{(k)}\!\sim\!\mathcal N(\mu^{(p)}_t,\mathrm{diag}(\sigma^{(p)2}_t))$; set $\theta_t^{(k)}=g_\eta(z_t^{(k)})$; output mixture 
$\displaystyle \hat p(y_t\mid x_t,\mathcal D_{1:t-1})\approx \tfrac{1}{K}\sum_k p_\vartheta(y_t\mid x_t;\theta_t^{(k)})$
\STATE \textbf{Receive $\mathcal D_t=(x_t,y_t)$ and update state (no gradients):} encode $\mathcal D_t\!\to e_t$; $h_t=\mathrm{RNN}(h_{t-1},e_t)$
\end{algorithmic}
\end{algorithm}

\section{VRNN and DSSM baselines}
\label{app:vrnn-dssm}

We model a sequence of inputs and targets 
$\{(x_t, y_t)\}_{t=1}^T$
with latent-variable recurrent baselines.
Both methods define a conditional model
$p_\theta(y_{1:T} \mid x_{1:T})$
via per-step latent variables $z_t$ and a deterministic recurrent state
$h_t$.
They differ primarily in \emph{where} the latent $z_t$ enters the dynamics.

Throughout, we assume diagonal Gaussians
$\mathcal{N}(\mu,\Sigma)$ with
$\Sigma = \mathrm{diag}(\sigma^2)$, and write
$\mathrm{KL}(\cdot\Vert\cdot)$ for the KL between diagonal Gaussians.

\subsection{VRNN baseline}

Our VRNN baseline follows the structure of \citet{chung2015vrnn}, adapted
to conditional regression.
Let $h_t \in \mathbb{R}^{d_h}$ denote a deterministic RNN state and
$z_t \in \mathbb{R}^{d_z}$ a stochastic latent.
We parameterize the model with neural networks
$\mathrm{prior}_\theta$,
$\mathrm{dec}_\theta$,
$\phi_x$ and $\phi_z$, and a GRU cell $f_\theta$.

\paragraph{Deterministic state.}
We initialize $h_0 = 0$ and update
\begin{equation}
  h_t 
  \;=\; f_\theta\big(h_{t-1},\, \phi_x(x_t),\, \phi_z(z_t)\big),
  \label{eq:vrnn-h}
\end{equation}
i.e., the RNN state depends on both the current input and the sampled latent.

\paragraph{Latent prior and likelihood.}
Conditioned on $h_{t-1}$, the prior over $z_t$ is
\begin{equation}
  p_\theta(z_t \mid h_{t-1})
  \;=\; \mathcal{N}\!\big(
      z_t;\,\mu^p_t,\,\mathrm{diag}((\sigma^p_t)^2)
    \big),
  \quad
  [\mu^p_t, \log (\sigma^p_t)^2] 
  \;=\; \mathrm{prior}_\theta(h_{t-1}).
  \label{eq:vrnn-prior}
\end{equation}
Given $x_t$, $z_t$, and $h_t$, the emission model is
\begin{equation}
  p_\theta(y_t \mid x_t, z_t, h_t)
  \;=\;
  \mathcal{N}\!\big(
    y_t;\,\mu^y_t,\,\mathrm{diag}((\sigma^y_t)^2)
  \big),
  \quad
  [\mu^y_t, \log(\sigma^y_t)^2]
  \;=\;
  \mathrm{dec}_\theta\big([h_t, z_t, x_t]\big).
  \label{eq:vrnn-dec}
\end{equation}

\paragraph{Variational posterior.}
For training we introduce a per-step amortized posterior
\begin{equation}
  q_\phi(z_t \mid x_t, y_t, h_{t-1})
  \;=\;
  \mathcal{N}\!\big(
    z_t;\,\mu^q_t,\,\mathrm{diag}((\sigma^q_t)^2)
  \big),
  \quad
  [\mu^q_t, \log(\sigma^q_t)^2]
  \;=\;
  \mathrm{enc}_\phi\big([h_{t-1}, x_t, y_t]\big).
  \label{eq:vrnn-post}
\end{equation}
Sampling uses the standard reparameterization
$z_t = \mu^q_t + \sigma^q_t \odot \varepsilon_t$ with
$\varepsilon_t \sim \mathcal{N}(0,I)$.

\paragraph{Conditional joint (derivation).}
Fix $x_{1:T}$ and introduce the full joint over $(y_{1:T},z_{1:T},h_{1:T})$.
Using the chain rule at each time step with the ordering
$z_t \rightarrow h_t \rightarrow y_t$ gives
\begin{align}
  p_\theta(y_{1:T}, z_{1:T}, h_{1:T} \mid x_{1:T})
  &\;=\;
  \prod_{t=1}^T
  p_\theta\!\big(
    z_t \mid y_{1:t-1}, z_{1:t-1}, h_{1:t-1}, x_{1:T}
  \big)\,
  p_\theta\!\big(
    h_t \mid y_{1:t-1}, z_{1:t}, h_{1:t-1}, x_{1:T}
  \big)\,
  p_\theta\!\big(
    y_t \mid y_{1:t-1}, z_{1:t}, h_{1:t}, x_{1:T}
  \big).
  \label{eq:vrnn-chain}
\end{align}
We now impose the VRNN conditional independences implied by
\cref{eq:vrnn-h,eq:vrnn-prior,eq:vrnn-dec}:
\begin{align}
  p_\theta(z_t \mid y_{1:t-1}, z_{1:t-1}, h_{1:t-1}, x_{1:T})
  &= p_\theta(z_t \mid h_{t-1}),
  \label{eq:vrnn-ci-prior}
  \\
  p_\theta(h_t \mid y_{1:t-1}, z_{1:t}, h_{1:t-1}, x_{1:T})
  &= \delta\!\big(
    h_t - f_\theta(h_{t-1}, \phi_x(x_t), \phi_z(z_t))
  \big),
  \label{eq:vrnn-ci-state}
  \\
  p_\theta(y_t \mid y_{1:t-1}, z_{1:t}, h_{1:t}, x_{1:T})
  &= p_\theta(y_t \mid x_t, z_t, h_t),
  \label{eq:vrnn-ci-like}
\end{align}
where $\delta(\cdot)$ is the Dirac delta.
Substituting \cref{eq:vrnn-ci-prior,eq:vrnn-ci-state,eq:vrnn-ci-like} into
\cref{eq:vrnn-chain} yields
\begin{equation}
  p_\theta(y_{1:T}, z_{1:T}, h_{1:T} \mid x_{1:T})
  \;=\;
  \prod_{t=1}^T
  p_\theta(z_t \mid h_{t-1})\,
  \delta\!\big(
    h_t - f_\theta(h_{t-1}, \phi_x(x_t), \phi_z(z_t))
  \big)\,
  p_\theta(y_t \mid x_t, z_t, h_t).
  \label{eq:vrnn-joint-yzh}
\end{equation}
Integrating out the deterministic states $h_{1:T}$ gives the conditional
joint over $(y_{1:T},z_{1:T})$:
\begin{align}
  p_\theta(y_{1:T}, z_{1:T} \mid x_{1:T})
  &\;=\;
  \int
  p_\theta(y_{1:T}, z_{1:T}, h_{1:T} \mid x_{1:T})\,dh_{1:T}
  \nonumber\\
  &\;=\;
  \prod_{t=1}^T
  p_\theta(z_t \mid h_{t-1})\,
  p_\theta(y_t \mid x_t, z_t, h_t),
  \label{eq:vrnn-joint}
\end{align}
where $h_t = f_\theta(h_{t-1}, \phi_x(x_t), \phi_z(z_t))$ is the
deterministic RNN state from \cref{eq:vrnn-h}.

\paragraph{ELBO (derivation).}
We optimize the conditional log-likelihood
$\log p_\theta(y_{1:T} \mid x_{1:T})$ via a variational lower bound.
Using \cref{eq:vrnn-joint},
\begin{align}
  \log p_\theta(y_{1:T} \mid x_{1:T})
  &= \log \int p_\theta(y_{1:T}, z_{1:T} \mid x_{1:T})\,dz_{1:T}
  \nonumber\\
  &= \log \int
  q_\phi(z_{1:T} \mid x_{1:T}, y_{1:T})\,
  \frac{
    p_\theta(y_{1:T}, z_{1:T} \mid x_{1:T})
  }{
    q_\phi(z_{1:T} \mid x_{1:T}, y_{1:T})
  }\,
  dz_{1:T}
  \nonumber\\
  &\ge
  \mathbb{E}_{q_\phi(z_{1:T} \mid x_{1:T}, y_{1:T})}
  \Big[
    \log p_\theta(y_{1:T}, z_{1:T} \mid x_{1:T})
    - \log q_\phi(z_{1:T} \mid x_{1:T}, y_{1:T})
  \Big],
  \label{eq:vrnn-jensen}
\end{align}
where the inequality is Jensen's inequality.
We now assume a mean-field factorization
$q_\phi(z_{1:T} \mid x_{1:T}, y_{1:T})
 = \prod_{t=1}^T q_\phi(z_t \mid x_t, y_t, h_{t-1})$ and plug in
\cref{eq:vrnn-joint}:
\begin{align}
  \mathbb{E}_{q_\phi}
  \big[\log p_\theta(y_{1:T}, z_{1:T} \mid x_{1:T})\big]
  &=
  \mathbb{E}_{q_\phi}
  \Big[
    \sum_{t=1}^T
    \big(
      \log p_\theta(z_t \mid h_{t-1})
      + \log p_\theta(y_t \mid x_t, z_t, h_t)
    \big)
  \Big]
  \nonumber\\
  &=
  \sum_{t=1}^T
  \mathbb{E}_{q_\phi(z_t \mid x_t, y_t, h_{t-1})}
  \big[
    \log p_\theta(y_t \mid x_t, z_t, h_t)
  \big]
  \nonumber\\
  &\quad
  + \sum_{t=1}^T
  \mathbb{E}_{q_\phi(z_t \mid x_t, y_t, h_{t-1})}
  \big[
    \log p_\theta(z_t \mid h_{t-1})
  \big],
  \label{eq:vrnn-exp-joint}
\end{align}
and similarly
\begin{align}
  \mathbb{E}_{q_\phi}
  \big[\log q_\phi(z_{1:T} \mid x_{1:T}, y_{1:T})\big]
  &=
  \sum_{t=1}^T
  \mathbb{E}_{q_\phi(z_t \mid x_t, y_t, h_{t-1})}
  \big[
    \log q_\phi(z_t \mid x_t, y_t, h_{t-1})
  \big].
  \label{eq:vrnn-exp-q}
\end{align}
Substituting \cref{eq:vrnn-exp-joint,eq:vrnn-exp-q} into
\cref{eq:vrnn-jensen} yields
\begin{align}
  \log p_\theta(y_{1:T} \mid x_{1:T})
  &\ge
  \sum_{t=1}^T
  \mathbb{E}_{q_\phi(z_t \mid x_t, y_t, h_{t-1})}
  \big[
    \log p_\theta(y_t \mid x_t, z_t, h_t)
  \big]
  \nonumber\\
  &\quad
  + \sum_{t=1}^T
  \mathbb{E}_{q_\phi(z_t \mid x_t, y_t, h_{t-1})}
  \big[
    \log p_\theta(z_t \mid h_{t-1})
    - \log q_\phi(z_t \mid x_t, y_t, h_{t-1})
  \big].
\end{align}
The second sum is $-\sum_t \mathrm{KL}(q_\phi(\cdot)\Vert p_\theta(\cdot))$,
giving the standard VRNN ELBO:
\begin{align}
  \mathcal{L}_{\text{VRNN}}
  &\;=\;
  \sum_{t=1}^T
  \mathbb{E}_{q_\phi(z_t \mid x_t, y_t, h_{t-1})}
  \Big[
    \log p_\theta\big(y_t \mid x_t, z_t, h_t\big)
  \Big]
  \;-\;
  \sum_{t=1}^T
  \mathrm{KL}\!\Big(
    q_\phi(z_t \mid x_t, y_t, h_{t-1})
    \,\Big\Vert\,
    p_\theta(z_t \mid h_{t-1})
  \Big).
  \label{eq:vrnn-elbo}
\end{align}
In the implementation, we use a $\beta$-weighted KL,
$\mathcal{L}_{\text{VRNN}}^\beta 
 = \sum_t \mathbb{E}[\log p_\theta(y_t \mid \cdot)]
   - \beta \sum_t \mathrm{KL}(\cdot\Vert\cdot)$,
and estimate each term using a single Monte Carlo sample from
$q_\phi$.

\subsection{DSSM baseline}

The DSSM baseline follows a deterministic+stochastic state-space model.
We maintain a deterministic state $h_t$ that summarizes the observed
pair $(x_t, y_t)$, and attach a stochastic latent $z_t$ that modulates
the emission.
Unlike VRNN, the latent does not enter the recurrent update.

\paragraph{Deterministic state.}
We define an encoder $\phi_e$ over the current pair $(x_t,y_t)$ and a GRU
cell:
\begin{equation}
  e_t = \phi_e([x_t, y_t]),
  \qquad
  h_t = f_\theta(h_{t-1}, e_t).
  \label{eq:dssm-h}
\end{equation}
We treat $h_t$ as a deterministic summary of all past observations
$(x_{1:t}, y_{1:t})$; it is not a random variable in the generative
model, but conditions both the prior and decoder.

\paragraph{Latent prior and likelihood.}
Conditioned on $h_{t-1}$, the prior over $z_t$ is again diagonal
Gaussian:
\begin{equation}
  p_\theta(z_t \mid h_{t-1})
  \;=\;
  \mathcal{N}\!\big(
    z_t;\,\mu^p_t,\,\mathrm{diag}((\sigma^p_t)^2)
  \big),
  \quad
  [\mu^p_t, \log(\sigma^p_t)^2]
  \;=\;
  \mathrm{prior}_\theta(h_{t-1}).
  \label{eq:dssm-prior}
\end{equation}
Given $x_t$, $z_t$, and $h_t$, the emission is
\begin{equation}
  p_\theta(y_t \mid x_t, z_t, h_t)
  \;=\;
  \mathcal{N}\!\big(
    y_t;\,\mu^y_t,\,\mathrm{diag}((\sigma^y_t)^2)
  \big),
  \quad
  [\mu^y_t, \log(\sigma^y_t)^2]
  \;=\;
  \mathrm{dec}_\theta\big([h_t, z_t, x_t]\big).
  \label{eq:dssm-dec}
\end{equation}

\paragraph{Variational posterior.}
The amortized posterior conditions on both the previous state and the
encoded current pair:
\begin{equation}
  q_\phi(z_t \mid x_t, y_t, h_{t-1})
  \;=\;
  \mathcal{N}\!\big(
    z_t;\,\mu^q_t,\,\mathrm{diag}((\sigma^q_t)^2)
  \big),
  \quad
  [\mu^q_t, \log(\sigma^q_t)^2]
  \;=\;
  \mathrm{post}_\phi\big([h_{t-1}, e_t]\big).
  \label{eq:dssm-post}
\end{equation}
Sampling again uses reparameterization.

\paragraph{Joint and ELBO.}
For DSSM, the same chain-rule argument as in
\cref{eq:vrnn-chain}--\cref{eq:vrnn-joint},
with the deterministic update $h_t = f_\theta(h_{t-1}, \phi_e([x_t,y_t]))$
in place of \cref{eq:vrnn-h}, yields the conditional joint
\begin{equation}
  p_\theta(y_{1:T}, z_{1:T} \mid x_{1:T})
  \;=\;
  \prod_{t=1}^T
  p_\theta(z_t \mid h_{t-1})\,
  p_\theta(y_t \mid x_t, z_t, h_t),
  \label{eq:dssm-joint}
\end{equation}
where $h_t$ is the deterministic state from \cref{eq:dssm-h}.
Repeating the variational derivation in
\cref{eq:vrnn-jensen}--\cref{eq:vrnn-elbo}
with the factorization
$q_\phi(z_{1:T} \mid x_{1:T}, y_{1:T})
 = \prod_t q_\phi(z_t \mid x_t, y_t, h_{t-1})$
and the definitions in
\cref{eq:dssm-prior,eq:dssm-dec,eq:dssm-post}
gives the DSSM ELBO
\begin{align}
  \mathcal{L}_{\text{DSSM}}
  &\;=\;
  \sum_{t=1}^T
  \mathbb{E}_{q_\phi(z_t \mid h_{t-1}, x_t, y_t)}
  \Big[
    \log p_\theta\big(y_t \mid x_t, z_t, h_t\big)
  \Big]
  \;-\;
  \sum_{t=1}^T
  \mathrm{KL}\!\Big(
    q_\phi(z_t \mid x_t, y_t, h_{t-1})
    \,\Big\Vert\,
    p_\theta(z_t \mid h_{t-1})
  \Big).
  \label{eq:dssm-elbo}
\end{align}
In practice, we again apply a $\beta$-weight on the KL and use a single
Monte Carlo sample from $q_\phi$ per step.

\subsection{Comparison}

Both baselines share the same conditional prior family
$p_\theta(z_t \mid h_{t-1})$ and Gaussian decoder
$p_\theta(y_t \mid x_t, z_t, h_t)$, and are trained with nearly identical
one-step ELBOs.
The key structural difference lies in the deterministic recurrence:
\begin{itemize}
  \item \textbf{VRNN} uses the latent in the core dynamics
  (\cref{eq:vrnn-h}):
  $h_t$ depends on $z_t$ (via $\phi_z(z_t)$), so information about
  past latents propagates forward through the RNN state.
  \item \textbf{DSSM} keeps the recurrent state purely
  observation-driven (\cref{eq:dssm-h}):
  $h_t$ depends on the encoded pair $e_t = \phi_e([x_t,y_t])$ but not
  on $z_t$; the latent $z_t$ only modulates the emission
  (\cref{eq:dssm-dec}).
\end{itemize}
This yields two complementary latent sequence baselines under a
matched ELBO objective:
VRNN represents a ``latent-in-the-dynamics'' recurrent model,
while DSSM represents a deterministic RNN with an auxiliary latent head
on top of its observation-driven state.

\newpage

\section{Tables and Figures}
\label{app:table_and_figs}

\begin{table}[H]
\centering
\caption{Training protocol, capacity constraints, and optimization settings used across all experiments.}
\label{tab:training_protocol}
\begin{tabular}{l c}
\toprule
\textbf{Setting} & \textbf{Value} \\
\midrule
Dataset & Jena Climate \\
Train / evaluation split & 80\% / 20\% (chronological) \\
Epochs & 6 \\
Random seeds & 25 \\
\midrule
Stateful models & LT, VRNN, DSSM \\
Training scheme (stateful) & TBPTT (no overlap) \\
TBPTT window length $W$ & 256 \\
Recency weighting & Fixed, $\lambda = 0.9$ \\
Latent dimension & $\dim(z_t) = 8$ \\
Hidden state dimension & $\dim(h_t) = 64$ \\
\midrule
Static models & MC-Dropout, BBB, Deep Ensembles \\
Training scheme (static) & Sliding window \\
Window length / stride & $W = 256$, $S = 32$ \\
Relative updates per epoch & $\approx 8\times$ stateful models \\
\midrule
KL annealing schedule & Linear \\
Maximum $\beta$ & $\beta_{\max} = 1$ \\
Warmup (stateful) & $\beta_{\text{warmup}} = 575$ updates \\
Warmup (static) & $\beta_{\text{warmup}} = 4600$ updates \\
\midrule
Generated weights (LT) & Entire predictor network \\
\midrule
Inference-time parameter budget & $20$k parameters ($\pm 5\%$) \\
Total trainable parameters & $20$k parameters ($\pm 10\%$) \\
\midrule
Inference-time sampling & $K = 100$ MC samples \\
Training-time sampling & $K = 1$ MC sample \\
Deep Ensemble size & $M = 10$ members \\
\midrule
Optimizer & Adam \\
Learning rate & $1 \times 10^{-4}$ \\
Gradient clipping & $\ell_2$ norm clipped at $1.0$ \\
\bottomrule
\end{tabular}
\end{table}

\begin{table}[H]
\centering
\caption{Model parameter counts. \emph{Total Parameters} includes all trainable parameters used
during training, while \emph{Inference-Time Parameters} reports the number of parameters active at
deployment.}
\label{tab:param_counts}
\begin{tabular}{l r r}
\toprule
\textbf{Model} & \textbf{Total Parameters} & \textbf{Inference-Time Parameters} \\
\midrule
Bayes-by-Backprop (BBB)     & 20{,}060 & 20{,}060 \\
DSSM                       & 22{,}262 & 20{,}630 \\
Deep Ensembles              & 19{,}100 & 19{,}100 \\
LT (Structured)    & 20{,}709 & 20{,}709 \\
LT (Unstructured)  & 21{,}888 & 20{,}848 \\
MC-Dropout                  & 19{,}112 & 19{,}112 \\
VRNN                        & 20{,}758 & 20{,}758 \\
\bottomrule
\end{tabular}
\end{table}


\begin{figure}[t]
\centering

\subfigure[Predictive mean vs t]{
    \includegraphics[width=\linewidth]{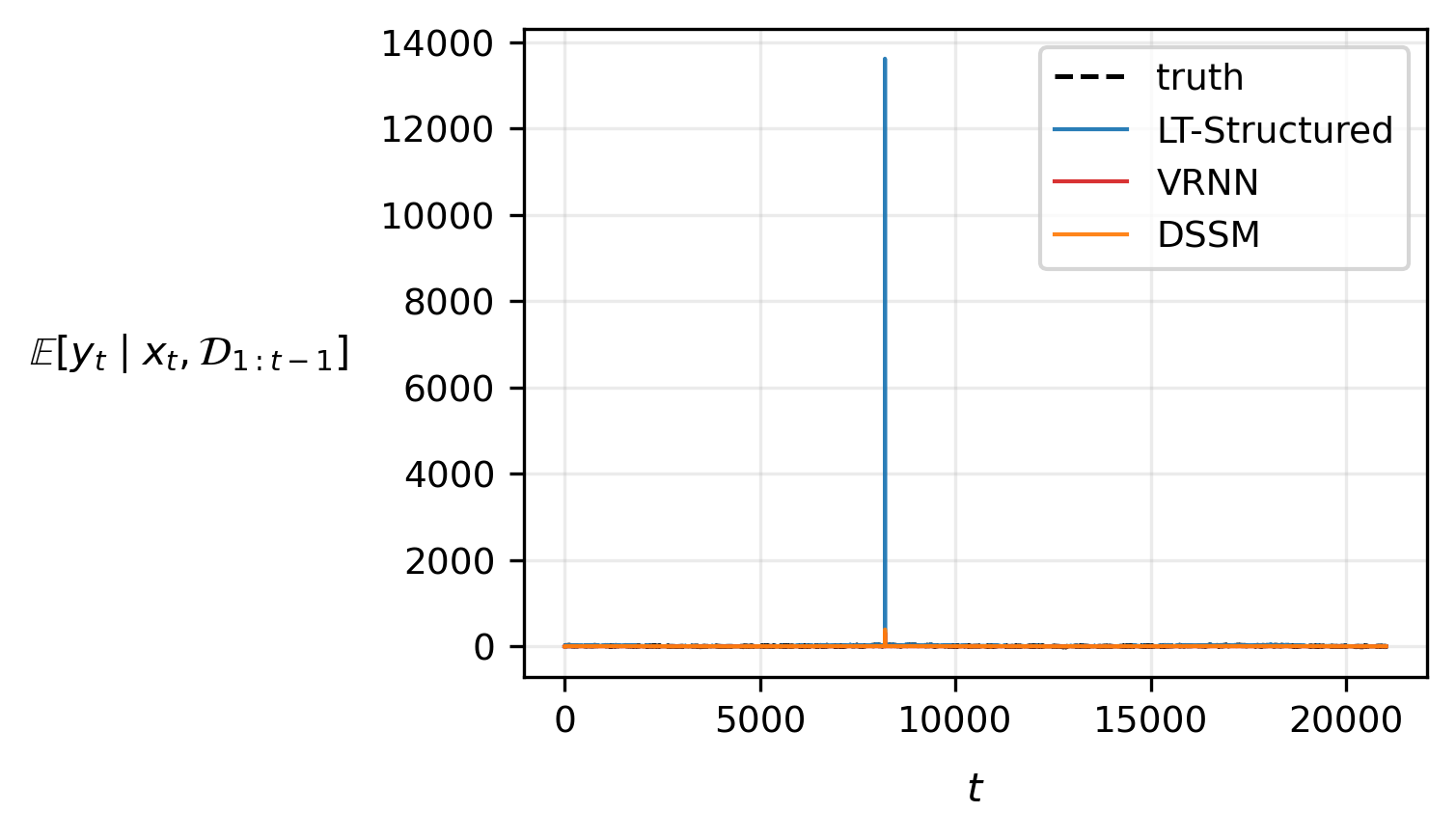}
}

\vspace{4mm}

\subfigure[Predictive mean vs t (clipped)]{
    \includegraphics[width=\linewidth]{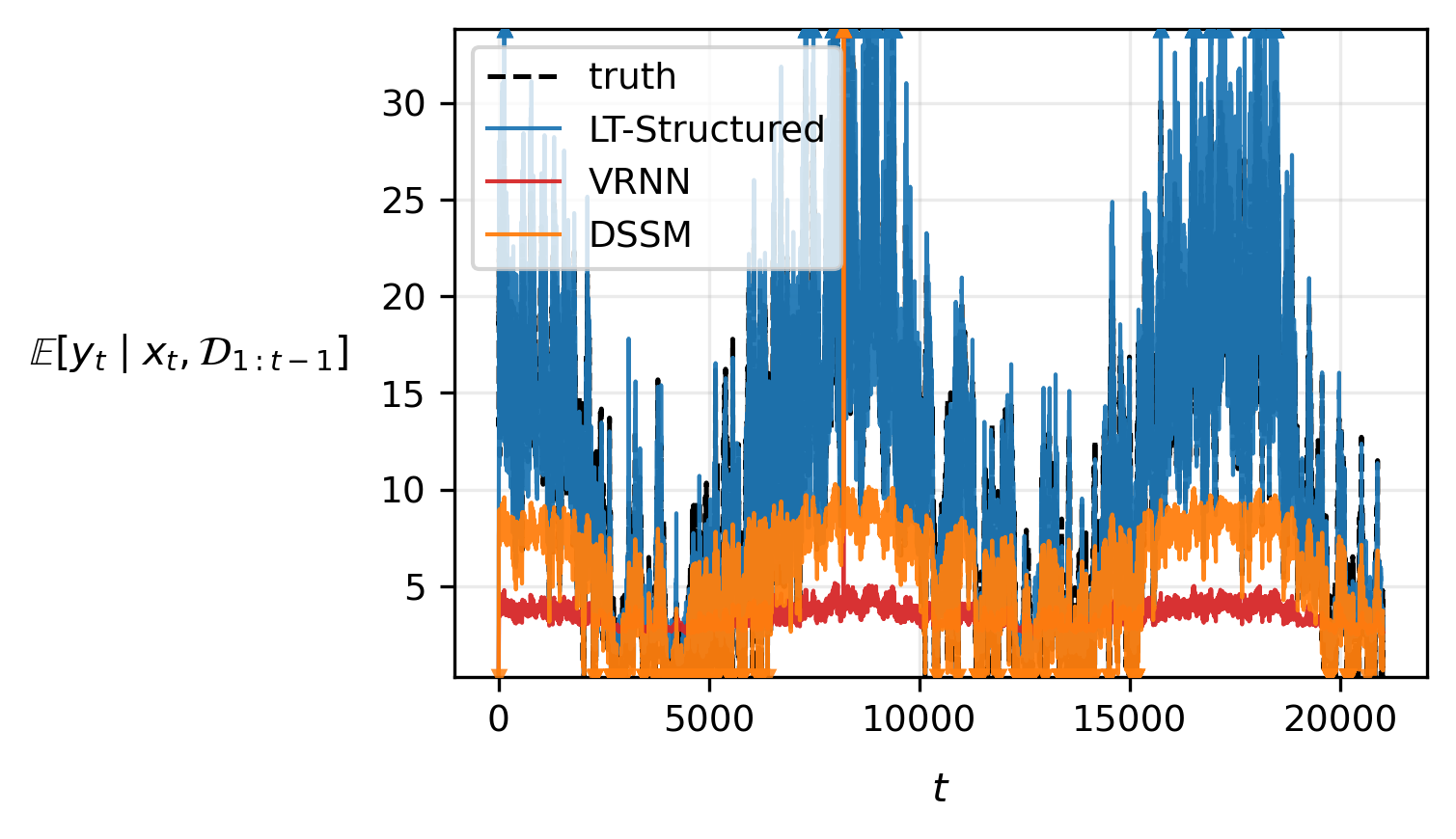}
}

\caption{
Predictive mean trajectories for a representative seed.
\textbf{(a)} Unclipped predictive mean $\mathbb{E}[y_t \mid x_t, \mathcal{D}_{1:t-1}]$ over the full time horizon, illustrating a single isolated divergence event for LT-Structured.
\textbf{(b)} Clipped view of the same trajectories to highlight typical predictive behavior across methods.
The isolated spike in (a) corresponds to a rare catastrophic failure and dominates untrimmed temporal mean statistics, while the clipped view in (b) shows that LT-Structured otherwise tracks the signal accurately over time.
}
\label{fig:full_mean_vs_time}

\end{figure}


\begin{figure}[t]
\centering

\subfigure[$\sigma_{Aleatoric}$ vs $t$]{
    \includegraphics[width=\linewidth]{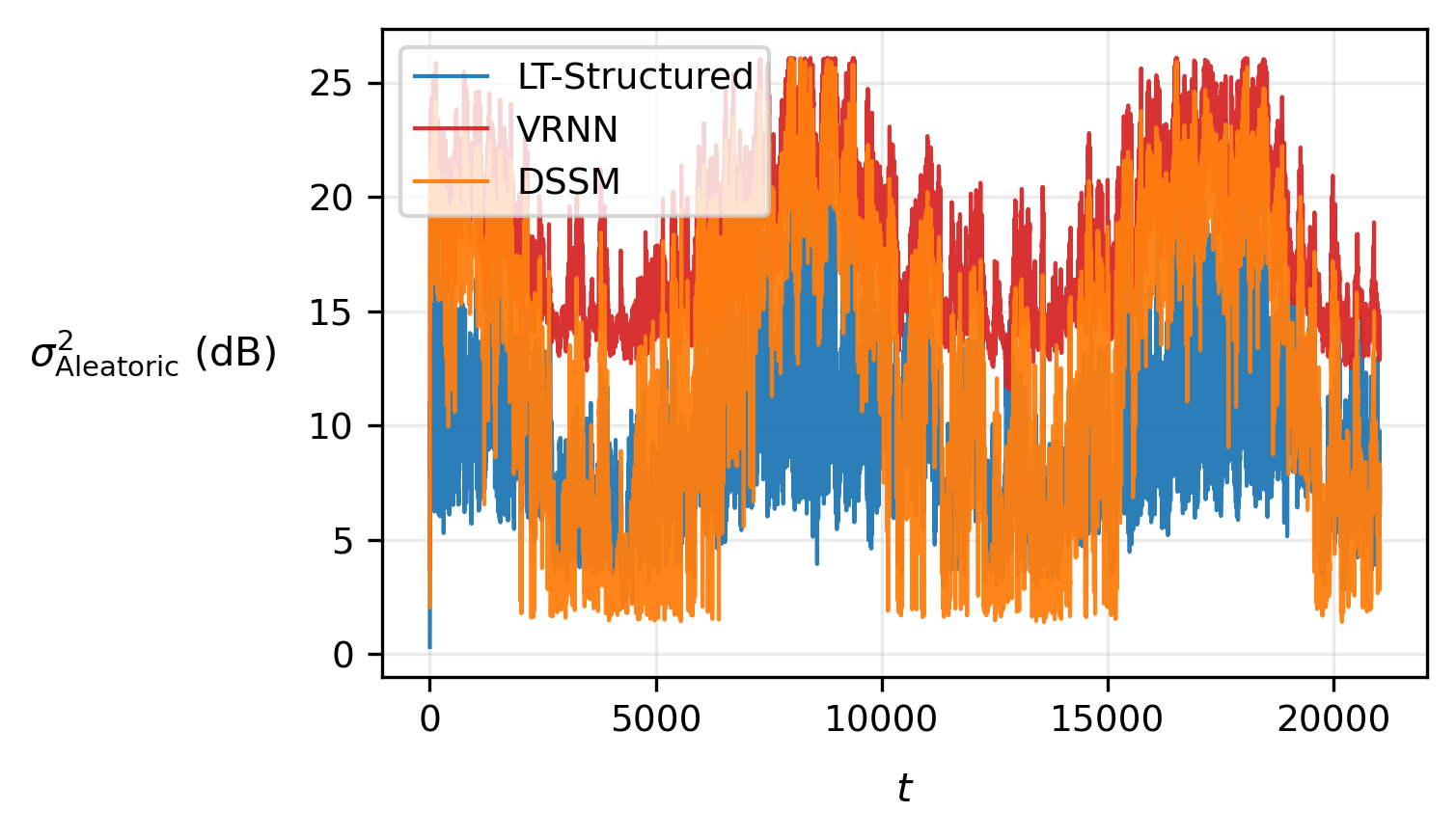}
}

\vspace{4mm}

\subfigure[$\sigma_{Epistemic}$ vs $t$]{
    \includegraphics[width=\linewidth]{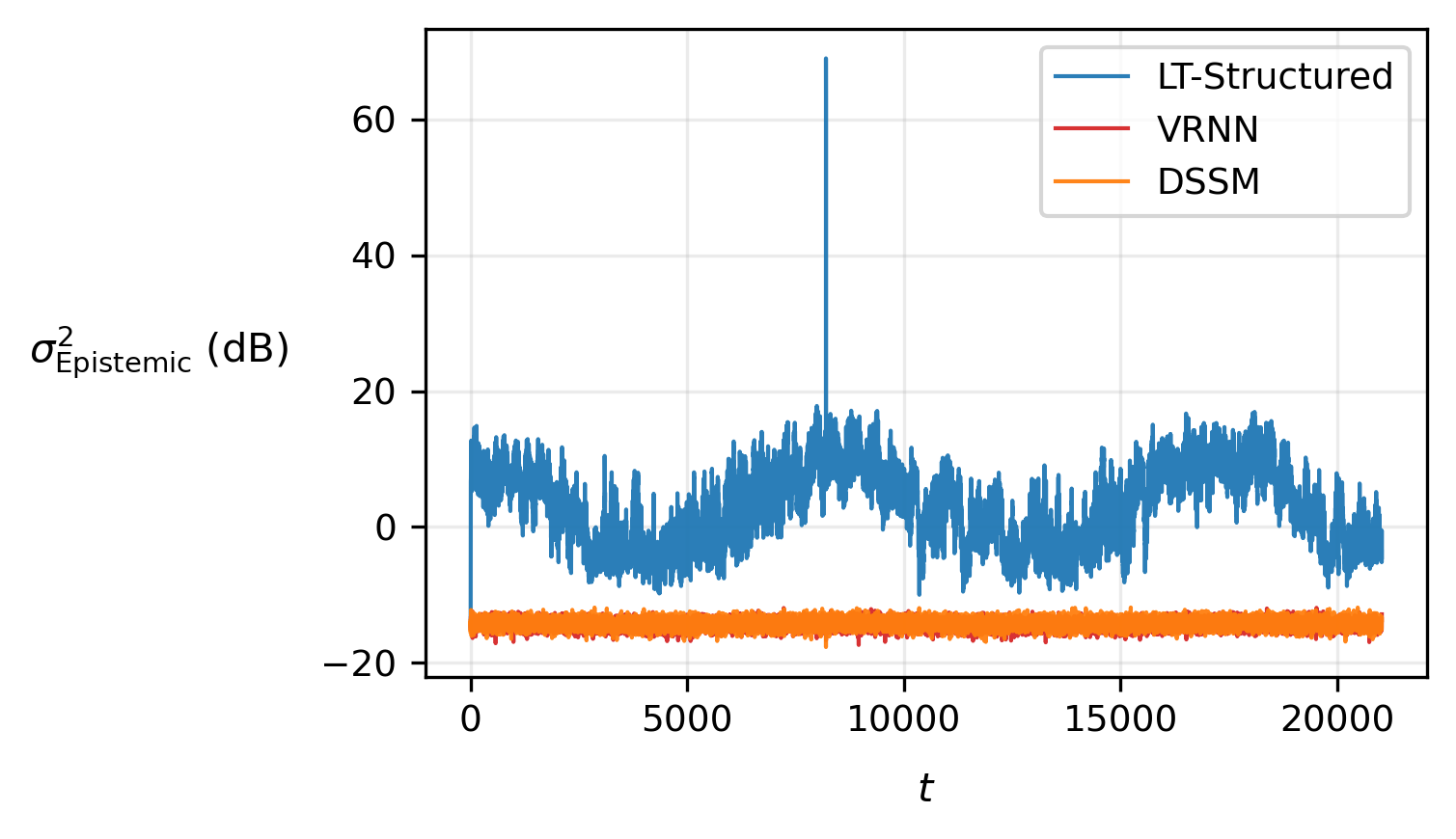}
}

\caption{
Decomposition of predictive uncertainty for the representative seed, reported in decibels (dB).
\textbf{(a)} Aleatoric variance $\sigma^2_{\mathrm{Aleatoric}}$ over time (dB), reflecting observation noise captured by each model.
\textbf{(b)} Epistemic variance $\sigma^2_{\mathrm{Epistemic}}$ over time (dB), reflecting model uncertainty estimated via predictive sampling.
LT-Structured exhibits temporally varying epistemic uncertainty with a single isolated spike corresponding to a rare divergence event, while VRNN and DSSM exhibit comparatively low and nearly constant epistemic variance.
}
\label{fig:full_var_vs_time}

\end{figure}


\begin{figure}[t]
\centering

\subfigure[$\sigma_{Total}$ vs $t$]{
    \includegraphics[width=\linewidth]{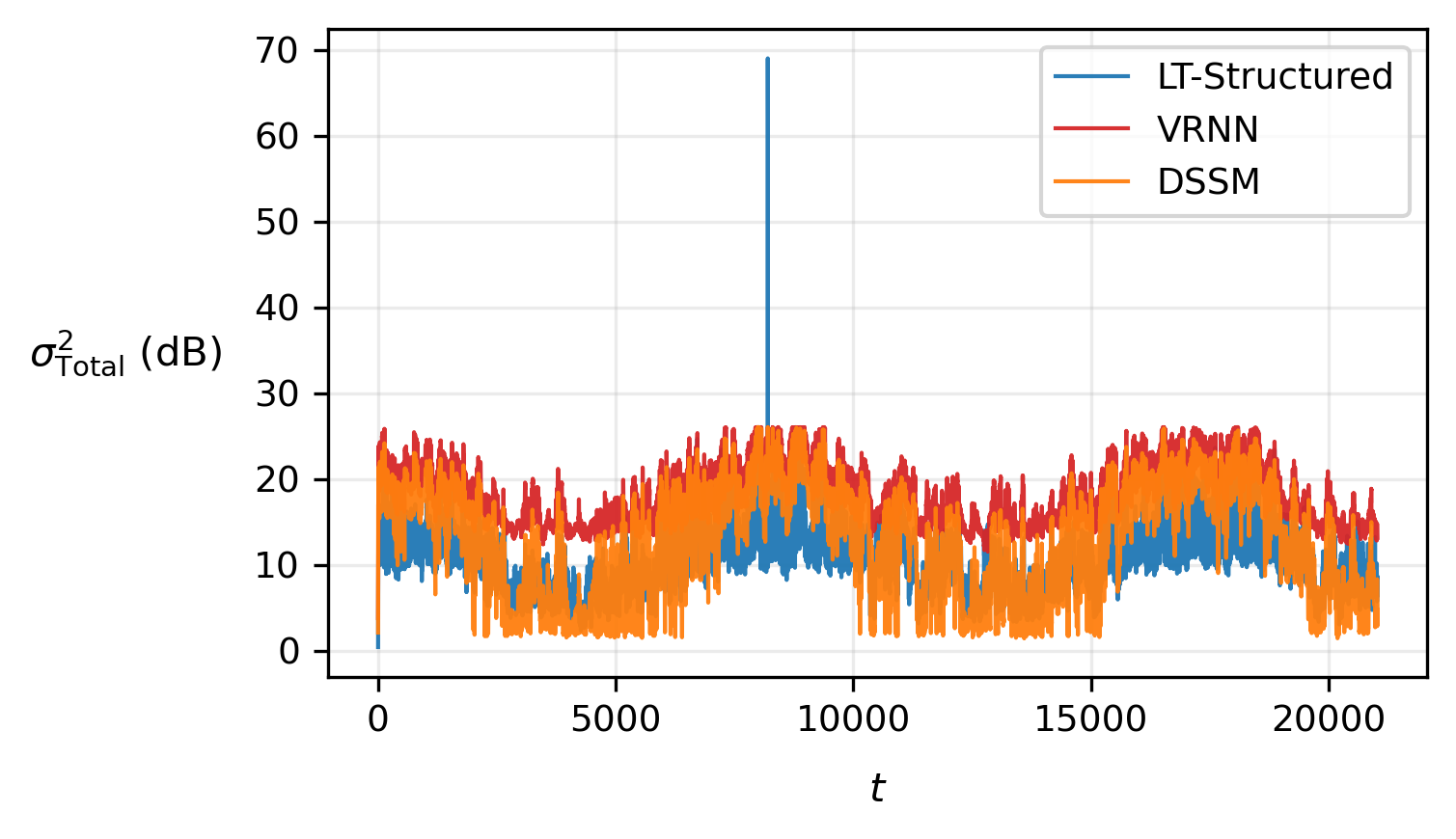}
}

\vspace{4mm}

\subfigure[$MSE$ vs $t$]{
    \includegraphics[width=\linewidth]{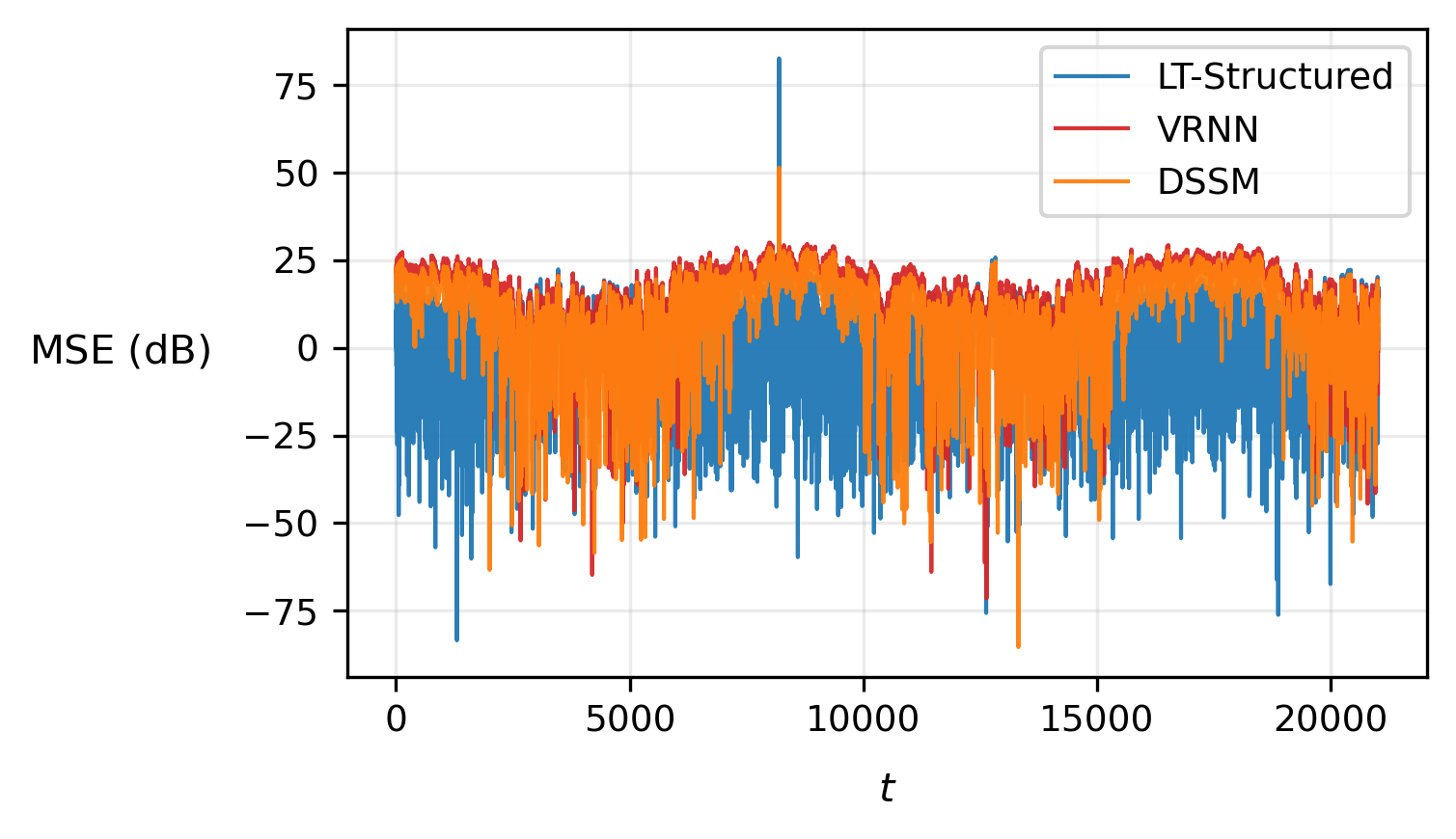}
}

\caption{
Total predictive variance and squared error over time for the representative seed, reported in decibels (dB).
\textbf{(a)} Total predictive variance $\sigma^2_{\mathrm{Total}} = \sigma^2_{\mathrm{Aleatoric}} + \sigma^2_{\mathrm{Epistemic}}$ as a function of time (dB).
\textbf{(b)} Corresponding mean squared error (MSE) over time (dB).
LT-Structured exhibits a single isolated spike in both variance and error corresponding to a rare divergence event, while otherwise maintaining stable uncertainty estimates and low error throughout the sequence.
VRNN and DSSM exhibit smoother but consistently higher error profiles over time.
}

\label{fig:full_mse_and_var_vs_time}

\end{figure}


\begin{figure}[t]
\centering

\subfigure[$NLL$ vs $t$]{
    \includegraphics[width=\linewidth]{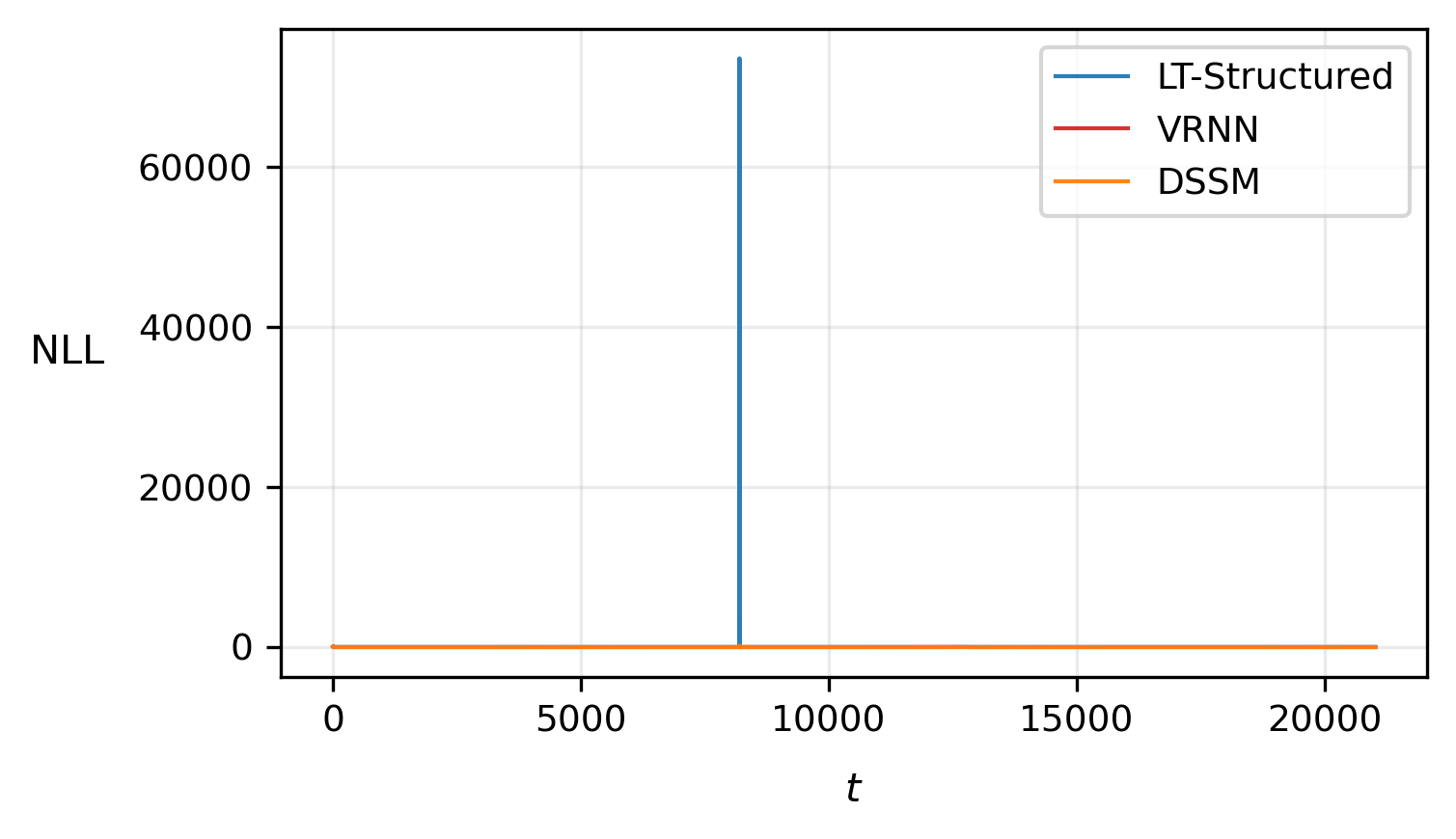}
}

\vspace{4mm}

\subfigure[$NLL$ vs $t$ (clipped)]{
    \includegraphics[width=\linewidth]{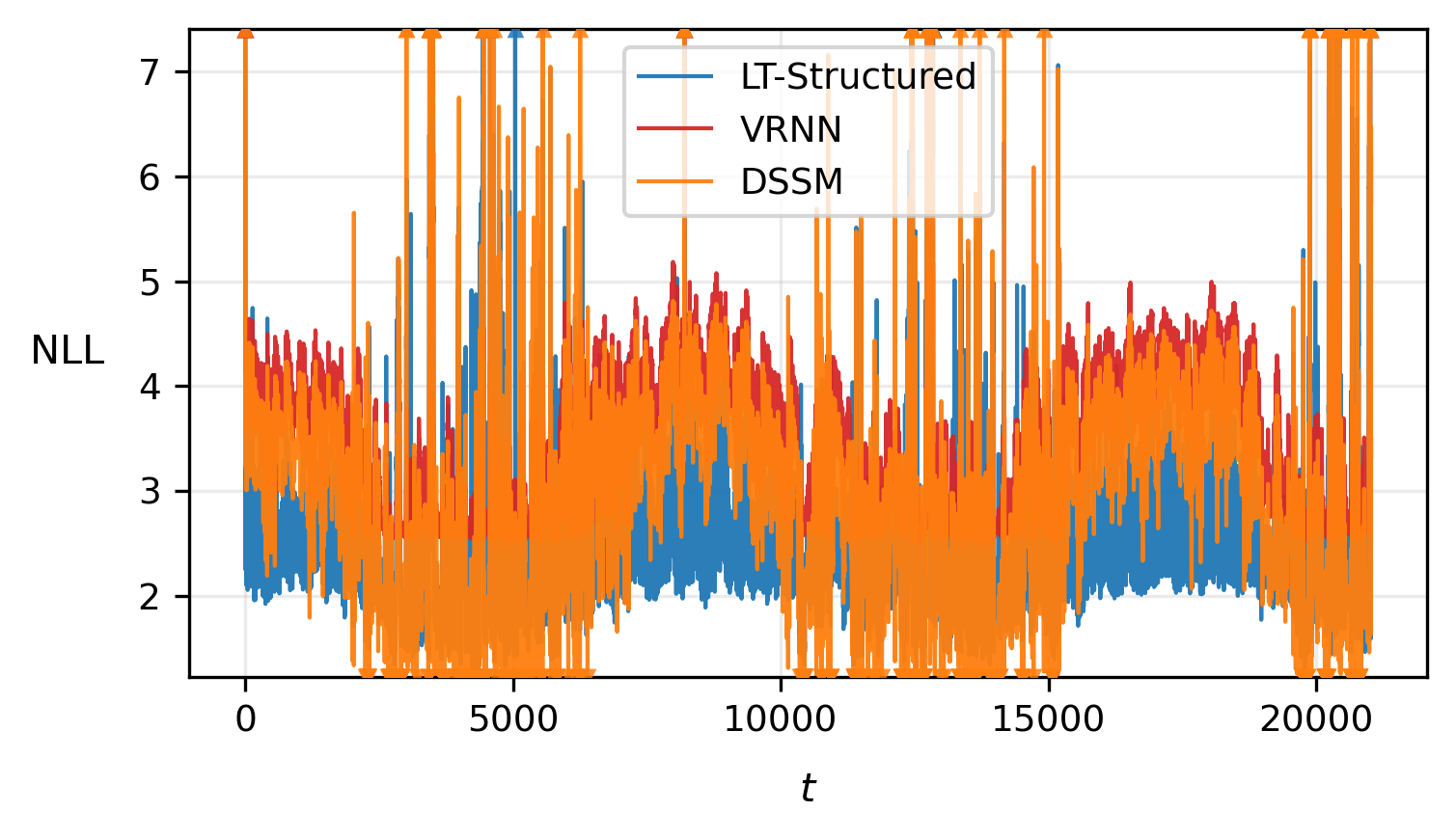}
}

\caption{
Negative log-likelihood (NLL) trajectories over time for the representative seed.
\textbf{(a)} Unclipped NLL as a function of time, illustrating a single isolated divergence event for LT-Structured that produces an extreme NLL spike.
\textbf{(b)} Clipped view of the same trajectories to highlight typical per-step likelihood behavior across methods.
The isolated spike in (a) dominates untrimmed temporal NLL statistics, while the clipped view in (b) shows that LT-Structured otherwise maintains consistently lower NLL over time relative to VRNN and DSSM.
}
\label{fig:full_nll_vs_time}

\end{figure}



\begin{figure}[t]
\centering

\subfigure[Predictive mean vs t]{
    \includegraphics[width=\linewidth]{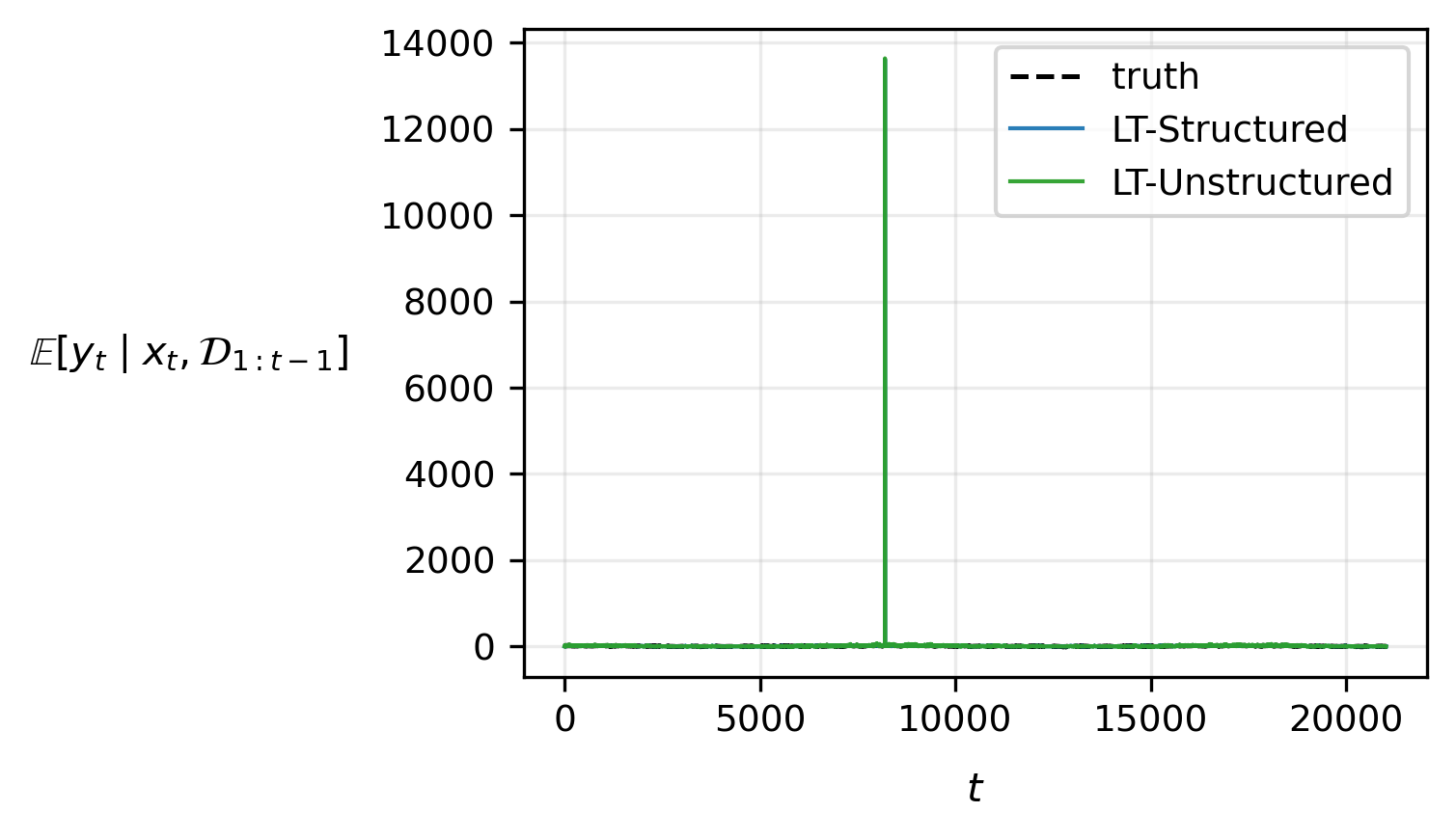}
}

\vspace{4mm}

\subfigure[Predictive mean vs t (clipped)]{
    \includegraphics[width=\linewidth]{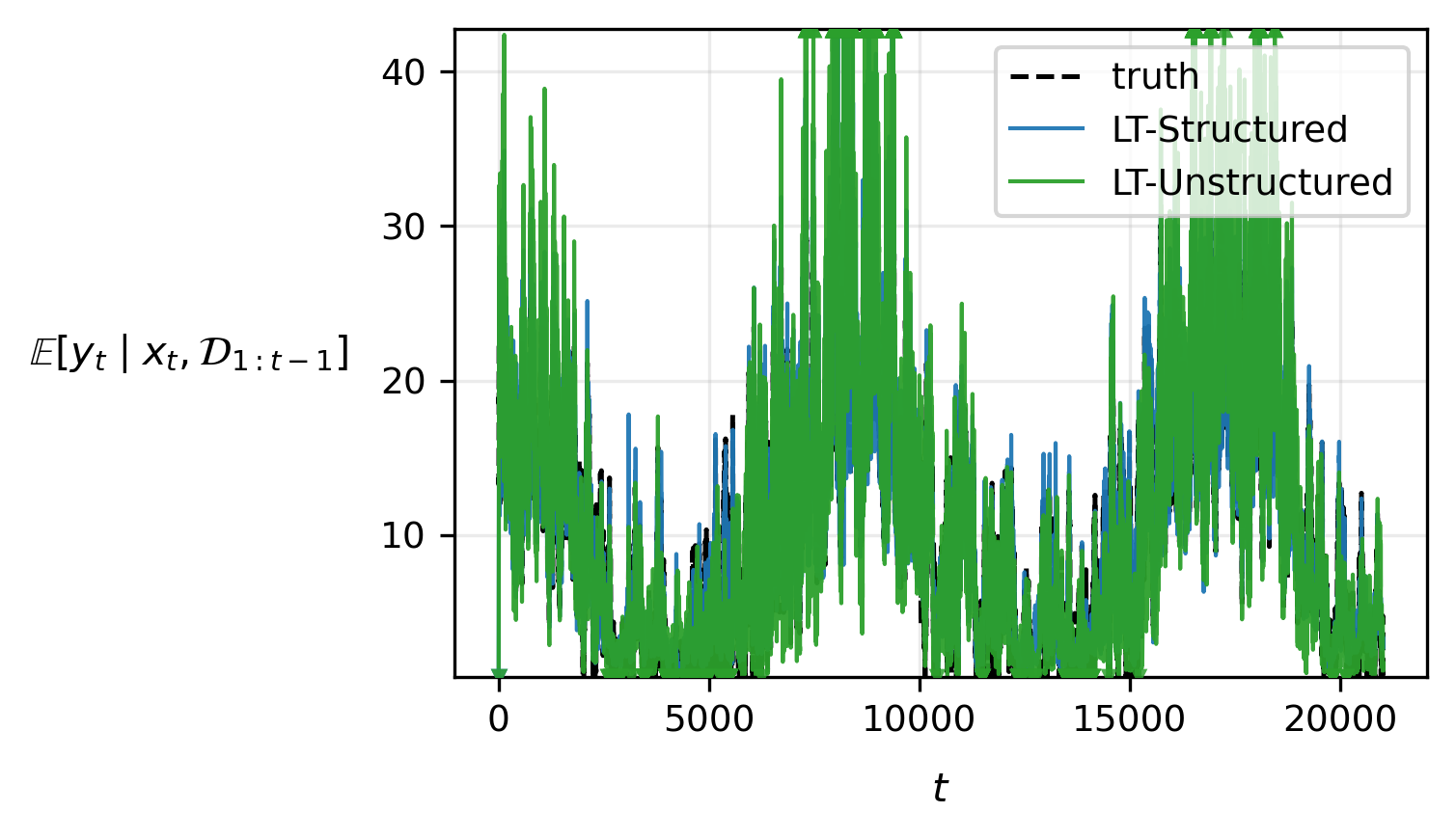}
}

\caption{
Predictive mean trajectories comparing LT variants for the representative seed.
\textbf{(a)} Unclipped predictive mean $\mathbb{E}[y_t \mid x_t, \mathcal{D}_{1:t-1}]$ over the full time horizon, illustrating an isolated divergence event for the unstructured variant.
\textbf{(b)} Clipped view of the same trajectories to highlight typical predictive behavior.
LT-Structured exhibits more stable tracking over time, while LT-Unstructured is more prone to large transient deviations; nevertheless, both variants maintain lower typical error than stateful baseline models under comparable clipping.
}
\label{fig:lt_comp_mean_vs_time}

\end{figure}


\begin{figure}[t]
\centering

\subfigure[$\sigma_{Aleatoric}$ vs $t$]{
    \includegraphics[width=\linewidth]{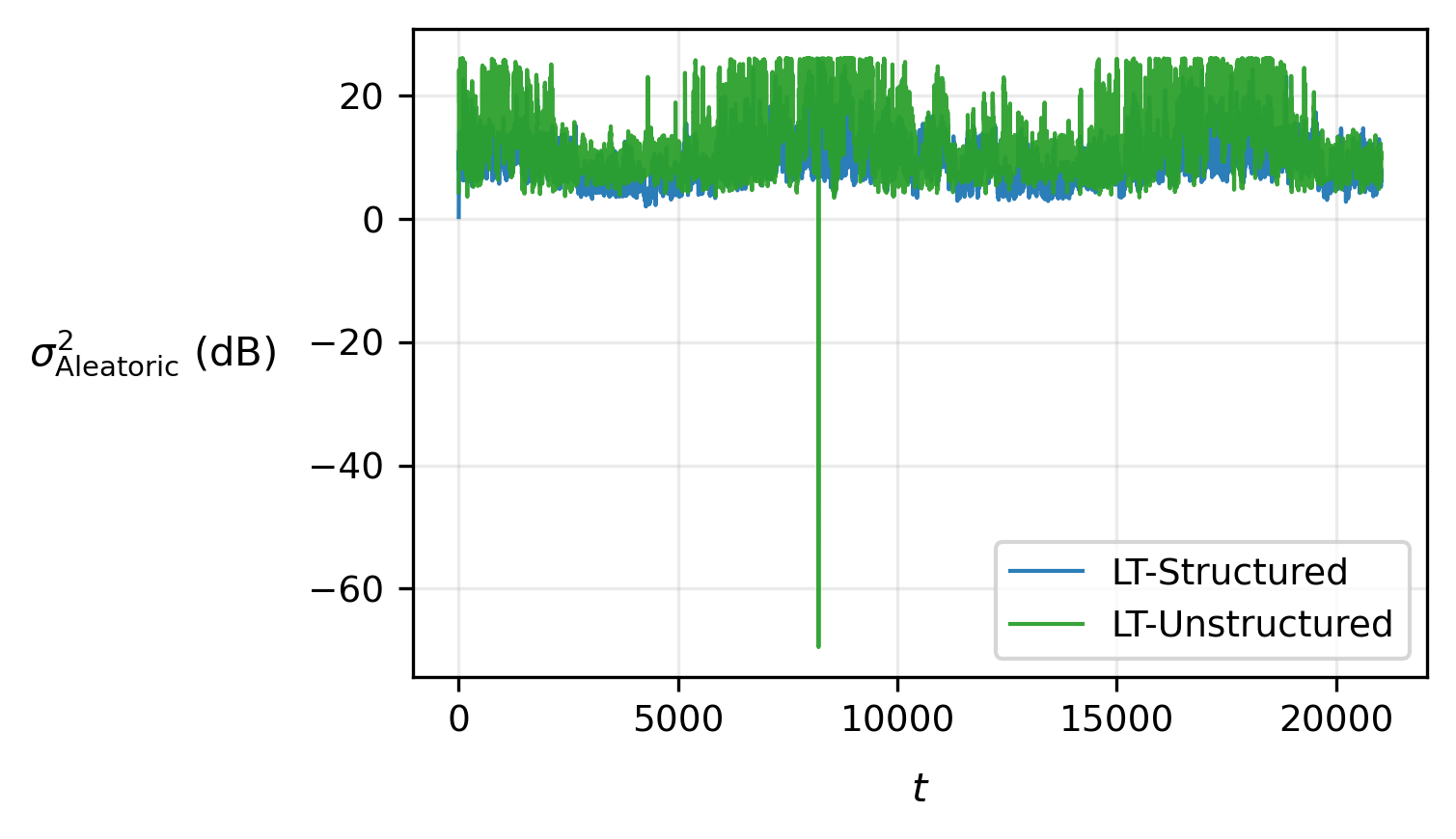}
}

\vspace{4mm}

\subfigure[$\sigma_{Epistemic}$ vs $t$]{
    \includegraphics[width=\linewidth]{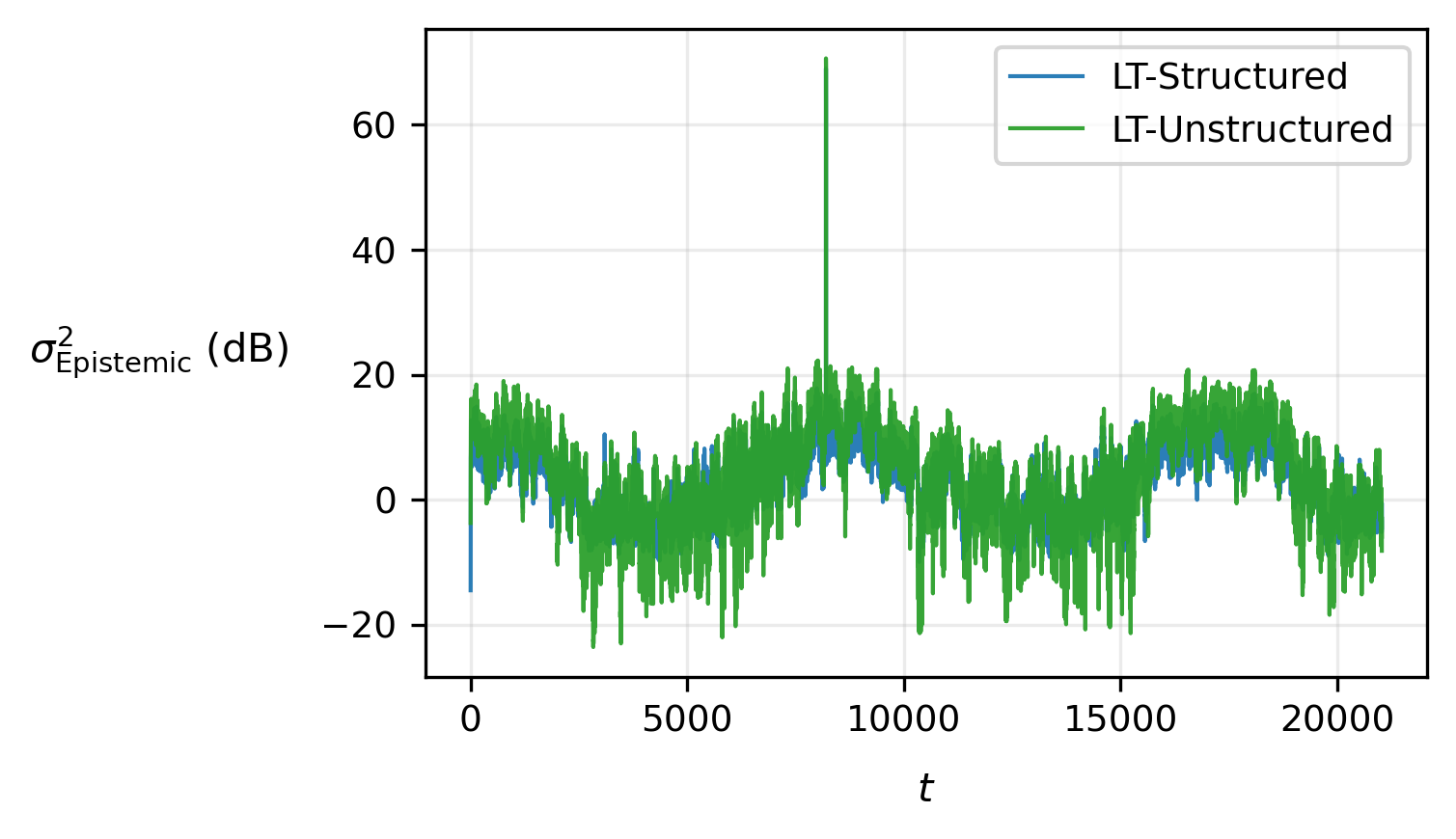}
}

\caption{
Decomposition of predictive uncertainty for LT variants on the representative seed, reported in decibels (dB).
\textbf{(a)} Aleatoric variance $\sigma^2_{\mathrm{Aleatoric}}$ over time (dB), reflecting observation noise captured by each model.
\textbf{(b)} Epistemic variance $\sigma^2_{\mathrm{Epistemic}}$ over time (dB), reflecting model uncertainty estimated via predictive sampling.
LT-Structured exhibits smoother and more stable uncertainty estimates over time, while LT-Unstructured shows larger transient fluctuations and an isolated spike corresponding to a rare divergence event.
}
\label{fig:lt_comp_var_vs_time}

\end{figure}


\begin{figure}[t]
\centering

\subfigure[$\sigma_{Total}$ vs $t$]{
    \includegraphics[width=\linewidth]{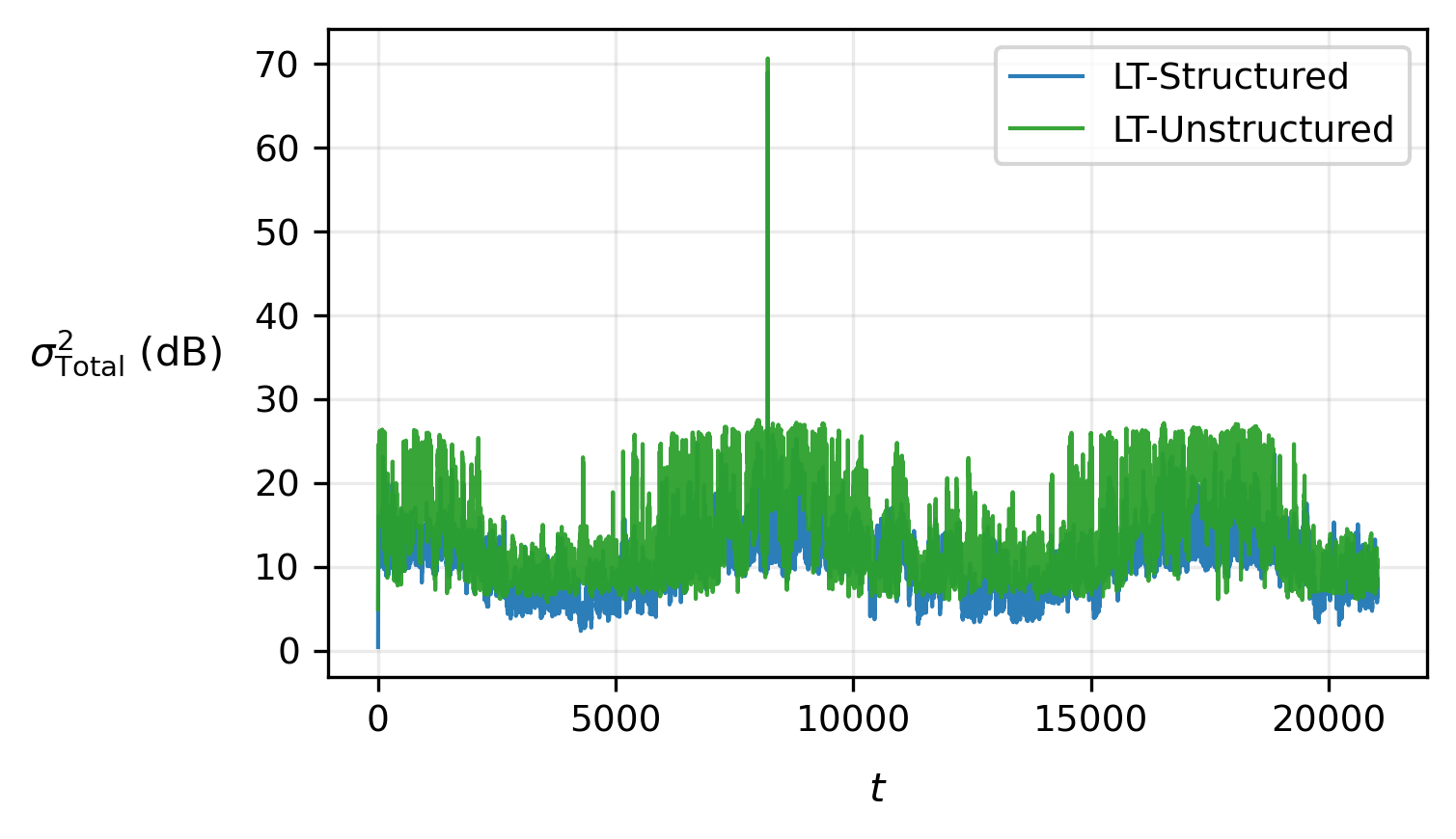}
}

\vspace{4mm}

\subfigure[$MSE$ vs $t$]{
    \includegraphics[width=\linewidth]{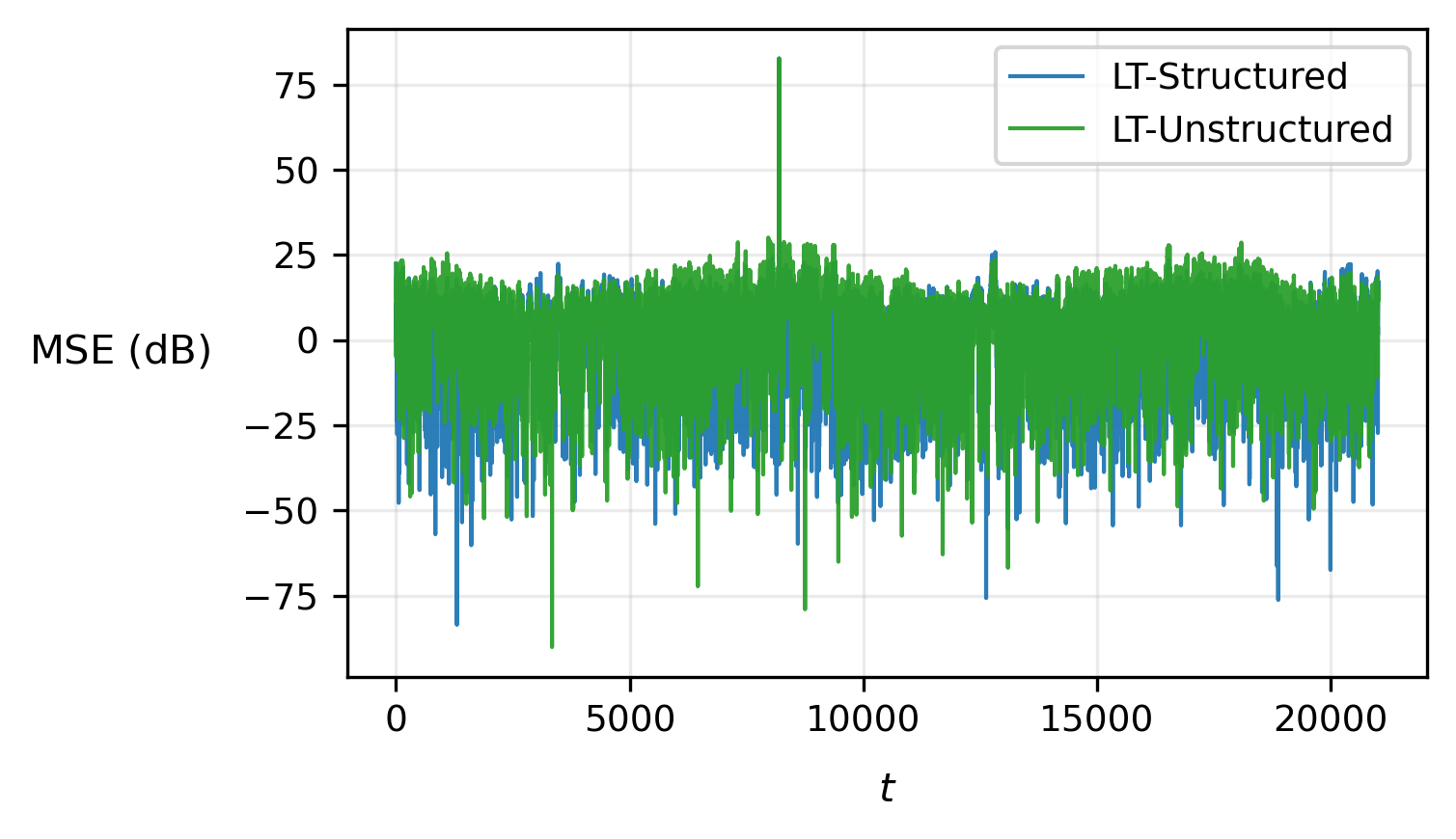}
}

\caption{
Total predictive variance and squared error over time for LT variants on the representative seed, reported in decibels (dB).
\textbf{(a)} Total predictive variance $\sigma^2_{\mathrm{Total}} = \sigma^2_{\mathrm{Aleatoric}} + \sigma^2_{\mathrm{Epistemic}}$ as a function of time (dB).
\textbf{(b)} Corresponding mean squared error (MSE) over time (dB).
LT-Unstructured exhibits a larger transient spike in both variance and error corresponding to a rare divergence event, while LT-Structured maintains more stable uncertainty estimates and lower typical error throughout the sequence.
}
\label{fig:lt_comp_mse_and_var_vs_time}

\end{figure}


\begin{figure}[t]
\centering

\subfigure[$NLL$ vs $t$]{
    \includegraphics[width=\linewidth]{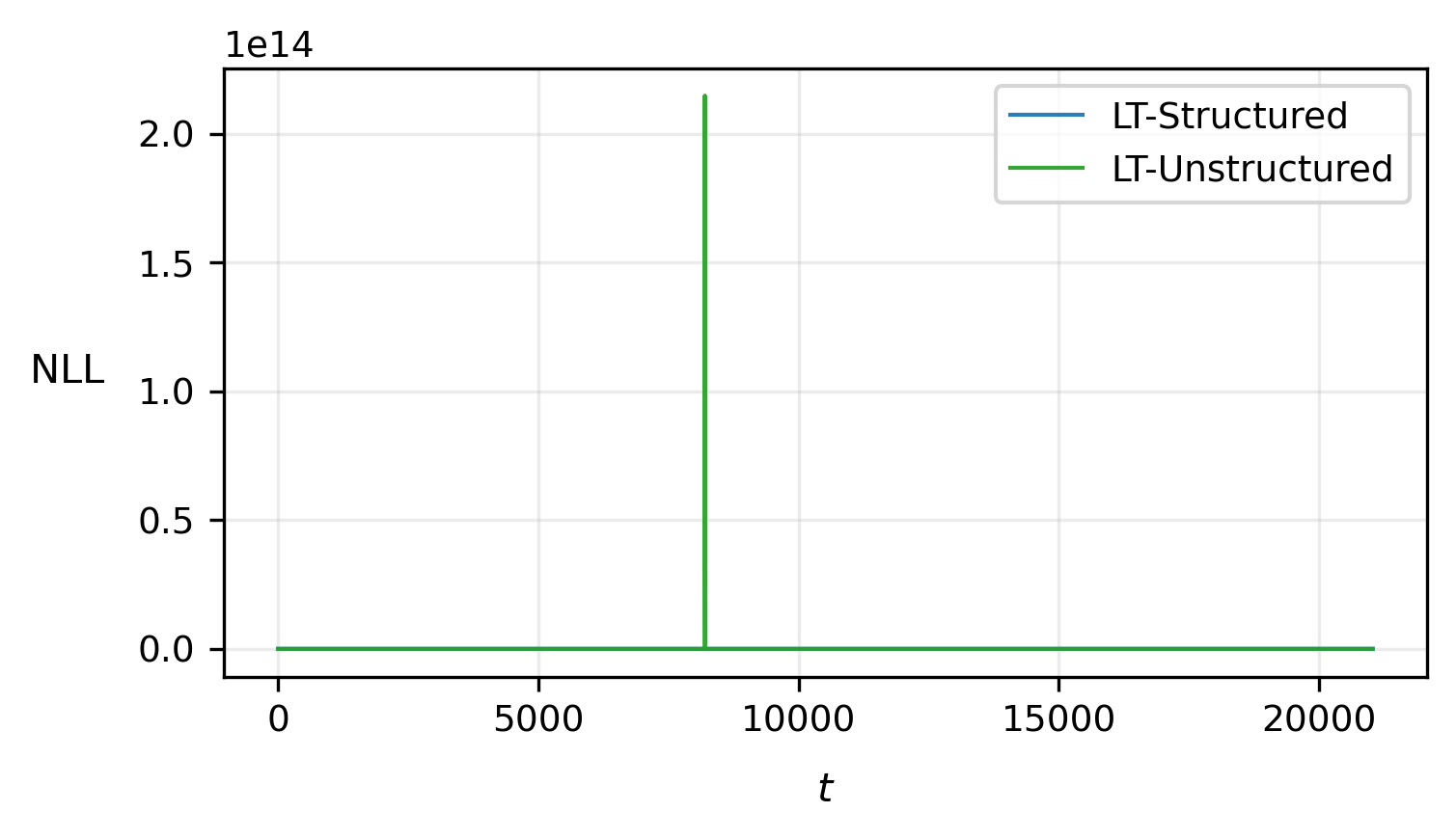}
}

\vspace{4mm}

\subfigure[$NLL$ vs $t$ (clipped)]{
    \includegraphics[width=\linewidth]{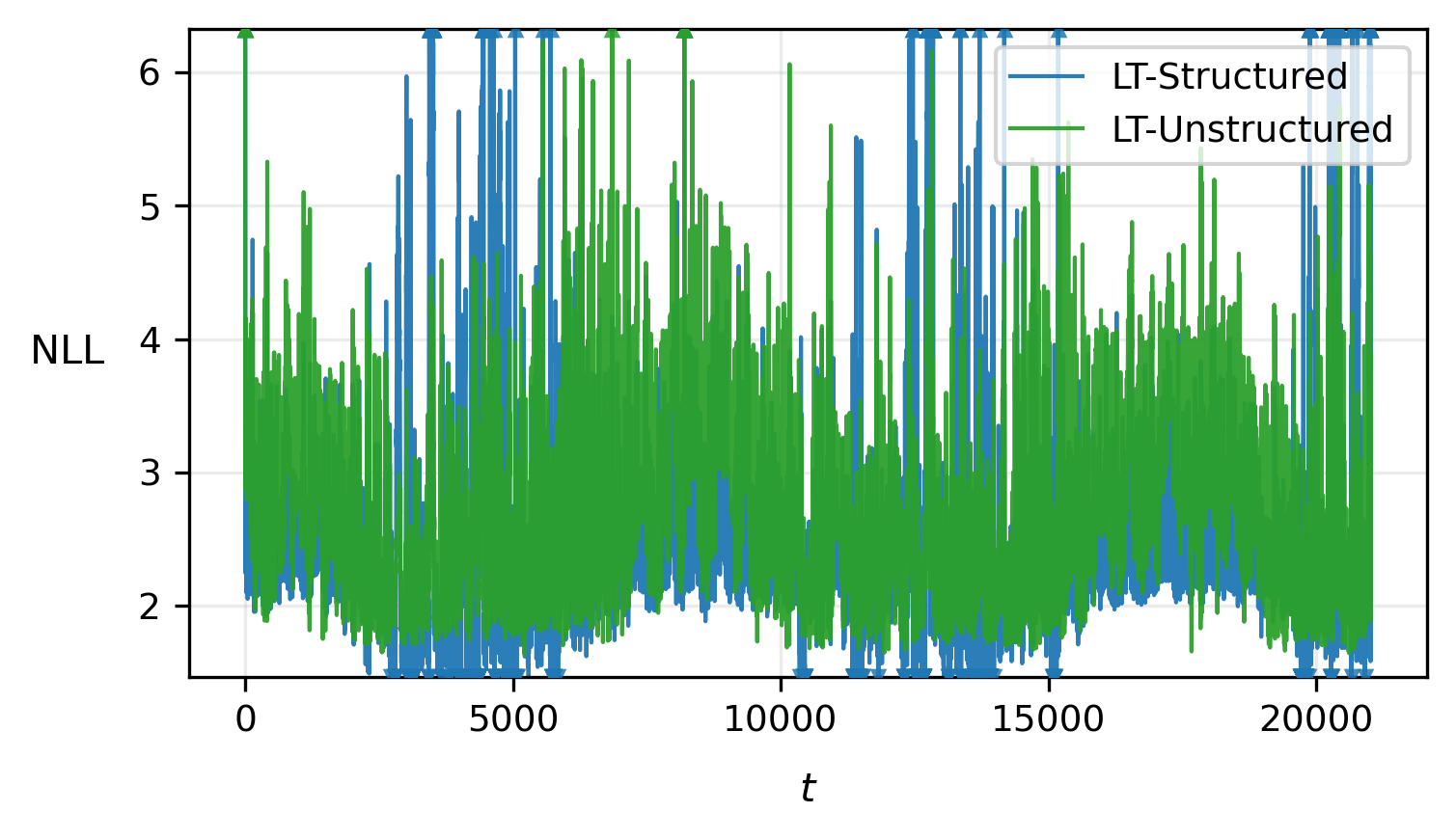}
}

\caption{
Negative log-likelihood (NLL) trajectories over time comparing LT variants for the representative seed.
\textbf{(a)} Unclipped NLL as a function of time, illustrating a single isolated divergence event for the unstructured variant that produces an extreme NLL spike.
\textbf{(b)} Clipped view of the same trajectories to highlight typical per-step likelihood behavior.
LT-Structured maintains more stable and consistently lower typical NLL over time, while LT-Unstructured is more prone to large transient deviations despite similar average behavior under clipping.
}
\label{fig:lt_comp_nll_vs_time}

\end{figure}


\begin{figure}[t]
  \centering
  \includegraphics[width=\linewidth]{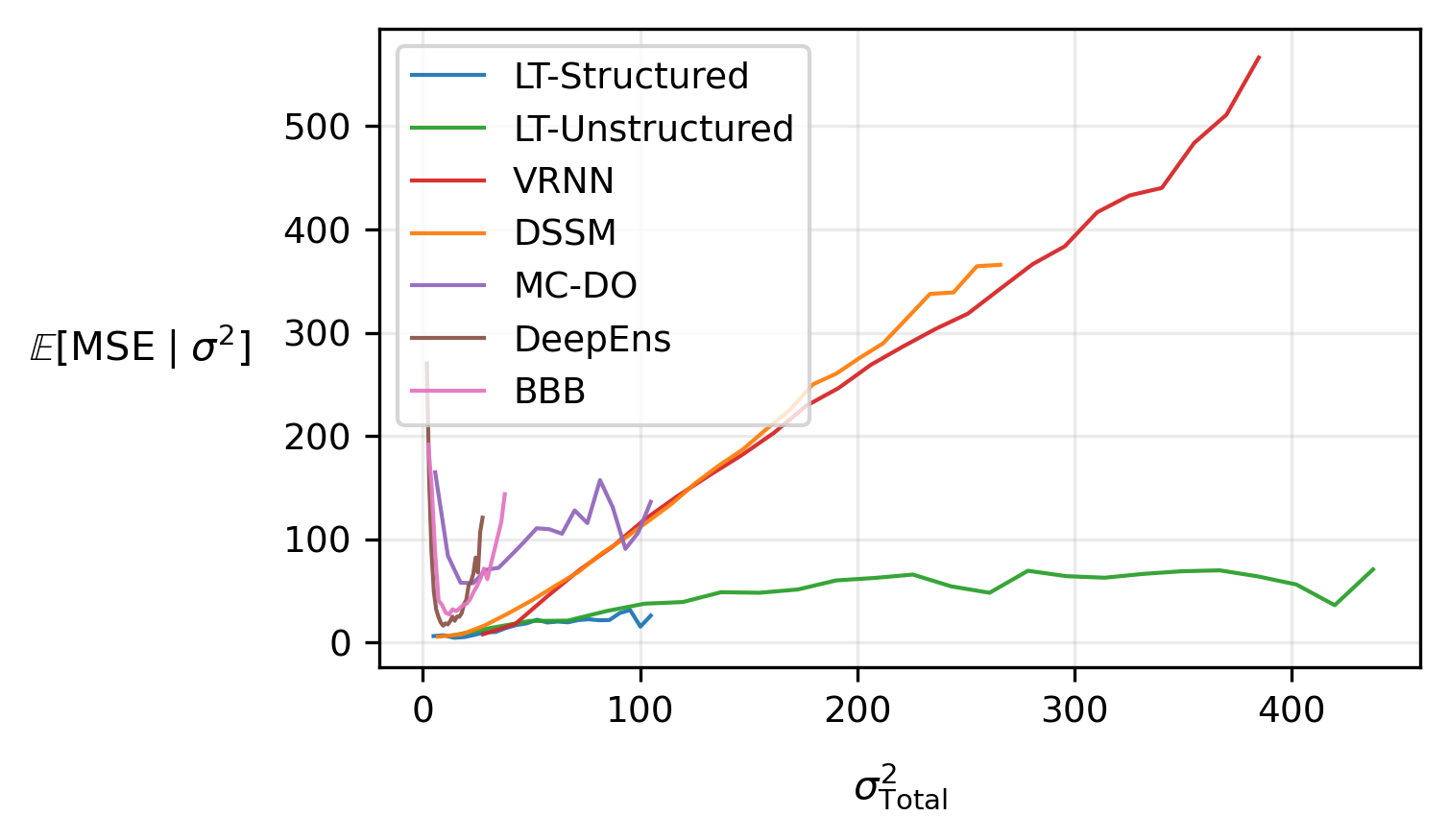}
    \caption{
    Empirical relationship between predictive uncertainty and squared error for the representative seed.
    The plot shows $\mathbb{E}[\mathrm{MSE} \mid \sigma^2_{\mathrm{total}}]$ as a function of total predictive variance $\sigma^2_{\mathrm{total}}$, both reported in linear scale, estimated by binning time steps according to predicted variance.
    To emphasize typical behavior, values are clipped to the $[1,99]$ percentile range prior to binning.
    Well-calibrated uncertainty corresponds to a monotonic and approximately proportional relationship between predicted variance and observed error.
    LT-Structured exhibits a comparatively shallow and stable error--variance relationship across uncertainty levels, whereas DSSM and VRNN display substantially steeper growth, indicative of poorly scaled uncertainty under high variance.
    }
  \label{fig:calib_curve}
\end{figure}

\begin{figure}[t]
  \centering
  \includegraphics[width=\linewidth]{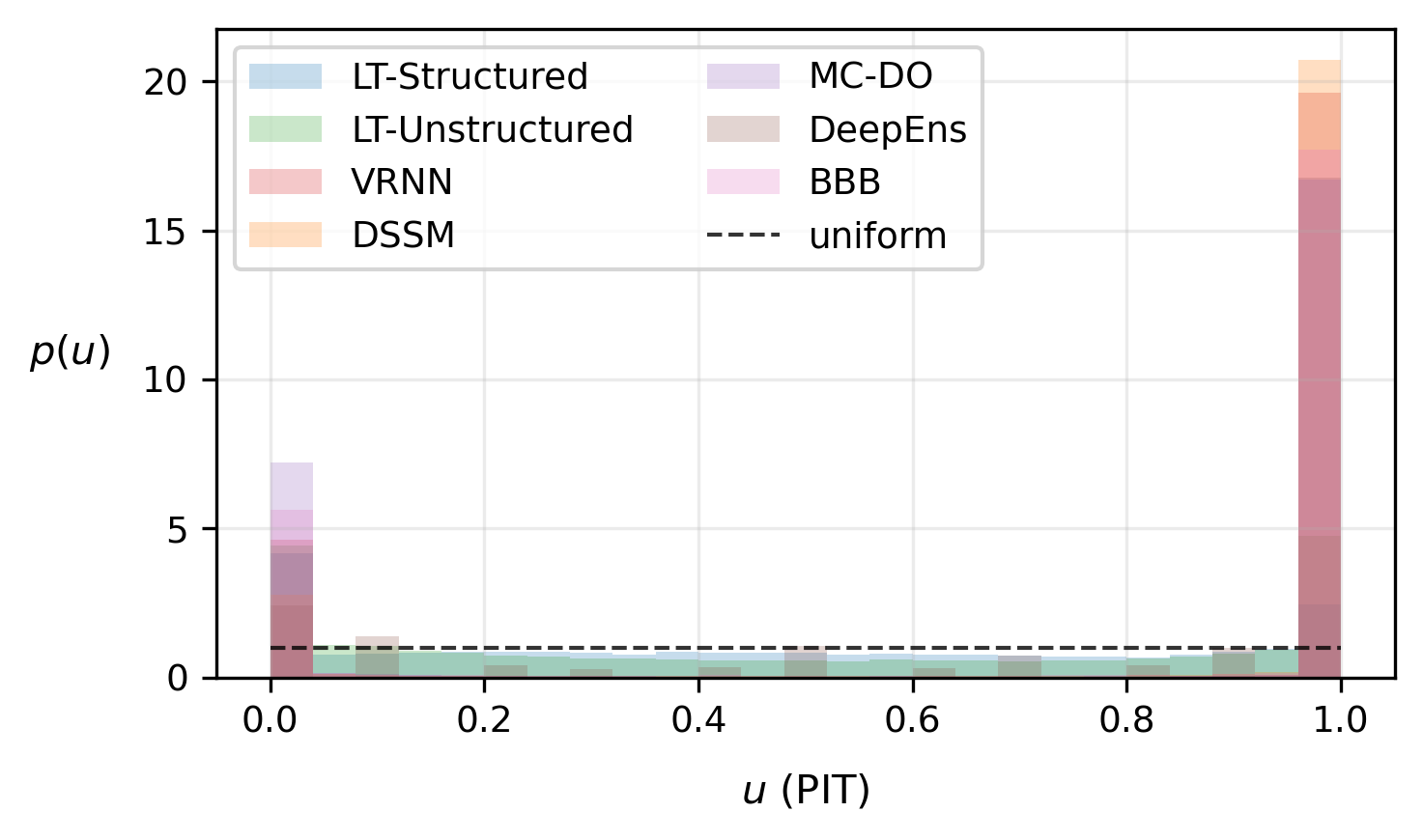}
    \caption{
    Probability integral transform (PIT) histograms for predictive distributions on the representative seed.
    For a well-calibrated probabilistic model, PIT values $u$ should be approximately uniformly distributed on $[0,1]$ (dashed line).
    LT-Structured produces a comparatively flatter PIT distribution, indicating improved calibration of predictive uncertainty, whereas several baseline methods exhibit pronounced boundary mass consistent with under- or overconfident predictions.
    } 
  \label{fig:calib_pit}
\end{figure}



\begin{figure}[t]
\centering

\subfigure[Truth vs $t$]{
    \includegraphics[width=\linewidth]{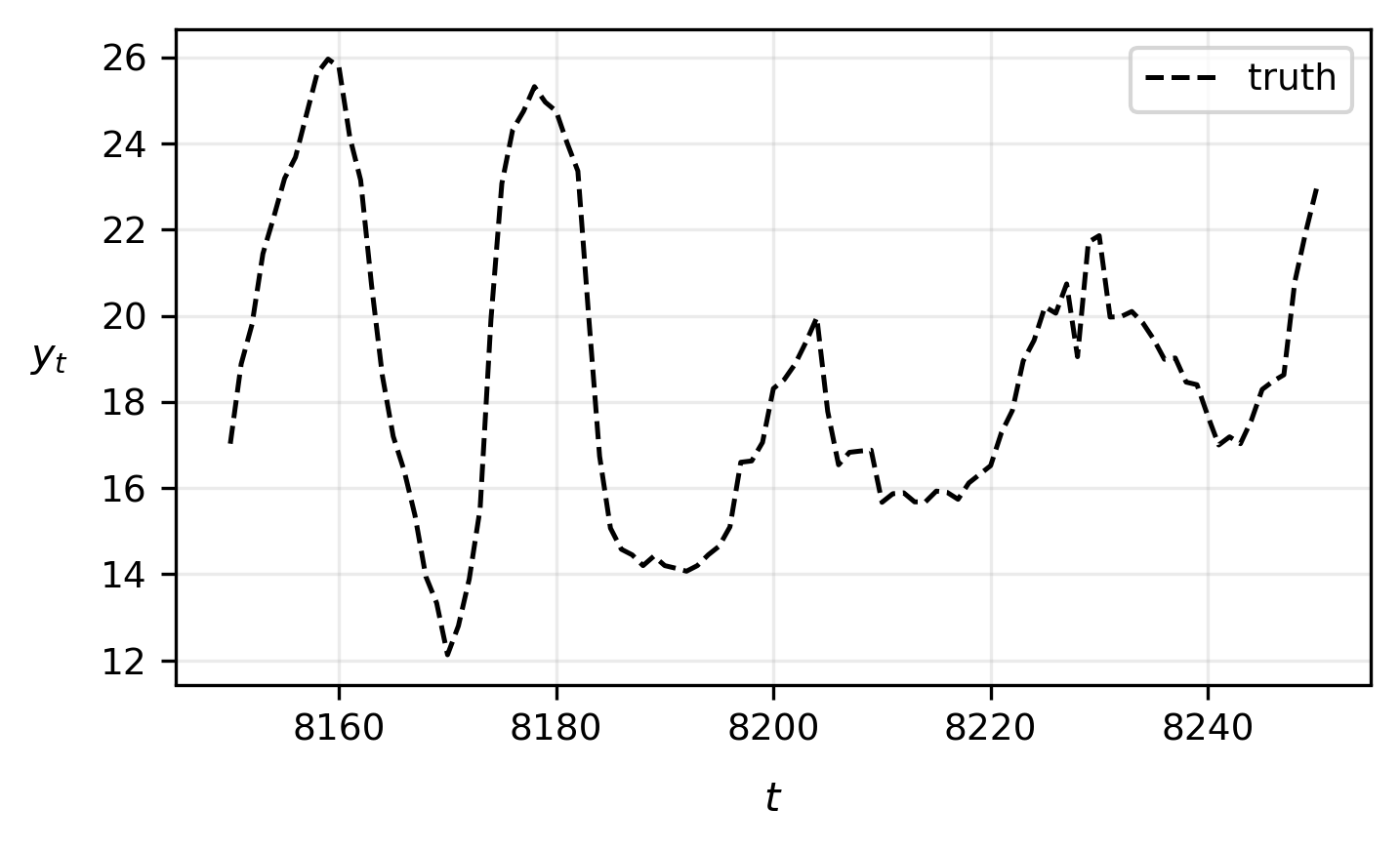}
}

\vspace{4mm}

\subfigure[Predictive mean vs $t$]{
    \includegraphics[width=\linewidth]{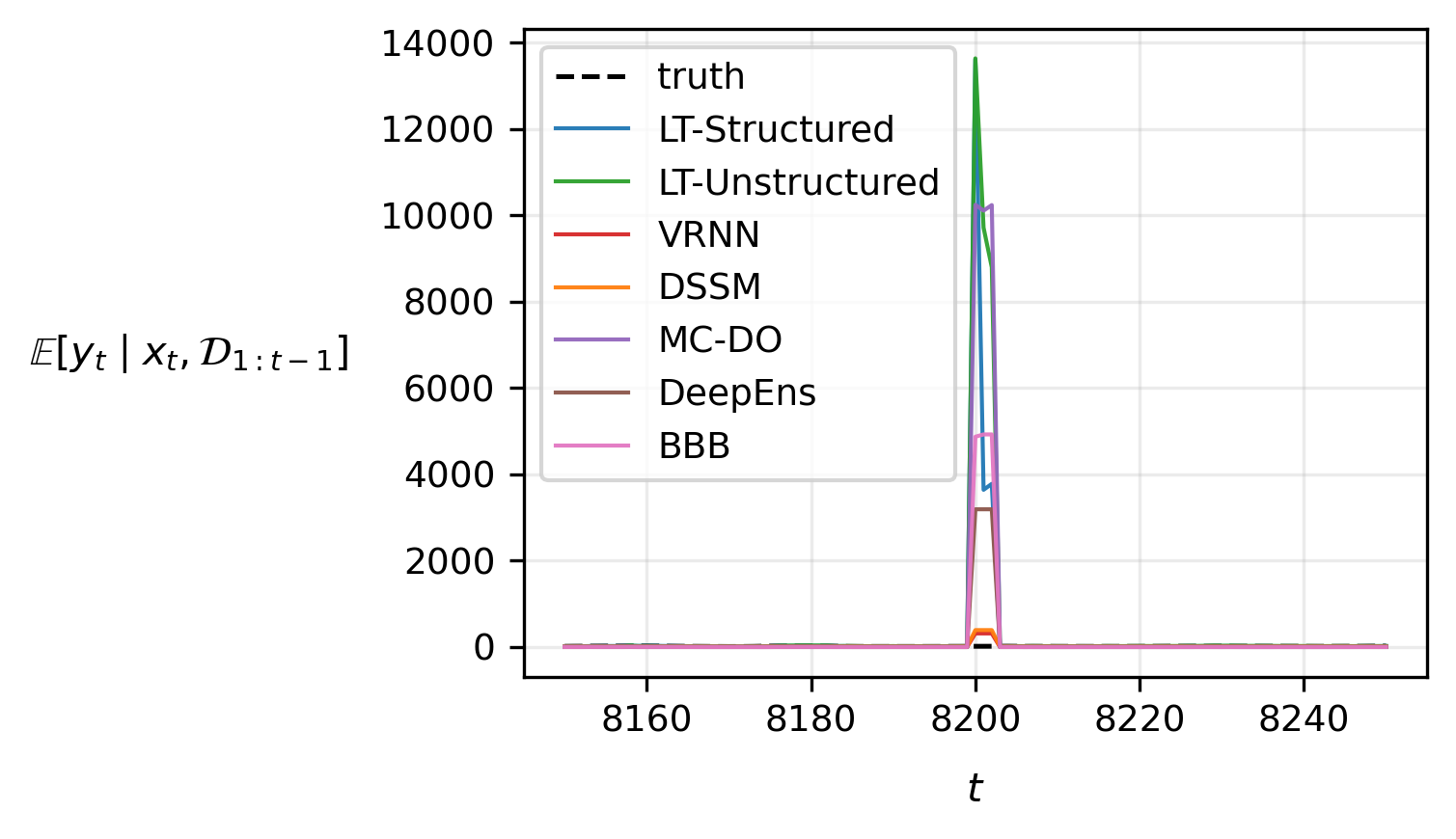}
}

\caption{Ground-truth signal and predictive mean trajectories around an isolated extreme observation.
Top: ground-truth target values $y_t$ over the selected time window.
Bottom: predictive means $\mathbb{E}[y_t \mid x_t, \mathcal{D}_{1:t-1}]$ for all models at the same time indices.
All models exhibit a sharp transient deviation at the same time step, reflecting a shared response to the extreme observation rather than model-specific instability.}

\label{fig:anom_mean_vs_time}
\end{figure}



\begin{figure}[t]
\centering

\subfigure[$\sigma_{Aleatoric}$ vs $t$]{
    \includegraphics[width=\linewidth]{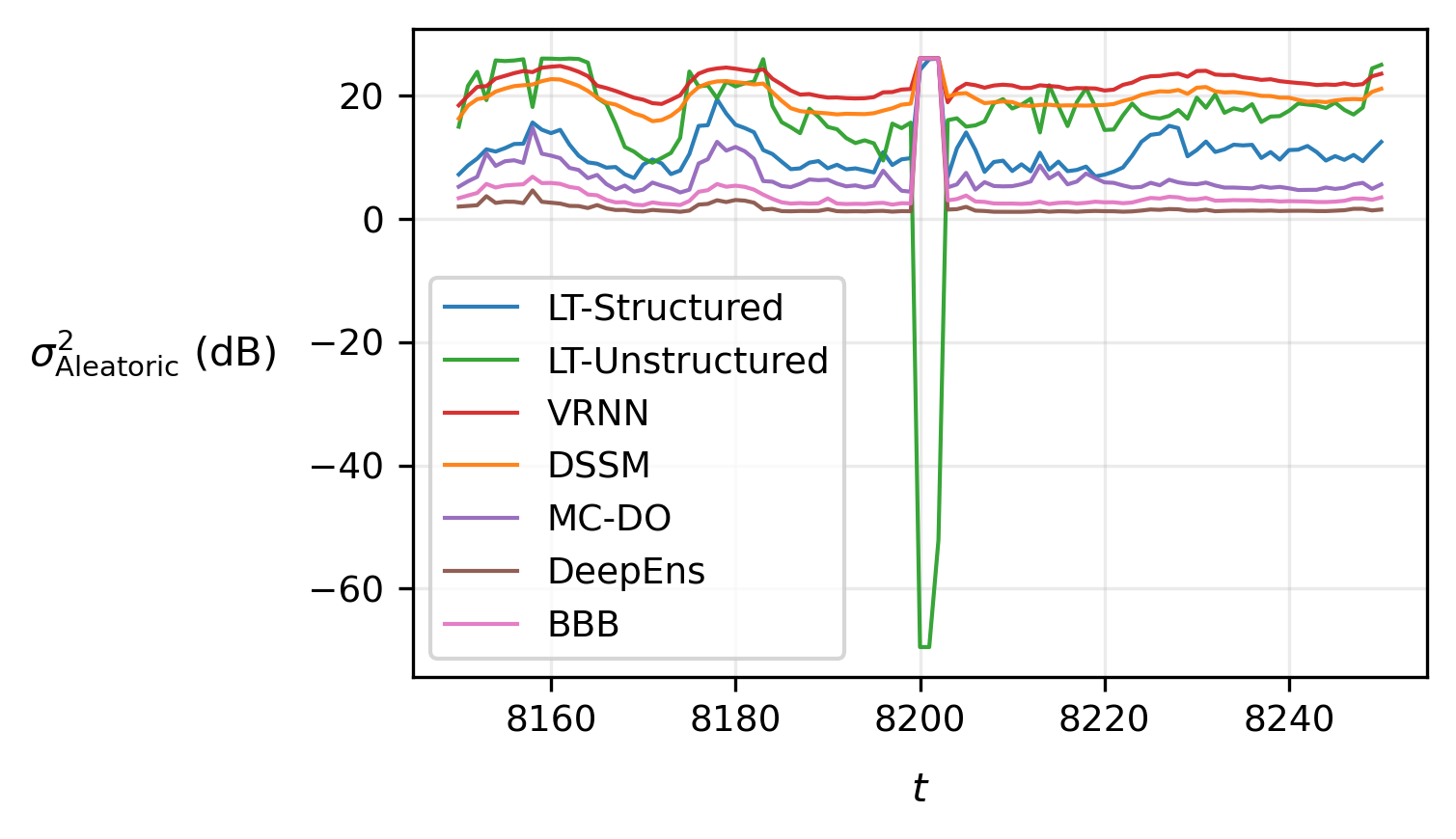}
}

\vspace{4mm}

\subfigure[$\sigma_{Epistemic}$ vs $t$]{
    \includegraphics[width=\linewidth]{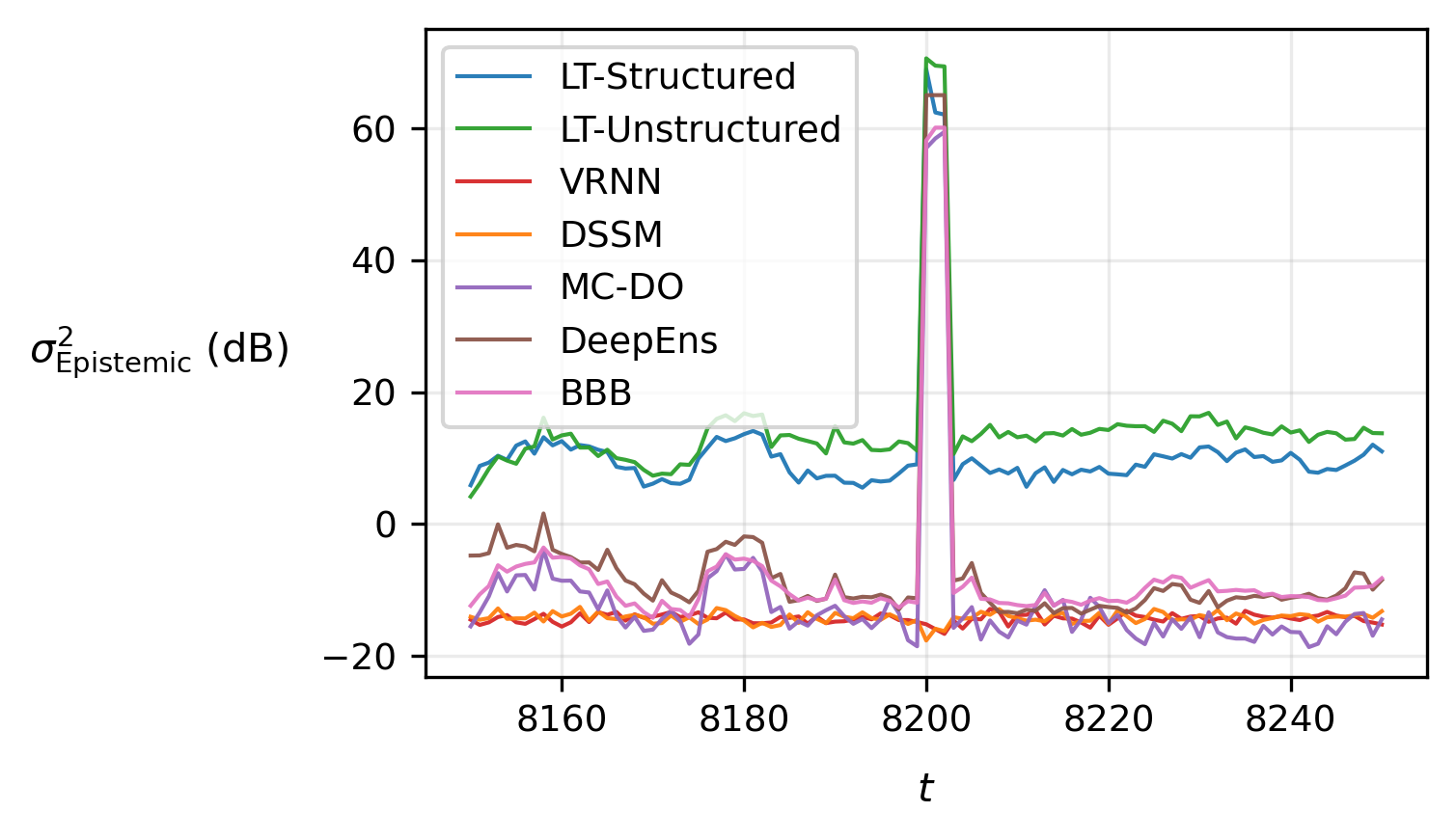}
}

\caption{Decomposition of predictive uncertainty around an isolated extreme observation (all quantities shown in dB).
Top: aleatoric variance $\sigma^2_{\text{Aleatoric}}$ over time (dB). Bottom: epistemic variance $\sigma^2_{\text{epistemic}}$ over time (dB). All models---including both stateful and static baselines---exhibit anomalous behavior at the same time step, indicating sensitivity to a shared extreme observation rather than model-specific instability. In all cases, uncertainty rapidly returns to baseline levels following the event, consistent with transient stress rather than persistent miscalibration.}

\label{fig:anom_var_vs_time}

\end{figure}


\begin{figure}[t]
\centering

\subfigure[$\sigma_{Total}$ vs $t$]{
    \includegraphics[width=\linewidth]{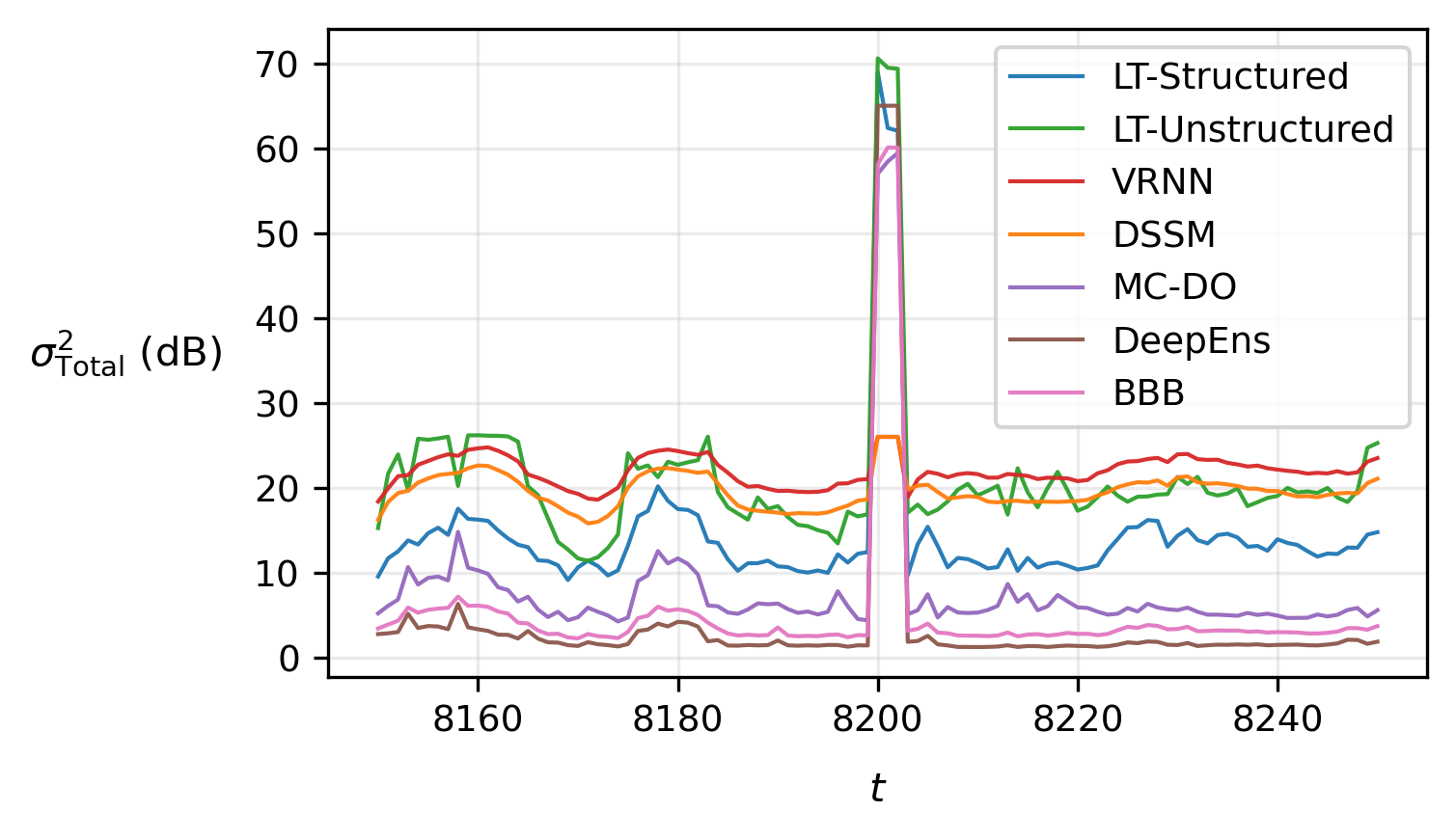}
}

\vspace{4mm}

\subfigure[$MSE$ vs $t$]{
    \includegraphics[width=\linewidth]{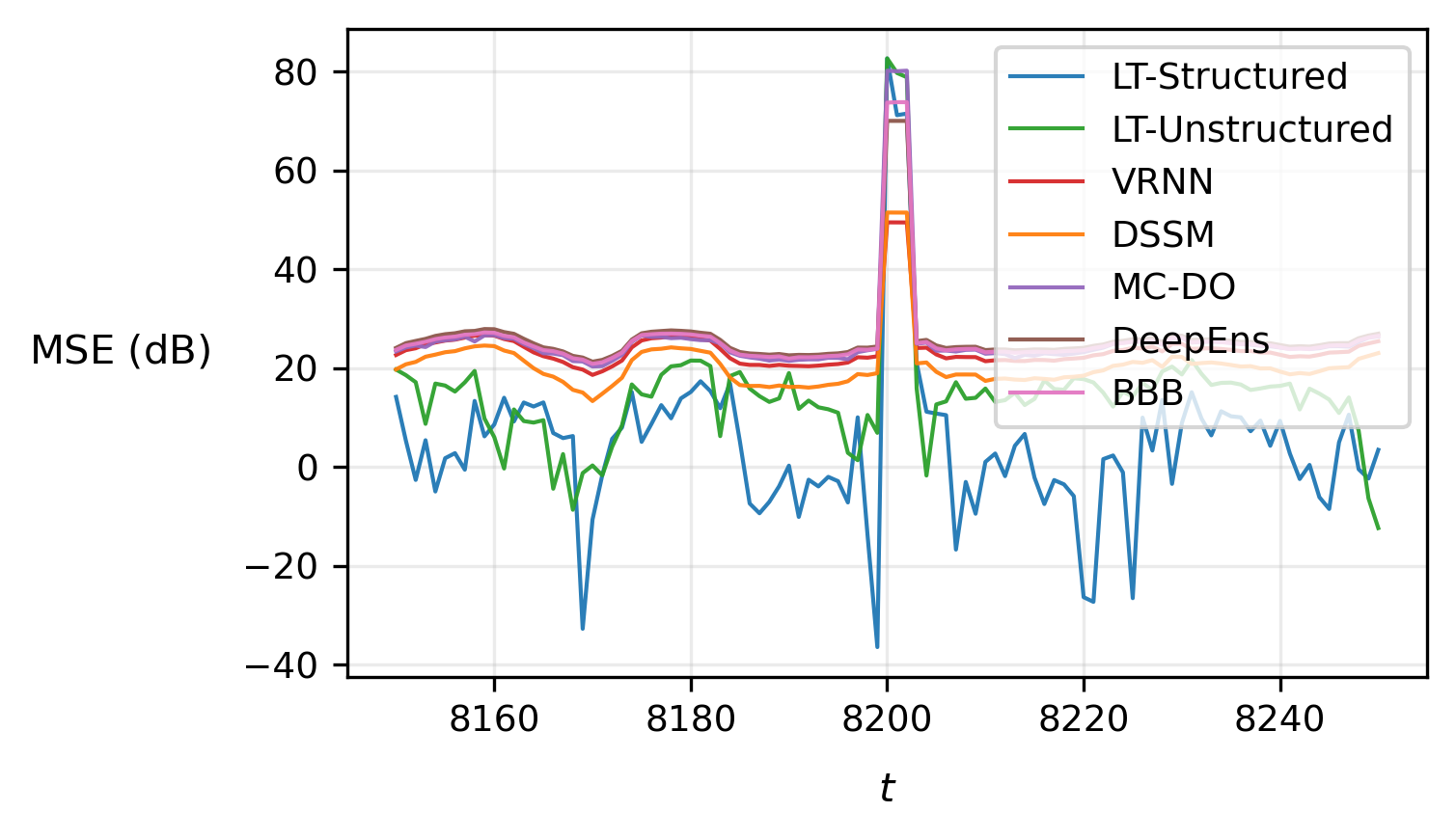}
}

\caption{Temporal evolution of total predictive uncertainty and error around an isolated extreme observation (all quantities shown in dB).
Top: total predictive variance $\sigma^2_{\text{Total}}$ over time (dB). Bottom: mean squared error (MSE) over time (dB). All models exhibit a sharp, synchronized spike at the same time step, indicating sensitivity to a shared extreme observation rather than model-specific instability. Following this event, both predictive uncertainty and error rapidly return to their typical ranges, consistent with a transient anomaly rather than sustained degradation in predictive performance.}
\label{fig:anom_mse_and_var_vs_time}

\end{figure}


\begin{figure}[t]
\centering

\subfigure[$NLL$ vs $t$]{
    \includegraphics[width=\linewidth]{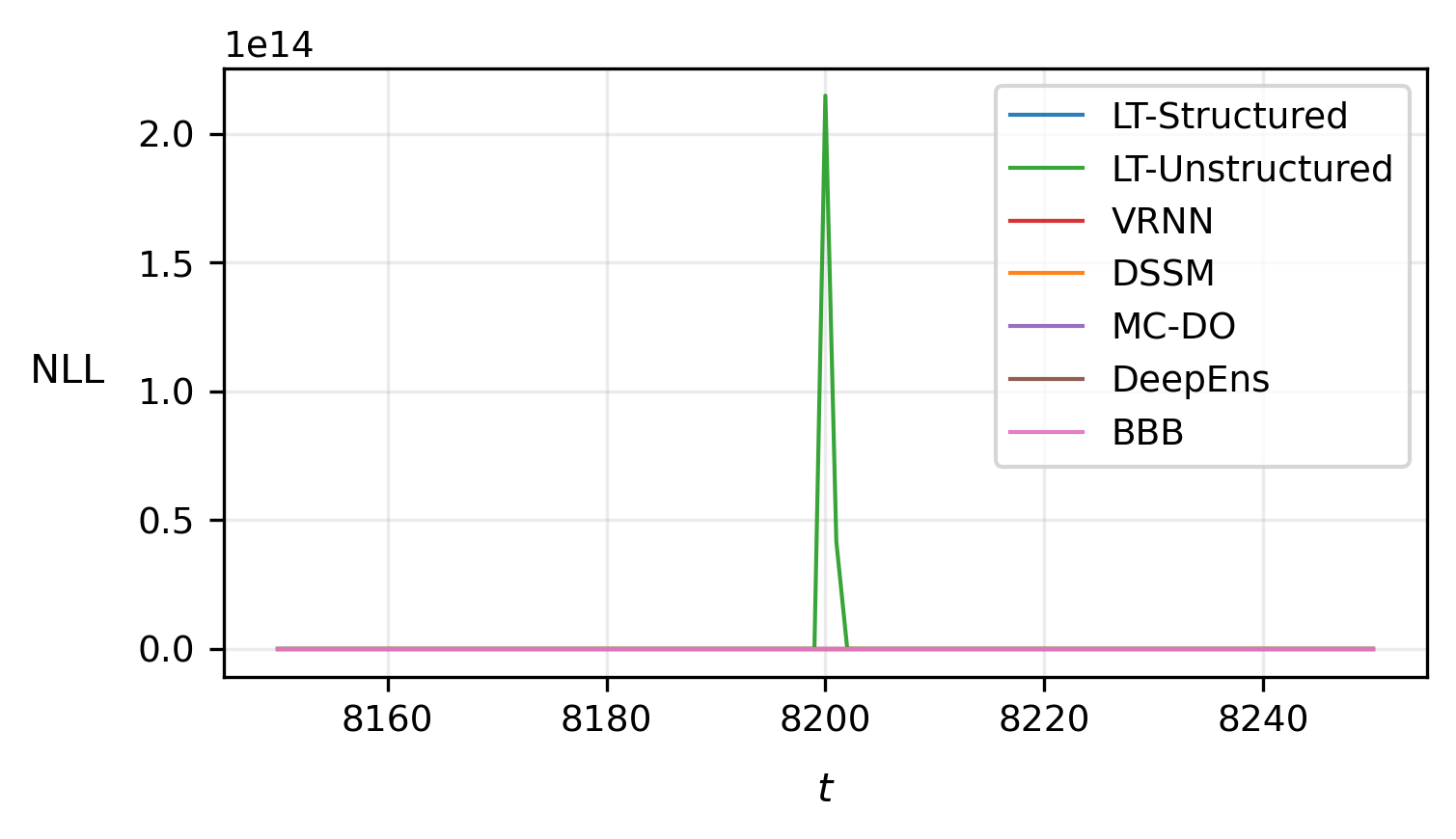}
}

\vspace{4mm}

\subfigure[$NLL$ vs $t$ (clipped)]{
    \includegraphics[width=\linewidth]{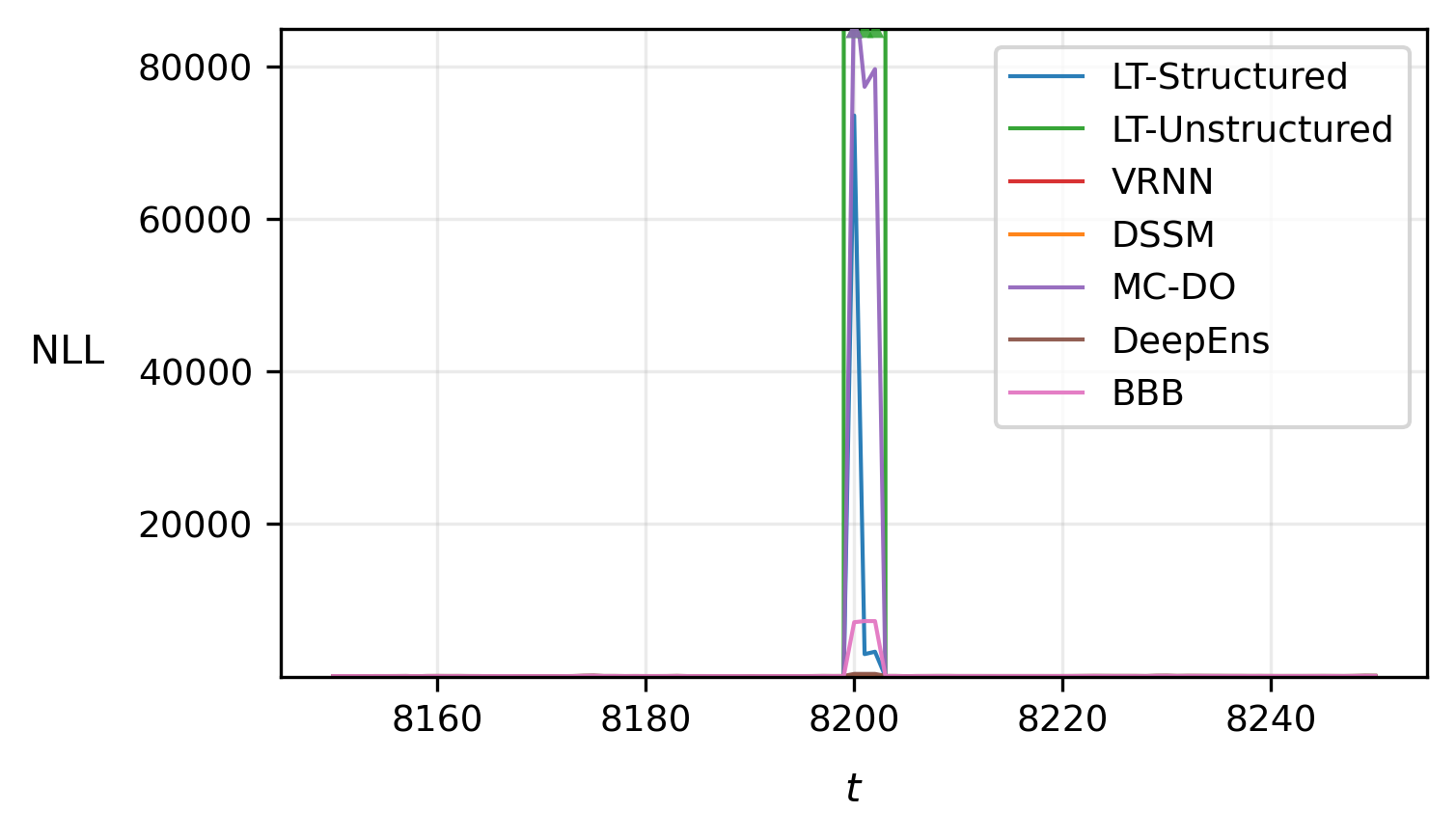}
}

\caption{Negative log-likelihood (NLL) around an isolated extreme observation.
Top: NLL over time shown on the full scale, highlighting a single catastrophic spike that dominates the dynamic range. Bottom: the same NLL with the vertical axis clipped to reveal relative behavior across models outside the extreme event. All models exhibit a synchronized spike at the same time step, indicating sensitivity to a shared anomalous observation rather than model-specific instability. Outside this isolated event, NLL values remain well-behaved and return rapidly to baseline levels.}

\label{fig:anom_nll_vs_time}

\end{figure}


\end{document}